%% file: icml2023.tex
\documentclass{article}

\usepackage{microtype}
\usepackage{graphicx}
\usepackage{subfigure}
\usepackage{booktabs} 

\usepackage{hyperref}

\usepackage[accepted]{icml2023}

\usepackage{amsmath}
\usepackage{amssymb}
\usepackage{mathtools}
\usepackage{amsthm}

\usepackage[capitalize,noabbrev]{cleveref}

\theoremstyle{plain}
\newtheorem{theorem}{Theorem}[section]
\newtheorem{proposition}[theorem]{Proposition}
\newtheorem{lemma}[theorem]{Lemma}

\theoremstyle{definition}

\theoremstyle{remark}

\usepackage[textsize=tiny]{todonotes}

\usepackage[utf8]{inputenc} 
\usepackage[T1]{fontenc}    
\usepackage{url}            
\usepackage{amsfonts}       
\usepackage{nicefrac}       
\usepackage{tcolorbox}
\usepackage{ulem,multicol}
\usepackage{booktabs} 
\tcbuselibrary{listings,breakable} 

\usepackage{natbib}

\usepackage{hyperref}       

\usepackage{etoc}

\makeatletter
\newlength{\trianglerightwidth}
\newcommand{\blocco}[1]{}

\newcommand{\bisevena}{\textit{Bi7a} }
\newcommand{\bisevenb}{\textit{Bi7b} }
\newcommand{\bisixty}{\textit{Bi60} }
\newcommand{\multi}{\textit{Mul10} }

\newcommand{\multib}{\textit{Mul100} }
\newcommand{\cnn}{\textit{Mod1} }
\newcommand{\res}{\textit{Mod2} }
\newcommand{\vgg}{\textit{Mod3} }
\newcommand{\drop}{\textit{minority initial drop} }
\newcommand{\denomin}{G^{(l)}}
\newcommand{\ttau}{\tau}

\icmltitlerunning{A Theoretical Analysis of the Learning Dynamics under Class Imbalance}

\input{defs}

\begin{document}

\twocolumn[
\icmltitle{A Theoretical Analysis of the Learning Dynamics under Class Imbalance}



\icmlsetsymbol{equal}{*}

\begin{icmlauthorlist}
\icmlauthor{Emanuele Francazi}{EPFL,Eawag}
\icmlauthor{Marco Baity-Jesi}{Eawag}
\icmlauthor{Aurelien Lucchi}{Basel}
\end{icmlauthorlist}

\icmlaffiliation{Basel}{Department of Mathematics and Computer Science, University of Basel, Switzerland}
\icmlaffiliation{EPFL}{Physics Department, EPFL, Switzerland}
\icmlaffiliation{Eawag}{SIAM Department, Eawag (ETH), Switzerland}

\icmlcorrespondingauthor{Emanuele Francazi}{emanuele.francazi@epfl.ch}

\icmlkeywords{Machine Learning, ICML}

\vskip 0.3in
]



\printAffiliationsAndNotice{}  

\begin{abstract}
Data imbalance is a common problem in machine learning that can have a critical effect on the performance of a model. Various solutions exist 
but their impact on the convergence of the learning dynamics is not understood. 
Here, we elucidate the significant negative impact of data imbalance on learning, showing that the learning curves for minority and majority classes follow sub-optimal trajectories when training with a gradient-based optimizer. This slowdown is related to the imbalance ratio and can be traced back to a competition between the optimization of different classes. 
Our main contribution is the analysis of the convergence of full-batch (GD) and stochastic gradient descent (SGD), and of variants that renormalize the contribution of each per-class gradient.
We find that GD is not guaranteed to decrease the loss for each class but that this problem can be addressed by performing a per-class normalization of the gradient.
With SGD, class imbalance has an additional effect on the direction of the gradients: the minority class suffers from a higher directional noise, which reduces the effectiveness of the per-class gradient normalization.
Our findings not only allow us to understand the potential and limitations of strategies involving the per-class gradients, but also the reason for the effectiveness of previously used solutions for class imbalance such as oversampling.
\end{abstract}
\vspace{-2mm}
\section{Introduction}
\label{sec:intro}
\vspace{-2mm}

In supervised classification, as well as other problems in machine learning, datasets are often affected by data imbalance. Although the situation can sometimes be solved or mitigated by changing the data collection method, this might induce an undesirable data shift~\cite{quinonero:09,morenotorres:12}. Importantly, many datasets are intrinsically unbalanced~\cite{van2017devil,feldman2020does,d2021tale}, as in the case of spam identification~\cite{liu:17}, fraud detection~\cite{makki:19}, or biodiversity monitoring~\cite{kyathanahally:21}.
\begin{figure}[t]
    \centering
    \includegraphics[width=1.\columnwidth]{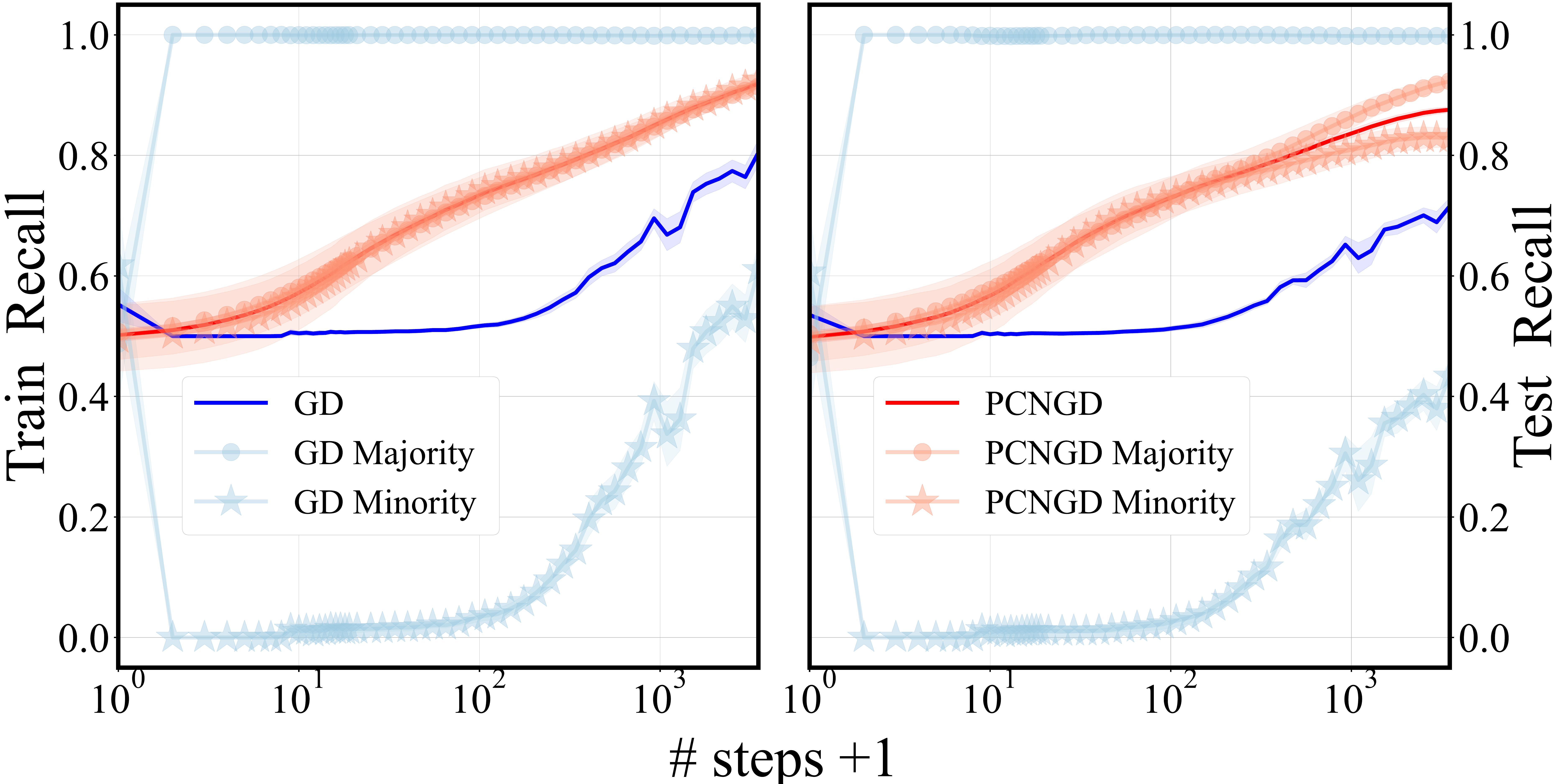}
    \vspace{-6mm}
    \caption{
    Training (left) and test (right) recall of the \cnn on the \bisixty datasets (see details in Sec.~\ref{sec:empirical_PCNGD}).
    Due to the data imbalance, GD (blue curves) first focuses on the majority class only, while the minority class stays at zero accuracy. 
    This effect is suppressed when using the PCNGD algorithm. 
    See Fig.~\ref{fig:loss60} for the related loss function curves.
}
\label{fig:warmup}
   \vspace{-10pt}
\end{figure}

The problem of data imbalance has already received attention in the literature and we broadly identify three most common ways to handle it. These either act on the data distribution \cite{he:09,huang:16}, adjust the objective function \cite{japkowicz:00, huang:16, alshammari:22} or modify the learning algorithm \cite{tang:20, anand:93}. We provide a detailed discussion of these methods in App.~\ref{Add_rel_works}. The important feature that makes our work depart from prior work is that we focus our study on the \textit{dynamics} of gradient-based learning in the presence of data imbalance. We will for instance analyze the theoretical convergence guarantees of these methods, which, to the best of our knowledge, is a novel contribution.

 \begin{figure*}[t]
    \centering
    \includegraphics[width=\textwidth]{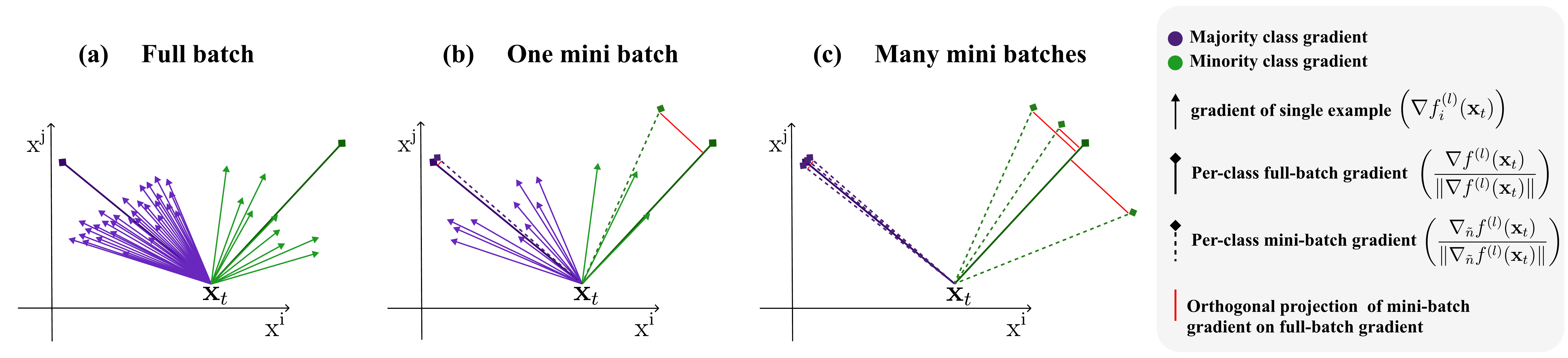}
    \vspace{-5mm}
    \caption{Diagram explaining the directional noise caused by class imbalance on (PCN)SGD {in binary classification}. We plot two generic components of the parameter vector $\x$ on the axes ($\mathrm{x^i}$, $\mathrm{x^j}$). Starting from a given iterate $\x_t$ at time $t$, the normalized per-class gradients associated with  individual batches, and the entire dataset are represented in the plots. The gradients of the individual examples that make them up are also shown.
    \textbf{(a)} The contribution of each example to the per-class full-batch gradient (FBG).
    \textbf{(b)} {Within a mini batch, instead, we consider only a randomly selected subset of dataset elements; the mini-batch gradient will therefore come from a random selection in this subset.}
    \textbf{(c)} We show several mini batches and observe that they are  more aligned to the FBG of the majority class than the FBG of the minority class. 
    We will see this has negative consequences on the dynamics. For more details, see Sec.~\ref{sec:directional-noise}.
    }
    \label{fig:diagram}
   \vspace{-10pt}
\end{figure*}

We start our investigation from the following two empirical observations: (i) The learning dynamics is delayed for imbalanced problems. {This is especially a problem during hyperparameter tuning, since it requires very long runs to assess the difference between distinct hyperparameter choices.} (ii) While the overall performance improves during the dynamics, that of the minority classes quickly deteriorates at first. 
This is shown in Fig.~\ref{fig:warmup} (blue curves) for a binary unbalanced classification problem. We call this initial deterioration the \textbf{\textit{\drop} (MID)}.\footnote{We show in Fig.~\ref{fig:Multi_PC_Comp} that the MID occurs also with SGD and multiclass classification.}
One can hypothesize why this happens {(we  will in fact  demonstrate this rigorously)}: at the beginning of learning, the gradient steps are dominated by the majority class, driving the classifier weights towards configurations that correctly classify the majority class, regardless of the minority class. In fact, due to the imbalance, decreasing the loss related to the majority class outweighs the loss increase coming from the minority class.
The minority classes will only be learned once the gradient of the majority class examples is small enough. To gain further insights, it therefore seems logical to express the loss as a sum over the examples belonging to each class, and analyze separately the gradient related to each class.

Guided by these observations, we characterize how class imbalance affects the learning dynamics, and study whether it is possible to adapt gradient-based algorithms in order to guarantee the decrease of the loss function of every single class. These adaptations provide us with a deeper insight into how class imbalance affects the training dynamics.
The main adaptation that we study is per-class normalization (PCN), \textit{i.e.} we normalize the (S)GD steps in such a way that the magnitude of the signal related to each class is the same. 
We name \textit{per-class-normalized GD (PCNGD)} the PCN version of GD. For the PCN version of SGD we use the acronym PCNSGD.


We identify our key contributions as follows:
\vspace{-3mm}
\begin{itemize}
\setlength\itemsep{0em}
 \item \textbf{Suboptimality of GD and SGD:}
 {Expanding on the observations of \cite{ye:21}, we provide further evidence for the deceleration caused by class imbalance. We relate this to the non-monotonicity of the minority class losses, providing {theorems and analytical arguments explaining the slowdown}. 
 }

 \item \textbf{Theoretical study of the PCNGD algorithm:} This algorithm is not completely new as it relates to a variant that was {empirically} motivated in~\cite{anand:93}. We derive a {theoretical convergence analysis} showing that, for a suitable step-size, PCNGD is guaranteed to decrease the loss of all classes~\footnote{More specifically, we will see that GD can also decrease the per-class loss but only under a restrictive condition on the angle between the per-class gradients.}. In addition, we also provide empirical evidence that {PCNGD performs better than GD}, in terms of both overall and per-class performance indicators (as shown in Fig.~\ref{fig:warmup}).
 
 \item \textbf{Imbalance causes directional noise in SGD:} {We show that, while normalizing the per-class gradients is sufficient to counter class imbalance in GD, this is not true for SGD, since the {imbalance influences the \textit{direction} of the per-class signal} in addition to its norm, and this suppresses the minority class learning (Fig.~\ref{fig:diagram}). 
 } 
 
 \item \textbf{A framework to understand imbalance:} 
 Our analysis also clarifies the effectiveness of other approaches already used heuristically, such as oversampling, since they can be interpreted in light of whether they help counter the directional noise; and shows how class imbalance should be seen as one of many factors (\textit{e.g.} class difficulty) that influence the per-class gradients.
\end{itemize}



\vspace{-2mm}
\section{Related work}
\label{sec:related_work}
\vspace{-2mm}

~\cite{ye:21} noticed empirically that, in imbalanced SGD dynamics, minority classes are learned later than majority ones. They call \textit{under-fitting phase of minor
classes} the initial part of the learning where the minority classes are not learned. Building on this observation, we provide a theory and didactic experiments, explaining this phenomenon formally and showing how it is qualitatively different between GD and SGD. We emphasize that this is {linked} to an initial deterioration in the minority class {recall/loss}, so we prefer to focus on avoiding the effect of the MID on the model performance, which leads to an under-fitting phase for the minor classes. 
Furthermore, \cite{ye:21} provide a method to mitigate the MID, while we show what are the theoretical requirements to eliminate it.

Noticing that the gradient magnitude of the minority class is smaller than that of the majority class, {\cite{anand:93} proposed an algorithm that takes the bisection between the per-class gradients to perform the update rule of the model parameters.
The PCN algorithms that we analyze in this work can be seen as an extension of the bisection algorithm, where the optimization steps are taken in the same direction but with a different modulus. Our adaptation allows us to use the algorithm also in the multiclass and SGD settings, and to derive convergence guarantees.}

{Gradient imbalance is also a problem discussed in multi-task learning. The latter is akin to classification problems; in fact, learning multiple classes can be thought of as learning multiple tasks together if each task is assumed to be just one class \textit{vs} the rest. The approaches proposed in \cite{chen2018gradnorm, yu2020gradient}, beyond some differences in implementation, are conceptually close to the PCN methods examined here. However, our work, analyzing separately the effects induced on GD and SGD, shows how the two cases, in classification problems, are characterized by substantial differences. In particular, we show how PCN alone is effective in the full-batch case, while in the stochastic case it is necessary to introduce variants to account, not only for the difference in norms, but also for the difference in directional noise.}
\vspace{-2mm}
\section{Structure of the article}
\vspace{-2mm}

{Sec.~\ref{sec:Deterministic_Section} focuses on full-batch algorithms. GD is analyzed in Sec.~\ref{sec:gd} where we prove that, while the dynamics converge to a critical point, the angle between the per-class gradients must be small to guarantee a monotonic decrease of the per-class loss functions and avoid the MID. The latter condition on the gradient angle becomes stronger with increasing imbalance [see Eq.~\eqref{eq:condition}]. To solve this problem, in Sec.~\ref{sec:pcngd} we introduce and study PCNGD, which equalizes the per-class gradient norms, resulting in a relaxed condition on the angles and removing the MID. We also demonstrate that for a specific type of loss function (known as gradient-dominated functions), PCNGD not only ensures a monotonic per-class loss but also converges to the global minimum (Th.~\ref{thm:convergence_PL}). In Sec.~\ref{sec:empirical_PCNGD}, we validate our conclusions through experiments.\\
In Sec.~\ref{sec:sgd}, we turn our attention to stochastic algorithms. Sec.~\ref{sec:directional-noise} shows that, unless the batch size is large, PCN is not sufficient to avoid the MID because imbalance affects not only the per-class gradient norms but also their direction (as illustrated in Fig.~\ref{fig:diagram}). Sec.~\ref{sec:pcnsgd-convergence} reaches the same conclusion by showing that the per-class losses in PCNSGD are not guaranteed to converge monotonically unless the batch size is large (Th.~\ref{th:PCNSGDdecreasing}). To support our results, Secs.~\ref{SubSec:StocAlgs} and~\ref{sec:empirical} experimentally show that using algorithms that take into account both effects induced by imbalance (the disproportions in the per-class norm \textit{and} direction) finally avoids the MID also when using stochastic gradient updates. \newline
The notation used in the article is summarized in App.~\ref{app:notation}. The theorems formulated in the main paper are in Apps.~\ref{app:convergence_GD} and~\ref{app:convergence_PCNGD}. Algorithms are discussed in more detail in App.~\ref{app:algorithms}. Model and dataset details used for experiments are in App.~\ref{app:numerical} and further experiments are in App.~\ref{app:more-exps}. App.~\ref{app:limitations} describes the limitations and ethics of this work.}

\vspace{-2mm}
\section{Convergence Guarantees with Full Batch}
\label{sec:Deterministic_Section}
\vspace{-2mm}

\begin{figure}[tb]
    \centering
    \includegraphics[width=.455\textwidth]{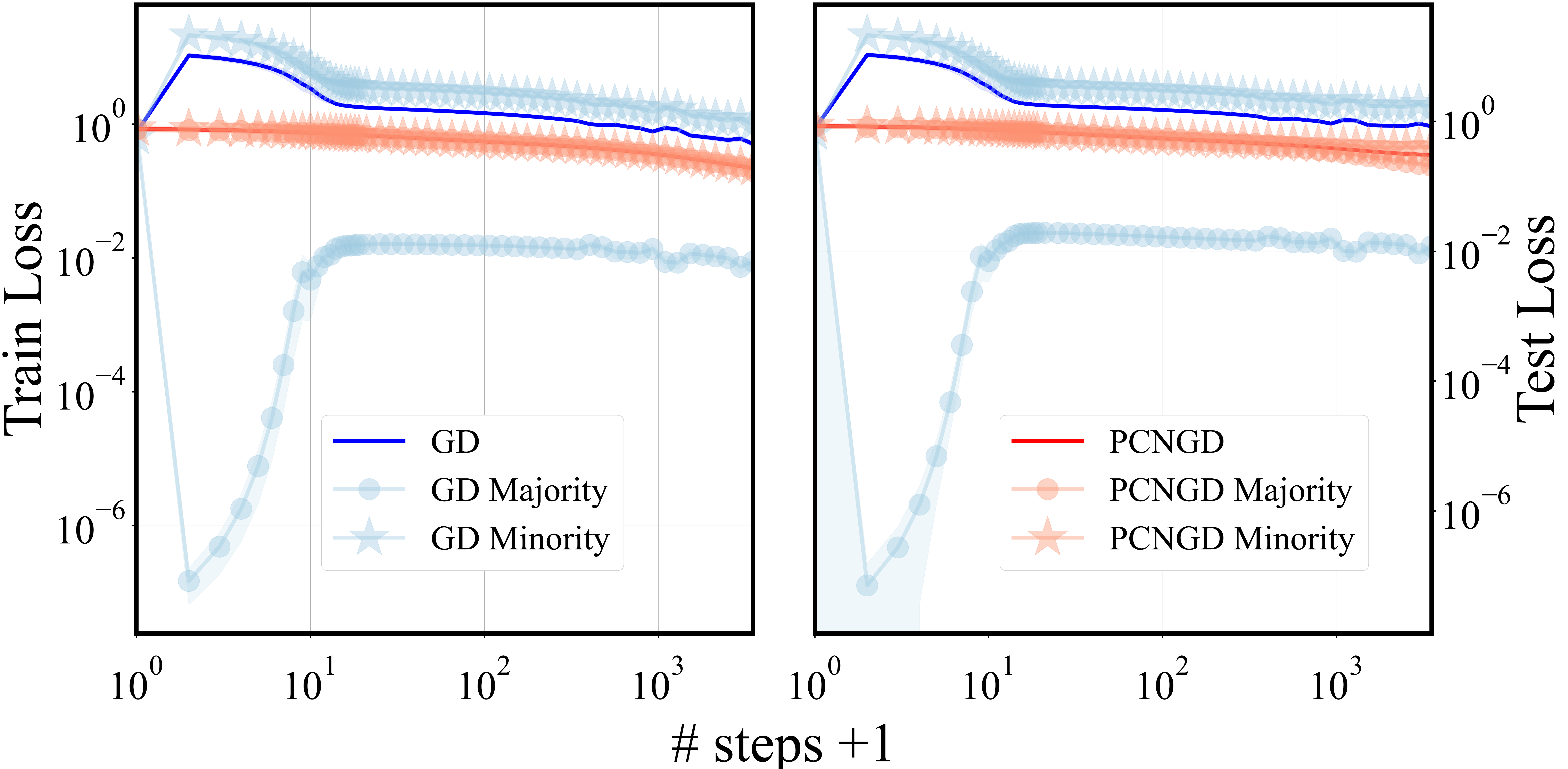}
    \caption{Training (left) and test (right) loss curves for GD and PCNGD algorithms. The corresponding recall is shown for the same setup in \ref{fig:warmup}.}
    \label{fig:loss60}
\end{figure}

{The loss curves in Fig.~\ref{fig:loss60} are illustrative of the differences between the loss of GD and PCNGD. In particular, the wide gap between per-class performances, in the early stage of GD dynamics, is absent in PCNGD. The absence of the MID phase at the beginning of the dynamics results in accelerated growth in the performances (see also Fig. \ref{fig:warmup}). In this section, we will analyze the differences in the two cases by studying the conditions that guarantee the decrease in per-class loss.}
\vspace{-3mm}
\subsection{Setting}
\vspace{-2mm}

Assume we are in a typical supervised setting where we are given a training set of samples $\D = (\xi_i, y_i)_{i=1}^n$ where $\xi_i \in \R^d$ are features, $y_i \in \{ 0, 1, \dots L - 1 \}$ are the corresponding labels and $L$ is the number of distinct labels. Our goal is to train a model parametrized by a vector $\x \in \R^m$ that minimizes a given loss function $f: \R^m \to \R$ as follows: 
$$
\min_{\x \in \R^m} \left[ f(\x) := \frac{1}{n} \sum_{i=1}^n f_i(\x) \right].
$$
\vspace{-1mm}
We also define the number of samples per class as $n_k$, and the fraction of samples per class as $\rho_k=\frac{n_k}{n}$. We will use the following notation to define the per-class losses:
$
f^{(l)}(\x) := \frac{1}{n} \sum_{i \in C_l} f_i(\x), \quad C_l = \{ i \mid y_i = l \}, \quad l \in \{ 0, 1, \dots L - 1 \}.
$

\vspace{-3mm}
\subsection{Gradient Descent}
\label{sec:gd}
\vspace{-2mm}

We start by analyzing the convergence of gradient descent run for $T$ iterations on the loss corresponding to each class $f^{(l)}(\x)$. For the sake of clarity, we prove our results for the binary case, where the number of classes $L = 2$ and $y_i \in \{0, 1 \}$, but we note that these results are extendable to the case where $L > 2$. {For example, by aggregating classes, our results are exact for step imbalance\footnote{Step imbalance: There is a set of majority classes each with $n_1$ elements, and a set of minority classes, all with $n_2$ elements.}
with any number of classes. Also, multiclass classifiers can be expressed as a combination of binary problems.} Additionally, in App.~\ref{app:multiclass-gd}, we provide a multiclass version of Th.~\ref{th:GDdecreasing} and~\ref{th:PCNGDdecreasing}.

To make our results more accessible, in the main paper we give an informal version of our theorems. We invite the reader to check App.~\ref{app:convergence_GD} and~\ref{app:convergence_PCNGD} for a formal version. In App.~\ref{app:convergence_GD} we also mention literature with similar convergence analyses.  Since we do not assume that the objective function $f$ is convex, we focus on showing that the gradient norms decrease with the number of iterations (which is the standard metric in the optimization literature, see \textit{e.g.}~\cite{ghadimi:13}). Specifically, our complexity measure will be $\min_{s \in S} \| \nabla f^{(l)}(\x_s) \|^2 \leq \epsilon$ for a given $\epsilon$ accuracy and over an interval $S$.


We first prove the per-class convergence of GD under imbalance, highlighting the requirements to obtain it. We denote the angle between two vectors $\x, \y \in \R^m$ by $\angle(\x,\y)$. We also define $C_t := \tfrac{\| \nabla f^{(1-l)}(\x_{t}) \|}{\| \nabla f^{(l)}(\x_{t}) \|}$.

\begin{theorembox}[Informal]
Assume that each $f^{(l)}$ for $l=0,1$ is $L_1$-Lipschitz and $L_2$-smooth\,\footnote{By $L_2$-smooth we mean that $\nabla f^{(l)}$ is $L_2$-Lipschitz continuous.
} and let $\alpha(\x_t) = \angle(\nabla f^{(l)}(\x_{t}), \nabla f^{(1-l)}(\x_{t}))$. If for all iterations $t\in [0,T-1]$, with $T < \infty$,
one has $\| \nabla f^{(1-l)}(\x_{t}) \| \neq 0$ and $\cos(\alpha(\x_t)) > - 1/C_t$, then, for all $\tilde T \in [0,T-1]$, the iterates of gradient descent with step size $\eta_t = \bigO\left(\tfrac{1}{\sqrt{\tilde T}}\right)$ satisfy
\begin{equation*}
\min_{t \in [0,\tilde T-1]} \| \nabla f^{(l)}(\x_{t}) \|^2 \leq  \tfrac{K}{\sqrt{\tilde T}} ,
\end{equation*}
for some constant $K$, independent from $\tilde T$.

\label{th:GDdecreasing}
\end{theorembox}
We stress that the upper bound $\mathcal{O}(\tfrac1{\sqrt{\tilde T}})$ depends on the imbalance. We show it explicitly in App.~\ref{app:convergence_GD} (constant {$C_t$}).

{Theorem~\ref{th:GDdecreasing} requires a restrictive condition on the angle $\alpha(\x_t)$ in order to guarantee a decrease of the loss of both classes; we find that (App.~\ref{app:convergence_GD}), in order to have a strictly decreasing loss for both classes, the angle between the per-class gradients must meet the condition
\vspace{-2mm}
\begin{equation}\label{eq:condition}
1 + \cos(\alpha(\x_t)) C_t > 0.
\end{equation}
We therefore see that this condition can not be satisfied when considering worst-case guarantees, especially when $C_t$ is large.
While the above result provides an upper bound, we discuss the tightness of this condition in App.~\ref{app:convergence_GD}, effectively demonstrating that one can not avoid this problem in the general case: GD
can not guarantee monotonic convergence for each class even for convex functions. }
Interestingly, condition \eqref{eq:condition} has an intuitive meaning.
Indeed, one can see that it depends on the ratio of the norms of the per-class gradients denoted by $C_t$, which at least in the initial phases of learning is proportional to the imbalance ratio. In the case where the two norms are equal, Eq.~\eqref{eq:condition} is trivially satisfied \textit{almost always} if the classes are equally difficult. On the contrary, if one gradient norm dominates the other, GD will only minimize the gradient of one class, until it gets to a point where the two gradient norms start having comparable values. As we will see in our experimental results, this leads to a suboptimal behavior if one equally cares about the loss of each class. 
{Building on this finding, we will now introduce a variant of GD which compensates the asymmetry in the gradients. Specifically, we will show that the restrictive condition required in Eq.~\eqref{eq:condition} can essentially be removed.}
\vspace{-3mm}
\subsection{Per-Class Normalized Gradient Descent}\label{sec:pcngd}
\vspace{-2mm}



Algorithmic solutions to class imbalance typically rescale the contribution of each gradient~\cite{anand:93}. Inspired by this prior work we present an algorithm named PCNGD that, starting from an initial guess $\x_0$, iteratively updates the parameter $\x_t$ as follows,
\begin{equation}\label{eq:pcngd}\tag{PCNGD}
\x_{t+1} = \x_t - \eta_t \left( \frac{\nabla f^{(0)}(\x_t)}{\| \nabla f^{(0)}(\x_t) \|} + \frac{\nabla f^{(1)}(\x_t)}{\| \nabla f^{(1)}(\x_t) \|} \right).
\end{equation}
This algorithm forces the updates to be independent of the gradient norm related to each class, but it does not act on the angle between the gradients, so the size of the steps depends on the angle between the gradients.

We want to study the convergence behavior of the loss corresponding to each class. We derive such an analysis in the broad setting where the loss function is smooth and non-convex.

\begin{theorembox}[Informal]
Assume that each $f^{(l)}$ for $l=0,1$ is $L_1$-Lipschitz and $L_2$-smooth. 
If, for all iterations $t\in [0,T-1]$, with $T < \infty$, one has $\cos \alpha(\x_t) \neq -1$, then, for all $\tilde T \in [0,T-1]$, the iterates of PCNGD with step size $\eta_t = \bigO\left(\tfrac{1}{\sqrt{\tilde T}}\right)$ satisfy
\begin{align*}
\min_{t \in [0,\tilde T-1]} \| \nabla f^{(l)}(\x_{t}) \| \leq  \tfrac{K}{\sqrt{\tilde T}},
\end{align*}
for some constant $K$, independent from $\tilde T$.
\label{th:PCNGDdecreasing}
\end{theorembox}
\begin{proof}[Proof sketch for $f^{(0)}$]


Since $f^{(0)}$ is $L_2$-smooth, we have
\begin{align*}
& f^{(0)}(\x_{t+1}) \\
&\leq f^{(0)}(\x_{t}) + \langle \nabla f^{(0)}(\x_{t}), \x_{t+1} - \x_t \rangle + \tfrac{L_2}{2} \sqnorm{\x_{t+1} - \x_t} \\
&\leq f^{(0)}(\x_{t}) - \eta_t (1 + \cos \alpha(\x_t)) \| \nabla f^{(0)}(\x_{t}) \| + 2 L_2 \eta_t^2,
\end{align*}
where we used the update step of PCNGD in the second equation.

Let $\omega_{\min} := \min_{t \in [0,T-1]} (1 + \cos \alpha(\x_t))$. By rearranging the terms in the equation above, we get
\begin{equation*}
\| \nabla f^{(0)}(\x_{t}) \| \leq \frac{1}{\eta_t \omega_{\min}} [f^{(0)}(\x_{t}) - f^{(0)}(\x_{t+1})] + \frac{2 L_2 \eta_t}{\omega_{\min}}
\end{equation*}

Taking the minimum over $t$, we get
\begin{equation*}
\min_{s \in [0,\tilde T-1]} \| \nabla f^{(0)}(\x_{s}) \| \leq \tfrac{1}{\tilde T} \sum\nolimits_{t=0}^{\tilde T-1} \|\nabla f^{(0)}(\x_{t}) \|.
\end{equation*}
\vspace{-2mm}

We conclude the proof by combining the last two equations and by choosing the step size stated in the theorem.

\end{proof}
\vspace{-2mm}

The full proof is given in App.~\ref{app:convergence_PCNGD} for different step-size schedules and for randomized iterates (as done in~\cite{ghadimi:13}).
An important feature of the above theorem is that it guarantees that each per-class loss function $f^{(l)} $decreases at the same rate. We note that the assumption $\cos \alpha(\x_t) \neq -1$ implies that the two gradients are not allowed to be in completely opposite directions. {This condition is a much milder assumption than the one required for GD in Theorem~\ref{th:GDdecreasing} (it is as restrictive as in balanced learning, Eq.~\eqref{eq:condition}), and does not become more restrictive when the class imbalance increases}.

Note that for a small enough step size, the per-class loss decrease is monotonic [Eq.~\eqref{eq:pcngd-decrease}]. This is desirable since it means that the MID is avoided. This does not however guarantee convergence to a global minimum, but we can ensure a per-class monotonous convergence to a global minimizer under an additional assumption shown next.

\vspace{-3mm}
\paragraph{Gradient-dominated functions}

We prove convergence of PCNGD for a class of gradient-dominated functions~\citep{nesterov2006cubic} which are related to the Polyak-Łojasiewicz (PL) condition~\cite{karimi2016linear} that has been shown to hold for overparametrized neural networks~\cite{liu2022loss}. Instead of requiring this variant of the PL condition to hold for $f$, we require it to hold for each class separately. Specifically, we say that a function satisfies the \textit{class-GD} inequality if the following holds for each class $l$,
\vspace{-1mm}
\begin{equation*}
\frac12 \| \nabla f^{(l)}(\x) \| \geq \mu^{(l)} | f^{(l)}(\x) - f^{(l)}(\x^*) |, \quad \forall \x \in \R^m,
\end{equation*}
\vspace{-1mm}
for some constant $\mu^{(l)} > 0$. For finite-sum objective functions, the class-GD inequality implies the gradient dominated inequality.

\begin{theorembox}
Assume that each $f^{(l)}$ for $l \in \{0,1\}$ is $L_2$-smooth and $\mu$-class-GD.
If, for all iterations $t \in [0,T-1]$, with $T < \infty$, $\cos \alpha(\x_t) \neq -1$, then, for all $\tilde T \in [0,T-1]$, the iterates of PCNGD with a decreasing step satisfy
\begin{align}
f^{(l)}(\x_{\tilde T}) - f^{(l)}_* \leq  \tfrac{K}{\tilde T},
\end{align}
for some constant $K$, independent from $\tilde T$, for each $l \in \{0,1\}$, with $f^{(l)}_* = \min_{\x \in \R^d} f^{(l)}(\x)$.
\label{thm:convergence_PL}
\end{theorembox}
We prove theorem~\ref{thm:convergence_PL} in App.~\ref{app:convergence_PCNGD}.

\vspace{-3mm}
\subsection{Empirical evidence in favor of PCNGD}
\label{sec:empirical_PCNGD}
\vspace{-2mm}

We consolidate the findings of our paper with experiments, and provide our code on \href{https://github.com/EmanueleFrancazi/PCNGD-Algorithms}{GitHub} (see also App.~\ref{app:numerical}).

In Fig.~\ref{fig:warmup} we compare GD and PCNGD with a 60:1 imbalance.
In GD, within the first epoch, the recall of the majority class goes to 1, and that of the minority class drops to 0.
As shown in Sec.~\ref{sec:gd}, the intuition behind this is that with GD in an unbalanced dataset, the gradient points predominantly toward the direction dictated by the majority class. 
Since the norm of the gradient related to a class scales with its abundance, in order to appreciate the signal of the minority class, the majority class gradient needs to become smaller by an amount that scales with the imbalance ratio $\rho$.\footnote{Therefore, we expect that the time required to learn the minority class scales with $\rho$.}\\
With PCNGD, instead, the MID is suppressed, since both classes are optimized from the beginning of the run.
After the first $\sim200$ epochs, while the model trained with GD is not yet learning, the PCNGD model already surpasses the peak performance achieved by GD within the entire 3500 epochs of the experiment.

We confirm these observations over several architectures and datasets. 
We used three different architectures, which we name \cnn\!, \res\!, \vgg\!, and five different imbalanced binary and multiclass datasets: \bisevena\!, \bisevenb\!, \bisixty\!, \multi\!, \multib \!. 
In all cases, while the training set is unbalanced, the test set is balanced. We give full details about the models in App.~\ref{Appendix:architecutes_info} and about the datasets in App.~\ref{Appendix:Data&HP}. The results of the experiments are reported in App.~\ref{app:det-exps}

\vspace{-3mm}
\subsection{On the better generalization of PCN algorithms}
\label{sec:generalize}
\vspace{-2mm}

In the experiments shown in Figs~\ref{fig:warmup}, \ref{fig:loss60}, and~\ref{fig:det_group}, PCNGD
not only has a better training performance, but it also has a better test performance.
To rationalize the better generalization of PCNGD, we can look at the fixed points of the dynamics.
In GD, fixed points satisfy $ \nabla f_{n_0}^{(0)} =- \nabla f_{n_1}^{(1)}$. If the training set is imbalanced and the per-class gradients are not exactly zero, this will imply that the minority gradients are about a factor $\rho$ larger than the majority ones, which is not what we would expect from a balanced dataset.
On the contrary, PCNGD allows an infinitely larger quantity of fixed points, since now the condition becomes
$\nabla f_{n_0}^{(0)} = -\gamma\nabla f_{n_1}^{(1)} $, $\forall\gamma>0$. If on one side this makes us expect a higher variability in the found solutions (which might be beneficial for ensembling~\cite{kyathanahally2022ensembles} or stochastic weighted averaging~\cite{izmailov:18}), on the other it means that the fixed points are insensitive to the data imbalance. \\
Another reason for a better generalization can be taken from \cite{ye:21}, who argue that the MID itself induces overfitting. Therefore, eliminating the MID can on its own be a reason for an improved test performance.


\vspace{-2mm}
\section{SGD: Data Imbalance affects both intensity and direction of the signal}
\label{sec:sgd}
\vspace{-2mm}

To elucidate the effect of imbalance on SGD, we turn our attention toward a stochastic variant of PCNGD. Perhaps the most obvious way to adapt Eq.~\eqref{eq:pcngd} to a stochastic setting is to replace the full gradient by a stochastic estimate computed over a mini-batch, with the condition that at least one element of each class is present in each minibatch.\footnote{This can be done in several manners, as long as the batch size is larger than the number of classes. For example, one possible approach is to discard the batches that do not contain both classes and reshuffle the data at every epoch.} We call the resulting algorithm Per-Class Normalized SGD (PCNSGD), whose update is defined as
\vspace{-2mm}
\begin{equation}\label{eq:pcnsgd-wrong}\tag{PCNSGD}
\x_{t+1} = \x_t - \eta_t \left( \frac{\nabla_{\tilde n} f^{(0)}(\x_t)}{\| \nabla_{\tilde n} f^{(0)}(\x_t) \|} + \frac{\nabla_{\tilde n} f^{(1)}(\x_t)}{\| \nabla_{\tilde n} f^{(1)}(\x_t) \|} \right)\,,
\end{equation}
where $\nabla_{\tilde n} f^{(l)}(\x_t)$ is the gradient over a random batch of size $\tilde n$, {taking only} examples belonging to the class $l$.

\vspace{-3mm}
\subsection{Imbalance causes directional noise}
\label{sec:directional-noise}
\vspace{-2mm}

\begin{figure}[t]
    \centering
    \includegraphics[width=0.9\columnwidth]{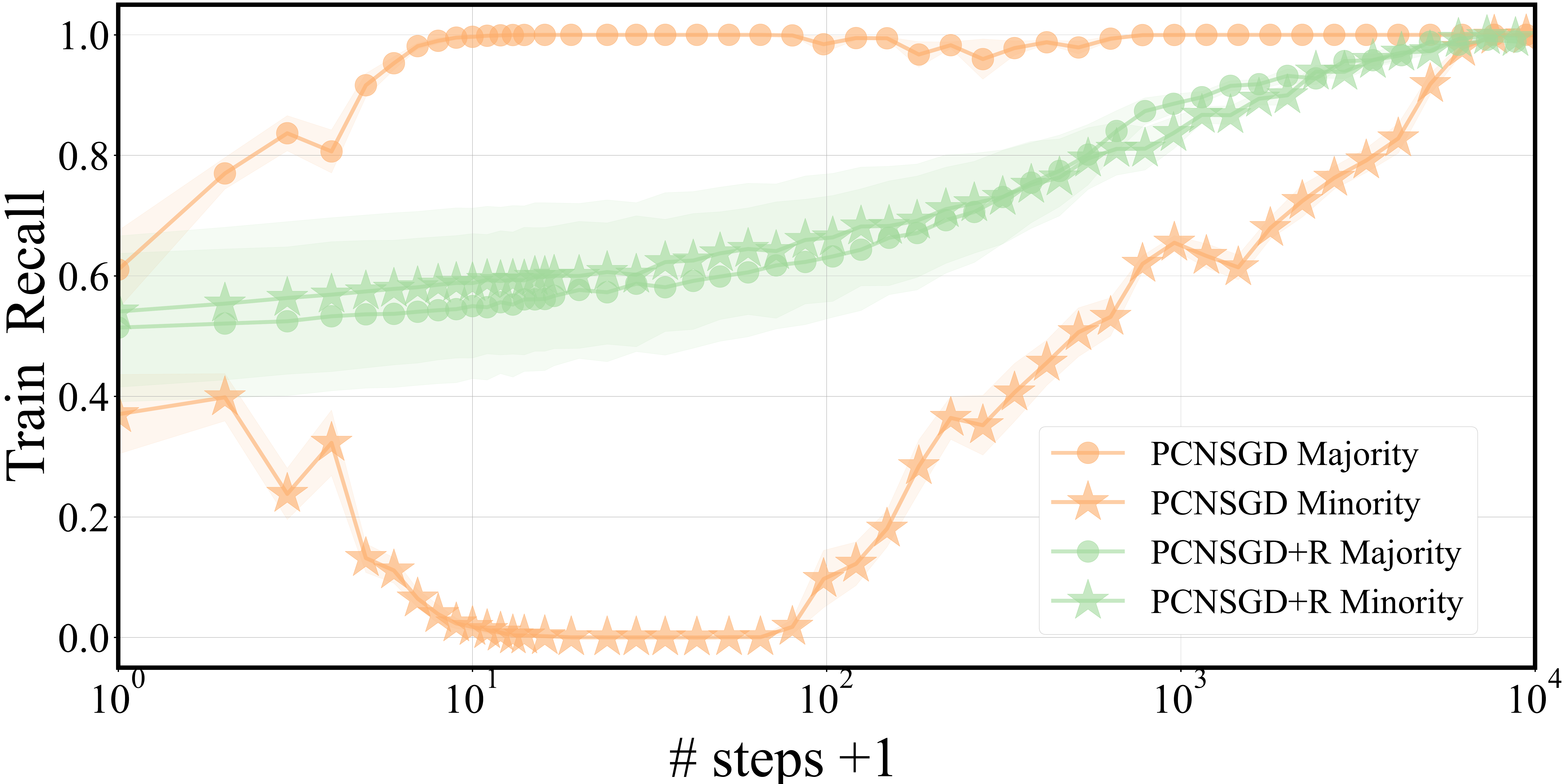}
    \caption{Per-class train recall of PCNSGD and of PCNSGD+R with \cnn on the \bisevena dataset. With the rescaling proposed in Eq.~\eqref{eq:rescaling}, leading to PCNSGD+R, the MID disappears. The macro-averaged curves are shown in Fig.~\ref{fig:PCNSGD}.}
   \label{fig:PCNSGD-wrong}
   \vspace{-10pt}
\end{figure}

In contrast to PCNGD, we empirically observed that PCNSGD still suffers dramatically from the MID, despite the correction on the gradient norms (Figs.~\ref{fig:PCNSGD-wrong} and~\ref{fig:PCNSGD}--orange curves). 


To understand why normalizing the per-class gradient norms does not have the same benefit as with GD, we take a closer look at the computation of the stochastic gradients.
To do so, we define $\nabla f_{\infty}^{(l)}(\x_t)$ as the mean gradient direction corresponding to class $l$ on the population loss, and we call $\nabla f_{n_l}^{(l)}(\x_t)$ the full-batch mean  gradient relative to class $l$.
More generally, given an arbitrary set $S$ of $\tilde n_l$ elements, all belonging to the same generic class $l$, the notation $\nabla f_{\tilde n_l}^{(l)}(\x_t)$  indicates:
\vspace{-2mm}
\begin{equation}
    \nabla f_{\tilde n_l}^{(l)}(\x_t) := \tfrac{1}{\tilde n_l} \sum_{i \in S} \nabla f_{i}(\x_t)
\end{equation}
where $f_{i}(\x_t)$ is the loss function corresponding to element $i \in S$.
If the underlying gradient distribution has a finite variance,\footnote{This condition can be relaxed by using a generalized CLT~\cite{darling:56, lam2011corrections}.} by the Central Limit Theorem (CLT) the per-class gradient calculated on $\tilde n_l$ examples can be written as fluctuations around the population gradient,
\begin{equation}\label{eq:clt}
\nabla f_{\tilde{n}_l}^{(l)} = \nabla f_{\infty}^{(l)} + \tfrac{1}{\sqrt{\tilde{n}_l}} \Z^{(l)}  +o\left( \tfrac{1}{\sqrt{\tilde{n}_l}} \right) \,,
\end{equation}
where $\Z^{(l)}$ is a zero-average multivariate Gaussian random variable whose distribution is fixed by the covariance matrix associated with the gradients of the $l^\mathrm{th}$ class.

We now consider a single batch; from now on $\tilde{n}_l$ will denote the number of elements, belonging to class $l$, present in the batch.
We want to compare $\nabla f_{\tilde{n}_l}^{(l)}$ with the corresponding full-batch gradient (FBG), in the regime where for both classes the batch size is much smaller than the size of the dataset, \textit{i.e.} $\tilde n_l\ll n_{l'}~\forall l,l'=0,1$. In this case, we can use the CLT again, obtaining
\vspace{-1mm}
\begin{equation}\label{eq:approx}
\nabla f_{\tilde{n_l}}^{(l)} \simeq \nabla f_{n_l}^{(l)} + \tfrac{1}{\sqrt{\tilde{n_l}}} \Z^{(l)} \equiv \denomin \,,
\end{equation}
\vspace{-1mm}
where we neglected terms of order $1/\sqrt{n_l}$.

Let us now take the updates in Eq.~\eqref{eq:pcnsgd-wrong}, and see how normalizing the per-class gradients over the minibatch differs from normalizing them over the full gradient.
By using Eq.~\eqref{eq:approx},
the PCNSGD steps can be written as 
\vspace{-1mm}
\begin{equation}\label{eq:pcn}
 \sum_l \tfrac{\nabla f_{\tilde{n_l}}^{(l)}}{\lVert \nabla f_{\tilde{n_l}}^{(l)} \rVert} 
 \simeq \sum_l \left( \tfrac{\nabla f_{n_l}^{(l)}}{\lVert \denomin \rVert} + \tfrac{\Z^{(l)}}{\sqrt{\tilde{n_l}}\lVert \denomin \rVert}\right) \,,
\end{equation}
\vspace{-1mm}
where the iterator $l$ goes through the different classes. We want to quantify the projection of the unit vector associated to a generic class $l$ along the corresponding FBG direction, for which Theorem~\ref{th:PCNGDdecreasing} shows that we can obtain a monotonous decrease for both classes.
The projection of these steps onto the FBG is (see Sec.~\ref{app:proj} for the derivation)
\vspace{-2mm}
\begin{equation}\label{eq:proj}\small
 \left(\frac{\nabla f_{\tilde{n_l}}^{(l)}}{\lVert \nabla f_{\tilde{n_l}}^{(l)} \rVert} \right)\cdot \frac{\nabla f_{n_l}^{(l)}}{\lVert \nabla f_{n_l}^{(l)} \rVert} =1 - \frac{  \lVert \Z^{(l)}\rVert^2 (\sin(\theta)^2)}{2\tilde{n_l}\lVert \nabla f_{n_l}^{(l)} \rVert^2} + o\left(\frac{1}{\tilde{n_l}}\right)\,,
\end{equation}
where $\theta$ indicates the angle between $\Z^{(l)}$ and $\nabla f_{n_l}^{(l)}$.
Eq.~\eqref{eq:proj} shows that the larger the number of examples of class $l$ in the mini-batch, the closer the steps are to the PCNGD direction. 
A direct consequence is that, despite the per-class normalization, the two classes do not have the same signal towards the optimal direction: the signal of the minority class is suppressed.

The signal related to each class is attenuated by $\tfrac{  \lVert \Z^{(l)}\rVert^2 (1-\cos(\theta)^2)}{2\tilde{n_l}\lVert \nabla f_{n_l}^{(l)} \rVert^2}$. At the beginning of learning, with random initial conditions, we can expect $\lVert \nabla f_{\n_0}^{(0)} \rVert\approx\lVert \nabla f_{\n_1}^{(1)} \rVert$, and we can expect that on average the noise fluctuations have a similar projection onto $\lVert \nabla f_{n_l}^{(l)} \rVert$. Consequently, the attenuation of the minority with respect to the majority signal is proportional to the imbalance ratio, $\frac{\n_0}{\n_1}$.
{As long as the gradient of the majority class remains large, the minority class updates point far from the direction that would allow the minority class to be optimized.}
Once the per-class gradient of the majority class converged, by Eq.~\eqref{eq:approx}, we have
$\tfrac{\nabla f_{\tilde{n_l}}^{(l)}}{\lVert \nabla f_{\tilde{n_l}}^{(l)} \rVert} \simeq \tfrac{ \Z^{(l)}}{\lVert  \Z^{(l)} \rVert}$, and the signal of the minority class can become relevant.
In Secs.~\ref{SubSec:StocAlgs} and \ref{sec:empirical} we will confirm our theory by showing that, if we account for this extra attenuation, the MID disappears.

\vspace{-3mm}
\subsection{Convergence of PCNSGD}
\label{sec:pcnsgd-convergence}
\vspace{-2mm}

We also derive a new convergence analysis for PCNSGD under the common assumption of bounded gradient variance, \textit{i.e.} $\E \| \nabla f^{(l)}(\x_{t}) - g^{(l)}(\x_{t}) \|^2 \leq \sigma_l^2$ where we introduced the shorthand notation $g^{(l)}(\x_t) := \nabla_{\tilde n} f^{(l)}(\x_t)$ (with $l=0,1$) to denote the stochastic gradients, and  $\sigma_l > 0$ (the expectation is over the randomness of the algorithm, both in terms of the choice of $\x_0$ and the choice of the mini-batch). We note that for finite-sum objective functions, one can precisely characterize $\sigma_l$ as a function of the mini-batch size (using standard concentration arguments such as Bernstein's inequality), with $\sigma_l \to 0$ as the batch size approaches the full batch. 

\begin{theorembox}[informal]
Assume that each $f^{(l)}$ is $L_1$-Lipschitz and $L_2$-smooth and $\E \| \nabla f^{(l)}(\x_{t}) - g^{(l)}(\x_{t}) \|^2 \leq \sigma_l^2$ where $\sigma_l > 0$. If, for all iterations $t \in [0, T-1]$, with $T < \infty$, $\cos \alpha(\x_t) \neq -1$, then for all $\tilde T \in [0,T-1]$, the iterates of PCNSGD with step size $\eta_t = \bigO(\tfrac{1}{\sqrt{\tilde T}})$ satisfy
\begin{align*}
\min_{t \in [0,\tilde T-1]} \E \| \nabla f^{(l)}(\x_{t}) \| \leq  \tfrac{K}{\sqrt{\tilde T}} + K'\sigma_l.
\end{align*}
for some constants $K, \; K'$, independent from $\tilde T$.
\label{th:PCNSGDdecreasing}
\end{theorembox}
We prove theorem~\ref{th:PCNSGDdecreasing} in App.~\ref{app:convergence_PCNGD}.\newline
Theorem~\ref{th:PCNSGDdecreasing} shows that the gradient converges at the same rate as in the deterministic case, but only up to a ball of radius $\sigma_l$. {One way of reducing the size of this ball is to increase the batch size. Another way is to make sure that there is no class $l$ for which $\sigma_l$ is large.} 
{We will later see that one can re-balance the variance $\sigma_l$ of each class using a simple oversampling technique.}

Finally, in App.~\ref{app:StocAlg} we show how the monotonicity, that characterized PCNGD, is now not recovered unless the batch size is large [Eq.~\eqref{eq:pcnsgd-mono}]. This is consistent with the analysis of Sec.~\ref{sec:directional-noise}.

\vspace{-3mm}
\subsection{Balancing the directional noise}\label{SubSec:StocAlgs}
\vspace{-2mm}

As shown in Secs.~\ref{sec:directional-noise} and \ref{sec:pcnsgd-convergence}, with SGD, the directional noise induced by class imbalance implies that per-class normalization is not enough to  avoid a suppression of the minority-class learning, thus not suppressing the MID. 
To confirm our theory with experiments, we make use of this knowledge by showing that remedies that compensate for this directional noise induced by imbalance suppress the MID.

\vspace{-3mm}
\paragraph{Per-class normalization with Rescaling (PCNSGD+R)}
We want to enforce that both majority and minority mini-batch signals project by the same amount onto the FBGs.
This is done by rescaling the minority-class signal by the appropriate amount, calculated from Eq.~\eqref{eq:proj}. The rescaling factor (we label with 0 the majority and with 1 the minority class),
\vspace{-2mm}
\begin{equation}\label{eq:rescaling}\small
    \nu = \left( 1 - \tfrac{  \lVert \Z^{(0)}\rVert^2 \sin(\theta_0)^2}{2\tilde{n}_0\lVert \nabla f_{n_0}^{(0)} \rVert^2} \right) \bigg/ \left( 1 - \tfrac{  \lVert \Z^{(1)}\rVert^2 \sin(\theta_1)^2}{2\tilde{n}_1\lVert \nabla f_{n_1}^{(1)} \rVert^2} \right) \,,
\end{equation}
relies on the calculation of the FBGs. To alleviate this burden we only calculate them every 5 steps.
The slowness of this algorithm is not a problem, since it is not meant as an alternative to SGD, but rather to provide empirical evidence that the arguments in Sec.~\ref{sec:directional-noise} are correct.
Indeed, we show in Fig.~\ref{fig:PCNSGD-wrong} (green lines) that, after this further rescaling, the MID is prevented: the minority class recall is monotonically increasing since the beginning of learning, and the majority and minority classes have a similar evolution.
{In practice, as soon as the signals related to each class are \textit{balanced} (taking into account \textit{all} sources of imbalance, including directional noise), the MID disappears.}

\vspace{-3mm}
\paragraph{Oversampling (O)}
An alternative way to rebalance the directional noise in a more computationally efficient manner is to impose that all minibatches are composed of the same number of examples from each class. This is similar to minority class oversampling, with the difference that we enforce that every batch is \textit{exactly} split between the classes (see App.~\ref{app:algorithms} for more details). Contrary to what we saw at the end of Sec.~\ref{sec:directional-noise}, in this case, $\frac{\n_0}{\n_1}$ does not scale with the imbalance ratio, being $\n_0 = \n_1$.

\vspace{-3mm}
\paragraph{Per-class normalization with Oversampling (PCNSGD+O)}
The idea of normalizing per-class gradient contributions can in principle be coupled with other existing methods that are used to deal with class imbalance. Here, we check whether there is a gain in combining per-class normalization with oversampling. In fact, while oversampling equalizes the number of examples per class within each batch, PCN puts an additional constraint on the per-class gradients. For some datasets, this could be influential in the initial phases of learning, in the cases in which the gradients related to one class are larger than others. {In fact the hardness of different classes may vary (some classes are harder than others, which reflects in the per-class gradients), and this could cause an effect analogous to class imbalance. PCNSGD+O should reduce this difference.}

\vspace{-3mm}
\subsection{Experiments with Stochastic Gradients}\label{sec:empirical}
\vspace{-2mm}

\begin{figure}
    \centering
    \includegraphics[width=\columnwidth, trim=0 300 0 0]{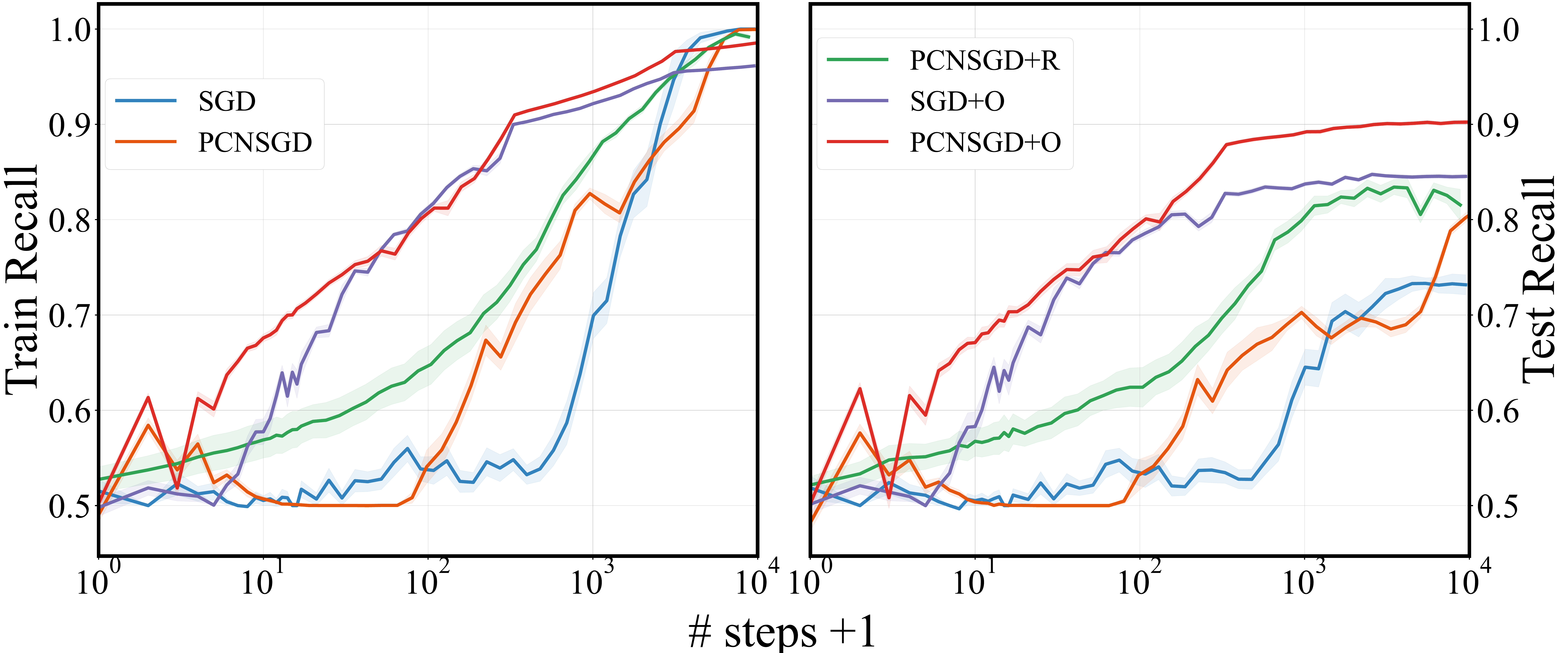}
    \caption{Comparison between different stochastic algorithms with model \cnn on the \bisevena dataset.
    }
    \label{fig:PCNSGD}
   \vspace{-3mm}
\end{figure}

In Fig.~\ref{fig:PCNSGD} we compare the described algorithms. As per our theory, while SGD and PCNSGD exhibit the MID, PCNSGD+R grows steadily since the beginning of learning.\\
PCNSGD+R is outperformed by SGD+O and PCNSGD+O, which is partly expected, since in PCNSGD+R we only calculate the FBG every 5 steps, and because PCNSGD+R acts on the signal of the minority class, but leaves the signal-to-noise ratio unaltered.\footnote{After the rescaling of the minority class through Eq.~\eqref{eq:proj}, all the per-class gradients project equally onto the respective full-batch direction. However, this rescaling also amplifies the minority-class signal in the direction orthogonal to the FBG.
}
From Fig.~\ref{fig:PCNSGD} it seems that PCNSGD+O outperforms SGD+O, since both the recall at short times and the final test recall are higher. This advantage of PCNSGD+O is even more visible with higher imbalance (Fig.~\ref{figure:60Stoc}). However, this is not systematic throughout all the experiments we performed (Tab.~\ref{tab:SGD} and Fig.~\ref{fig:stoc_group} in App.~\ref{app:more-exps}).
It is instead systematic that the MID is present with SGD dynamics, but it is always avoided when using oversampling. This can be seen from the value of $\ttau$ in Tab.~\ref{tab:SGD}, which is considerably smaller.
Extra runs are shown in App.~\ref{app:more-exps}.

\vspace{-2mm}
\section{Discussion}
\vspace{-2mm}

We presented a new analysis of the learning dynamics of gradient-based algorithms for imbalanced problems, focusing on the learning curves related to each class. Our results highlight the suboptimal behavior of GD, which becomes especially acute for high class imbalance. In particular, at the initial stages of learning, we observe the MID, where the minority classes are classified worse than random. This is particularly problematic when many short runs are needed, such as during hyperparameter tuning.
Only in late stages, at times that increase with the imbalance, are the minority classes learned, in agreement with observations by~\cite{ye:21}.

From our analysis, we see that the influence of the imbalance on the dynamics is indirect, and it fades over time. 
In fact, we find that the learning is not governed by the imbalance ratio itself, but rather by the ratio $C_t$ between per-class gradient norms. This highlights a limitation of methods which renormalize the loss (or the dynamics) in terms of the number of examples per class.
While at the beginning of learning $C_t$ is reasonably correlated with the imbalance, it is not so at later stages of learning.
This also highlights that it is not class imbalance \textit{per se} that causes suboptimal dynamics, but rather the disparity between per-class gradients. While class imbalance does affect this quantity, there are other sources, \textit{e.g.} related to the  difficulty of each class, which could have the same effect. It would therefore be interesting to investigate the relationship between different factors that give rise to a disparity among per-class gradients.

One of our main contributions is a convergence analysis of gradient-based algorithms under class imbalance, with a focus on the \textit{per-class} losses and their gradients.
Furthermore, we studied how imbalance differently affects GD and SGD, by analyzing their PCN variants.
We found fundamental differences between the deterministic and stochastic settings. While, in the former, class imbalance mostly affects the modulus of the per-class gradients, in the latter it also causes a directional noise, which is stronger with small batch sizes. Consequently, the approaches effective in dealing with these effects need to be different in the two cases. While in the full-batch case, PCN is enough to eliminate the MID, in the stochastic case it is no longer sufficient. Instead, procedures that also balance the effects of the directional noise, as \textit{e.g.} oversampling, allow us to avoid the MID.

Overall, PCN algorithms were successful in addressing the suboptimal MID behavior of gradient descent under class imbalance. We proved that, under certain conditions on learning rate and batch size, they exhibit a monotonous per-class loss decrease, which results in faster training in the initial stages of learning. 
While per-class normalization directly acts on the dynamical update rule, this is not the case for most of the typically used methods to counter the class imbalance. Therefore, it can (and should) be used in synergy with other methods. We showed this with PCNSGD+O.
{In this regard, we also emphasize that our analysis is rather general and not related to a specific loss or architecture. It is possible, therefore, to combine the proposed approaches with loss functions typically employed for long-tailed datasets (\textit{e.g.} \cite{lin2017focal, cao2019learning, leng2022polyloss}), or with more complex architectures. However, we note that larger architectures have their own potential pitfalls, as evidenced by ~\cite{sagawa2020investigation} who showed that, in the overparametrized regime, the performance of minority groups deteriorates (despite the performance improving overall). }

Finally, there are several extensions of interest regarding the analysis of our algorithm, including a specialized analysis to escape saddle points~\cite{jin2017escape, daneshmand2018escaping} or analyze the effect of momentum~\cite{nesterov2003introductory}.


\section{Acknowledgements}
We thank Emma Chollet Ramampiandra for feedback on the manuscript and valuable tips on how to make it more clear.
This work was supported by the Swiss National Foundation, SNF grant \# 196902.


\bibliography{icml2023}
\bibliographystyle{unsrtnat}

\newpage
\appendix

\counterwithin{figure}{section}
\onecolumn

\vbox{
  {\hrule height 2pt \vskip 0.15in \vskip -\parskip}
  \centering
  {\LARGE\bf Appendix\par}
  {\vskip 0.2in \vskip -\parskip \hrule height 0.5pt \vskip 0.09in}
}
\vspace{-5mm}

\newcommand\invisiblepart[1]{%
  \refstepcounter{part}%
  \addcontentsline{toc}{part}{\protect\numberline{\thepart}#1}%
}

\invisiblepart{Appendix}
\setcounter{tocdepth}{2}
\localtableofcontents


\makeatletter
\renewcommand{\thefigure}{S\@arabic\c@figure}
\makeatletter

\clearpage

\section{Additional Related Work}\label{Add_rel_works}

We now elaborate more on existing strategies that have been used to mitigate class imbalance. 
As mentioned in Sec.~\ref{sec:intro}, we identify three broad classes of methods to handle class imbalance.

First, at the data level, resampling is arguably the most known approach where one either under- or over-samples datapoints in order to achieve a class-balanced distribution~\cite{he:09,huang:16}. These two methods have obvious drawbacks: while undersampling discards training data, oversampling causes longer training time and might lead to overfitting~\cite{chawla2004special, johnson:19}. Various techniques use more advanced sampling techniques, such as rebalancing the dataset through synthetic data~\cite{chawla:02}, relying on a K-NN classifier~\cite{mani2003knn} or other distance metrics to select samples near the boundary between classes, or reassigning the labels favoring minority classes~\cite{chou:20}. 

A second class of approaches acts on the loss function. Arguably the simplest approach in this category is class reweighting (cost-sensitive training), which  assigns a larger weight in the loss function to examples from minority classes. This can be done by rescaling the loss related to each example by the frequency of its class~\cite{japkowicz:00}, or by making use of an auxiliary balanced validation dataset~\cite{ren:18}. However, in overparameterized regimes, the reweighting approach may be ineffective in improving the performance related to minority groups~\cite{sagawa2020investigation}. 
In addition, it has been observed that resampling often outperforms reweighting significantly (see \cite{an2020resampling} and references therein). \newline
Another strategy is to increase the margins of the loss, which was shown to be effective in reducing the imbalance problem~\cite{huang:16,dong:18,menon:20}. 
{Other recently successful techniques include tailored regularizers~\cite{alshammari:22} or adapting the loss to allow for supervised contrastive learning~\cite{zhu:22}.}

A final way to mitigate class imbalance is to rely on algorithmic solutions. This can be done by: acting on the initial conditions, given empirical evidence that pretraining can be beneficial~\cite{hendrycks:19}; acting on the loss margins, but doing this dynamically according to a learning schedule~\cite{cao:19}; perturbing the inputs of the majority class~\cite{ye:21};
by using an alternative momentum which properly suits imbalance~\cite{tang:20};
or by adapting the direction of the gradient steps in order to suppress the domination of the majority class~\cite{anand:93}.


As discussed in~\cite{johnson:19}, all these methods achieve different levels of performance depending on a multitude of conditions (classifier, performance metric, $\ldots$). Importantly, these methods do not have well-understood convergence guarantees (in contrast to the PCN algorithms studied in this paper).

\section{Notation}\label{app:notation}
We summarize here some of the notation that is employed in the paper: 
\begin{itemize}
\item $|\cdot|$ : Absolute value of a scalar. When referred to a set, it denotes instead its cardinality, \textit{i.e.} the number of elements that make up the set 
\item $\| \cdot \|$ : $L_2$ Norm
\item $\%$ : Modulo operator
\item $\x \cdot \y = \langle \x , \y \rangle$ : inner product between two vectors $\x, \y $
\item $\angle(\x,\y)$ : angle between two vectors $\x, \y $
\item $C_l = \{ i \mid y_i = l \}$  : subgroup of indices belonging to class $l$ 
\item $C_t = \frac{\| \nabla f^{(1)}(\x_{t}) \|}{\| \nabla f^{(0)}(\x_{t}) \|}$
\item $C_{\mu,l} = \min_{t \in [0, T-1]} 2 \mu^{(l)} (1 + \cos \alpha(\x_t))$
\item $\mathcal{D} = (\xi_i, y_i)_{i=1}^n$ : dataset
\item $\mathcal{D}_l = (\xi_i, y_i)_{i \in C_l}$ : Subgroup of $\mathcal{D}$ elements belonging to class $l$
\item $D_0^{(l)} = [f^{(l)}(\x_{0}) - f^{(l)}_*]$: in the mini-batch case the same quantity is computed as an expectation value, \textit{i.e.} $D_0^{(l)} = \E[f^{(l)}(\x_{0}) - f^{(l)}_*]$
\item $f(\x_t) \equiv \frac{1}{|\mathcal{D}|} \sum_{i \in \mathcal{D}} f_i(\x_t)$ : Average loss function calculated over all elements in the dataset.
\item $f^{(l)}(\x_t) = \frac{1}{|\mathcal{D}|} \sum_{ i \in C_l} f_i(\x_t)$ : contribution to $f(\x_t)$ from class $l$
\item  $f^{(l)}_* = \min_{\x \in \R^d} f^{(l)}(\x)$
\item $\denomin = \nabla f_{n_l}^{(l)} + \tfrac{1}{\sqrt{\tilde{n_l}}} \Z^{(l)} $
\item $g^{(l)}(\x_t) = \nabla_{\tilde n} f^{(l)}(\x_t)$: see $\nabla_{\tilde n} f^{(l)}(\x_t)$
\item $L$: total number of classes
\item $L_1$ : Lipschitz constant of the generic per-class loss $f^{(l)}$
\item $L_2$ : Smooth constant of the generic per-class loss $f^{(l)}$
\item $N_e$ : total number of simulation epochs
\item $n$ : dataset size, \textit{i.e.} the number of elements that makes up the dataset
\item $n_l$ : Number of examples in the dataset belonging to class $l$
\item $\tilde n = |\gamma_t|$ : batch size at step $t$; note that for full-batch algorithms (\textit{e.g.} GD) $|\gamma_t|=n$ 
\item $\tilde n_l$ : Number of examples in the batch belonging to class $l$
\item $\x_t$ : set of network parameters at time $t$
\item $y_i \in  [0, \dots, L - 1]$ : label ; by convention, label " $0$ " identifies the majority class of the dataset
\item $\Z^{(l)}$: zero-average multivariate Gaussian random variable whose distribution is fixed by the covariance matrix associated with the gradients of the $l^\mathrm{th}$ class
\item $\alpha(\x_t) = \angle(\nabla f^{(0)}(\x_{t}), \nabla f^{(1)}(\x_{t}))$ : angle between the 2 gradients per-class in binary classification problems (in multi-class problems it is necessary to introduce additional indices to identify the classes considered in the angle calculation)
\item $\beta_t = \cos(\alpha(\x_t))$
\item $\gamma_t$: batch selected at step $t$
\item $\{ \gamma_t \}_e$: set of batches defined for the epoch $e$
\item $\eta_t$ : learning rate
\item $\mu^{(l)}$ : Per-class gradient dominated constant
\item $\xi_i \in \mathbb{R}^d$ : input vector
\item $\rho_k = \frac{n_k}{n}$: fraction of elements in the dataset belonging to class $k$ 
\item $\sigma_l^2$: finite upper bound for $\E \| \nabla f^{(l)}(\x_{t}) - g^{(l)}(\x_{t}) \|^2$
\item $\omega_{min}$ is defined differently in the various theorems; in each of them the definition is reported before the beginning of the proof
\item $\PCMBG$ : is the gradient over a random batch of size $\tilde n$, for examples belonging to the class $l$
\item     $ \APCMBG = \tfrac{1}{\tilde n_l} \sum_{\substack{i \in C_l \\ i \in \gamma_t }} \nabla f_{i}(\x_t)$ : Average gradient computed on the elements in the batch belonging to class $l$. Note that $\APCMBG$ and $\PCMBG$ differ by only one factor in particular $\APCMBG = \frac{\tilde n}{\tilde n_l} \PCMBG$

\end{itemize}

\section{Convergence rate GD}
\label{app:convergence_GD}

In this section, we derive a worst-case convergence analysis of the GD iterates for the per-class loss. Similar analyses are typically performed on a loss that combines all classes. We refer the reader to~\cite{nesterov2003introductory} for convex functions, to~\cite{ghadimi:13} for non-convex (but smooth) functions, and to~\cite{karimi2016linear} for PL functions.

We start by stating a more formal version of Theorem~\ref{th:GDdecreasing}, followed by its proof. For the sake of clarity, we prove our results for the binary case, where the number of classes $L = 2$. 

In the following, we denote by $f^{(l)}_* = \min_{\x \in \R^d} f^{(l)}(\x)$ the global minimum loss for each class $l$ (assuming each loss $f^{(l)}$ is lower bounded). Note that we do not require the minimum itself for each class to be unique.

\paragraph{Remark}
The following proof focuses on analyzing convergence within a finite time interval $t \in [0, T-1] \; , \;\; T < \infty$. Specifically, we examine the convergence of $\| \nabla f^{(l)}(\mathbf{x}_{s}) \|^2$ within this time frame to a neighborhood of zero. It is important to note that analyzing the convergence of $\| \nabla f^{(l)}(\mathbf{x}_{s}) \|^2$ directly to zero, rather than to its vicinity, would necessitate an asymptotic study, \textit{i.e.}, $T \rightarrow \infty$. Such an extension falls outside the scope of this work.

\begin{mybox}{gray}
\begin{theorem}[Formal version of Theorem~\ref{th:GDdecreasing}]
Assume that each $f^{(l)}$ is $L_1$-Lipshitz and $L_2$-smooth and let $\alpha(\x_t) = \angle(\nabla f^{(l)}(\x_{t}), \nabla f^{(1-l)}(\x_{t}))$. If for all iterations $t \in [0, T-1]$, with $T < \infty$, $\| \nabla f^{(1-l)}(\x_{t}) \| \neq 0$ and $\cos(\alpha(\x_t)) > - \frac{1}{C_t}$, with $C_t := \frac{\| \nabla f^{(1-l)}(\x_{t}) \|}{\| \nabla f^{(l)}(\x_{t}) \|}$, then for all $\tilde T \in [0, T]$, the iterates of gradient descent with step size $\eta_t = \min \left( \frac{1 + \cos(\alpha(\x_t)) C_t}{2 (1 + C_t^2) L_2}, \frac{c}{\sqrt{\tilde T}} \right)$ where $c > 0$ satisfy,
\begin{equation*}
\min_{s \in [0,\tilde T-1]} \| \nabla f^{(l)}(\x_{s}) \|^2 \leq \frac{2 (1 + C_{\max}) L_2}{(\omega_{min}^{(l)})^2 \tilde T} D_0^{(l)} + \frac{1}{\omega_{min}^{(l)} c \sqrt{\tilde T}}  D_0^{(l)},
\end{equation*}
for each $l \in \{0,1\}$, where $D_0^{(l)} \equiv [f^{(l)}(\x_{0}) - f^{(l)}_*]$, $\omega_{min}^{(l)} \equiv \min_{t \in [0,T-1]} 1 + \cos(\alpha(\x_t)) C_t > 0$, and $C_{\max} \equiv \max_{t \in [0,T-1]} C_t^2$.
\end{theorem}
\end{mybox}
\begin{proof}

We only write the detail for the function $f^{(0)}$ since the proof for $f^{(1)}$ is identical.\\

Since each function $f^{(0)}$ is $L_2$-smooth, we have
\begin{align}
f^{(0)}(\x_{t+1}) &\leq f^{(0)}(\x_{t}) + \langle \nabla f^{(0)}(\x_{t}), \x_{t+1} - \x_t \rangle + \frac{L_2}{2} \sqnorm{\x_{t+1} - \x_t} \\
&= f^{(0)}(\x_{t}) - \eta_t \langle \nabla f^{(0)}(\x_{t}), \nabla f(\x_{t}) \rangle + \frac{L_2 \eta_t^2}{2} \sqnorm{\nabla f(\x_{t})} \nonumber \\
&\leq f^{(0)}(\x_{t}) - \eta_t \sqnorm{\nabla f^{(0)}(\x_{t})} - \eta_t \langle \nabla f^{(0)}(\x_{t}), \nabla f^{(1)}(\x_{t}) \rangle + L_2 \eta_t^2 \sqnorm{\nabla f^{(0)}(\x_{t})} + L_2 \eta_t^2 \sqnorm{\nabla f^{(1)}(\x_{t})} \nonumber \\
&= f^{(0)}(\x_{t}) - \eta_t (1 - L_2 \eta_t) \sqnorm{\nabla f^{(0)}(\x_{t})} - \eta_t \langle \nabla f^{(0)}(\x_{t}), \nabla f^{(1)}(\x_{t}) \rangle + L_2 \eta_t^2 \sqnorm{\nabla f^{(1)}(\x_{t})} \nonumber \\
&= f^{(0)}(\x_{t}) - \eta_t (1 - L_2 \eta_t) \sqnorm{\nabla f^{(0)}(\x_{t})} - \eta_t \cos(\alpha(\x_t)) \| \nabla f^{(0)}(\x_{t}) \| \| \nabla f^{(1)}(\x_{t}) \| + L_2 \eta_t^2 \sqnorm{\nabla f^{(1)}(\x_{t})}, \nonumber
\end{align}
where the inequality in the third line is simply due to $\| \x + \y \|_2^2 \leq 2 \| \x \|_2^2 + 2 \| \y \|_2^2$ for any $\x, \y \in \R^d$.

Let $C_t := \frac{\| \nabla f^{(1)}(\x_{t}) \|}{\| \nabla f^{(0)}(\x_{t}) \|}$. We get
\begin{align}
& f^{(0)}(\x_{t+1}) \leq f^{(0)}(\x_{t}) - \eta_t \left(1 + \cos(\alpha(\x_t)) C_t - L_2 \eta_t - L_2 \eta_t C_t^2\right) \sqnorm{\nabla f^{(0)}(\x_{t})} \nonumber \\
\implies & \eta_t \left(1 + \cos(\alpha(\x_t)) C_t - (1 + C_t^2) L_2 \eta_t \right) \sqnorm{\nabla f^{(0)}(\x_{t})} \leq f^{(0)}(\x_{t}) - f^{(0)}(\x_{t+1}).
\end{align}

At this point, we see that we need the following condition on the angle $\alpha(\x_t)$:
\begin{equation}
\cos(\alpha(\x_t)) > -\frac{1}{C_t}.
\end{equation}

Taking $\eta_t = \min \left( \frac{1 + \cos(\alpha(\x_t)) C_t}{2 (1 + C_t^2) L_2}, \frac{c}{\sqrt{\tilde T}} \right)$ (where $c > 0$), we have
\begin{equation*}
\eta_t (1 + \cos(\alpha(\x_t)) C_t) - \eta_t^2 (1 + C_t^2) L_2 \geq \frac{\eta_t}{2} (1 + \cos(\alpha(\x_t)) C_t),
\end{equation*}
therefore
\begin{equation}
\frac{\eta_t}{2} (1 + \cos(\alpha(\x_t)) C_t) \sqnorm{\nabla f^{(0)}(\x_{t})} \leq f^{(0)}(\x_{t}) - f^{(0)}(\x_{t+1}).
\end{equation}

Let $\omega_t := 1 + \cos(\alpha(\x_t)) C_t$, then
\begin{align}
\sqnorm{\nabla f^{(0)}(\x_{t})} &\leq \frac{2}{\omega_t \eta_t}  (f^{(0)}(\x_{t}) - f^{(0)}(\x_{t+1})) \nonumber \\
&\leq \left( \max_t \frac{1}{\omega_t} \right) \frac{2}{\eta_t}  (f^{(0)}(\x_{t}) - f^{(0)}(\x_{t+1})).
\end{align}

Let $\omega_{min}^{(0)} = \min_{t \in [0,T-1]} \omega_t$ and $C_{\max} := \max_{t \in [0,T-1]} C_t^2$. By summing from $t = 0$ to $\tilde T$,
\begin{align}
\min_{s \in [0,\tilde T-1]} \| \nabla f^{(0)}(\x_{s}) \|^2 &\leq \frac{1}{\tilde T} \sum_{t=0}^{\tilde T-1} \|\nabla f^{(0)}(\x_{t}) \|^2 \nonumber \\
&\leq \frac{1}{\omega_{min}^{(0)} \tilde T} \max \left( \frac{2 (1 + C_{\max}) L_2}{\omega_{min}^{(0)}}, \frac{\sqrt{\tilde T}}{c} \right) [f^{(0)}(\x_{0}) - f^{(0)}(\x_{\tilde T})] \nonumber \\
&\leq \frac{1}{\omega_{min}^{(0)} \tilde T} \left( \frac{2 (1 + C_{\max}) L_2}{\omega_{min}^{(0)}} + \frac{\sqrt{\tilde T}}{c} \right) [f^{(0)}(\x_{0}) - f^{(0)}_*] \nonumber \\
&\leq \frac{2 (1 + C_{\max}) L_2}{(\omega_{min}^{(0)})^2 \tilde T} [f^{(0)}(\x_{0}) - f^{(0)}_*] + \frac{1}{\omega_{min}^{(0)} c \sqrt{\tilde T}}  [f^{(0)}(\x_{0}) - f^{(0)}_*],
\label{eq:min}
\end{align}
where $f^{(0)}_*$ denotes the global minimum of $f^{(0)}(\x)$.  Note that  Eq.~\eqref{eq:min} is related to the imbalance, since both $c$ and $\omega_{min}^{(0)}$ depend on $C_t$, which at least at the beginning of the dynamics depends on the imbalance.

\end{proof}

We note that the condition required in the theorem, $\cos(\alpha(\x_t)) > -\frac{\| \nabla f^{(0)}(\x_{t}) \|}{\| \nabla f^{(1)}(\x_{t}) \|}$ is quite restrictive and might not be satisfied in practice. We discuss this in more detail next.

\iftrue

\paragraph{Tightness of the upper bound}

Consider a quadratic function with a constant diagonal Hessian where all eigenvalues are equal to $L$, and for which
\begin{align*}
f^{(0)}(\x_{t+1}) &= f^{(0)}(\x_{t}) + \langle \nabla f^{(0)}(\x_{t}), \x_{t+1} - \x_t \rangle + \frac{L}{2} \sqnorm{\x_{t+1} - \x_t} \nonumber \\
&= f^{(0)}(\x_{t}) - \eta_t \langle \nabla f^{(0)}(\x_{t}), \nabla f(\x_{t}) \rangle + \frac{L \eta_t^2}{2} \sqnorm{\nabla f(\x_{t})} \nonumber \\
&= f^{(0)}(\x_{t}) - \eta_t \sqnorm{\nabla f^{(0)}(\x_{t})} - \eta_t \langle \nabla f^{(0)}(\x_{t}), \nabla f^{(1)}(\x_{t}) \rangle + \frac{L \eta_t^2}{2} \sqnorm{\nabla f^{(0)}(\x_{t})} + \frac{L \eta_t^2}{2} \sqnorm{\nabla f^{(1)}(\x_{t})} \nonumber \\
& \quad + L \eta_t^2 \langle \nabla f^{(0)}(\x_{t}), \nabla f^{(1)}(\x_{t}) \rangle \nonumber \\
&= f^{(0)}(\x_{t}) - \eta_t \left(1 - \frac{L \eta_t}{2} \right) \sqnorm{\nabla f^{(0)}(\x_{t})} - \eta_t (1 - L \eta_t) \langle \nabla f^{(0)}(\x_{t}), \nabla f^{(1)}(\x_{t}) \rangle + \frac{L \eta_t^2}{2} \sqnorm{\nabla f^{(1)}(\x_{t})}.
\end{align*}

Let $C_t := \frac{\| \nabla f^{(1)}(\x_{t}) \|}{\| \nabla f^{(0)}(\x_{t}) \|}$. We get
\begin{align}
f^{(0)}(\x_{t+1}) &= f^{(0)}(\x_{t}) - \eta_t \left(1 - L \eta_t/2 - L \eta_t C_t^2/2 + (1 - L \eta_t) \cos(\alpha(\x_t)) C_t \right) \sqnorm{\nabla f^{(0)}(\x_{t})}.
\end{align}

We see that in order to decrease $f^{(0)}$, we require the following condition to hold:
\begin{equation*}
1 - (1 + C_t^2) L \eta_t/2 + (1 - L \eta_t) \cos(\alpha(\x_t)) C_t > 0.
\end{equation*}

Taking $\eta_t = \frac{c}{\sqrt{\tilde T}}$, we see that we require the following condition on the step-size:
\begin{equation*}
c \leq \frac{(1 + \cos(\alpha(\x_t)) C_t) \sqrt{\tilde T}}{(1 + C_t^2) L/2 + L \cos(\alpha(\x_t)) C_t},
\end{equation*}
which in turns require $1 + \cos(\alpha(\x_t)) C_t > 0$. We therefore see that this condition can not be avoided when considering worst-case guarantees.

\fi

\paragraph{Alternate step size}

We also derive a convergence result for the case where the step size does not depend on the total number of iterations $\tilde T$. This allows us to obtain a faster rate of convergence, but we still observe a severe restriction on the angle between the gradient of the two classes.

\begin{theorembox}
Assume that each $f^{(l)}$ is $L_1$-Lipschitz and $L_2$-smooth. If for all iterations $t \in [0, T-1]$, with $T < \infty$, 
\[
\omega = \min_{t \in [0, T-1]} \eta (1 + \beta_t C_t - L_2 \eta (1+ C_t^2) ) > 0,
\]
where $\beta_t = \cos(\alpha(\mathbf{x}_t))$, then, for any $\tilde{T} \in [0, T-1]$, there exists a choice of step size that, under \textbf{restricted conditions} on the angle $\alpha(\mathbf{x}_t)$, yields
\begin{equation}
\min_{t \in [0,\tilde{T}-1]} \| \nabla f^{(l)}(\mathbf{x}_{t}) \|^2 \leq \frac{f^{(l)}(\mathbf{x}_{0}) - f^{(l)}_*}{\omega \tilde{T}}.
\end{equation}

\label{th:GD_alternate_step_size}
\end{theorembox}
\begin{proof}

Since each function $f^{(0)}$ is $L_2$-smooth, we have
\begin{align}
f^{(0)}(\x_{t+1}) &\leq f^{(0)}(\x_{t}) + \langle \nabla f^{(0)}(\x_{t}), \x_{t+1} - \x_t \rangle + \frac{L_2}{2} \sqnorm{\x_{t+1} - \x_t} \\
&= f^{(0)}(\x_{t}) - \eta \langle \nabla f^{(0)}(\x_{t}), \nabla f(\x_{t}) \rangle + \frac{L_2 \eta^2}{2} \sqnorm{\nabla f(\x_{t})} \nonumber \\
&\leq f^{(0)}(\x_{t}) - \eta \sqnorm{\nabla f^{(0)}(\x_{t})} - \eta \langle \nabla f^{(0)}(\x_{t}), \nabla f^{(1)}(\x_{t}) \rangle + L_2 \eta^2 \sqnorm{\nabla f^{(0)}(\x_{t})} + L_2 \eta^2 \sqnorm{\nabla f^{(1)}(\x_{t})} \nonumber \\
&= f^{(0)}(\x_{t}) - \eta (1 - L_2 \eta) \sqnorm{\nabla f^{(0)}(\x_{t})} - \eta \langle \nabla f^{(0)}(\x_{t}), \nabla f^{(1)}(\x_{t}) \rangle + L_2 \eta^2 \sqnorm{\nabla f^{(1)}(\x_{t})} \nonumber \\
&= f^{(0)}(\x_{t}) - \eta (1 - L_2 \eta) \sqnorm{\nabla f^{(0)}(\x_{t})} - \eta \cos(\alpha(\x_t)) \| \nabla f^{(0)}(\x_{t}) \| \| \nabla f^{(1)}(\x_{t}) \| + L_2 \eta^2 \sqnorm{\nabla f^{(1)}(\x_{t})}. \nonumber
\end{align}

Let $\| \nabla f^{(1)}(\x_{t}) \| = C_t \| \nabla f^{(0)}(\x_{t}) \|$ for $C_t \geq 0$ and $\beta_t = \cos(\alpha(\x_t))$. Note that $C_t$ depends on time.

\paragraph{Case 1: $\beta_t < 0$}

\begin{align}
f^{(0)}(\x_{t+1}) &\leq  f^{(0)}(\x_{t}) - \eta (1 - L_2 \eta) \sqnorm{\nabla f^{(0)}(\x_{t})} + \eta |\beta_t| C_t \| \nabla f^{(0)}(\x_{t}) \|^2 + L_2 \eta^2 C_t^2 \sqnorm{\nabla f^{(0)}(\x_{t})} \nonumber \\
&=  f^{(0)}(\x_{t}) - \eta (1 - |\beta_t| C_t - L_2 \eta (1+ C_t^2) ) \sqnorm{\nabla f^{(0)}(\x_{t})}.
\end{align}

We need
\begin{align}
(1 - |\beta_t| C_t - L_2 \eta (1+C_t^2)) > 0 \implies
\eta < \frac{1 - |\beta_t| C_t}{L_2 (1+C_t^2)}.
\end{align}

We also need $C_t < \frac{1}{|\beta_t|}$ for the function to decrease.

\paragraph{Case 2: $\beta_t \geq 0$}

\begin{align}
f^{(0)}(\x_{t+1}) &\leq  f^{(0)}(\x_{t}) - \eta (1 - L_2 \eta) \sqnorm{\nabla f^{(0)}(\x_{t})} - \eta |\beta_t| C_t \| \nabla f^{(0)}(\x_{t}) \|^2 + L_2 \eta^2 C_t^2 \sqnorm{\nabla f^{(0)}(\x_{t})} \nonumber \\
&=  f^{(0)}(\x_{t}) - \eta (1 + |\beta_t| C_t - L_2 \eta (1+C_t^2)) \sqnorm{\nabla f^{(0)}(\x_{t})}.
\end{align}

We need
\begin{align}
(1 + |\beta_t| C_t - L_2 \eta (1+C_t^2)) > 0 \implies
\eta < \frac{1 + |\beta_t| C_t}{L_2 (1+C_t^2)}.
\end{align}

\paragraph{Function decrease}

In both cases, we have an equation of the form
\begin{align}
f^{(0)}(\x_{t+1}) &\leq f^{(0)}(\x_{t}) - \omega_t \sqnorm{\nabla f^{(0)}(\x_{t})},
\end{align}
where $\omega_t = \eta (1 \pm |\beta_t| C_t - L_2 \eta (1+ C_t^2) )$.\\

Let $\omega = \min_{t \in [0, T-1]} \omega_t$. We then sum up the above inequality:
\begin{equation}
\omega \sum_{t=0}^{\tilde T -1} \| \nabla f^{(0)}(\x_{t}) \|^2 \leq f^{(0)}(\x_{0}) - f^{(0)}(\x_{T+1}) \leq f^{(0)}(\x_{0}) - f^{(0)}_*\,,
\end{equation}
where $f^{(0)}_*$ is the global minimum loss of class 0.

Finally, we lower bound $\| \nabla f^{(0)}(\x_{t}) \|$ by its minimum,
\begin{equation}
\min_{t \in [0,\tilde T-1]} \| \nabla f^{(0)}(\x_{t}) \|^2 \leq \frac{f^{(0)}(\x_{0}) - f^{(0)}_*}{\omega \tilde T}
\end{equation}

We see that we are only guaranteed a function decrease if $\omega_t > 0$, \textit{i.e.} $\beta_t \geq 0$ or $\beta_t \leq 0 \text{ and } C_t < \frac{1}{|\beta_t|}$.

\end{proof}

\section{Convergence rate PCNGD}
\label{app:convergence_PCNGD}

This section contains the formal version of Theorem~\ref{th:PCNGDdecreasing}, with a detailed proof.

\begin{theorembox}[Formal version of Theorem~\ref{th:PCNGDdecreasing}]
Assume that each $f^{(l)}$ is $L_1$-Lipschitz and $L_2$-smooth. 
If for all iterations $t \in [0, T-1]$, with $T < \infty$,  $\cos \alpha(\x_t) \neq -1$  , then for all $\tilde T \in [0, T-1]$, the iterates of PCNGD, with step size $\eta_t = \frac{c}{\sqrt{\tilde T}}$, where $c > 0$, satisfy
\begin{align*}
\min_{s \in [0,\tilde T-1]} \| \nabla f^{(l)}(\x_{s}) \| \leq \frac{1}{\omega_{\min} \sqrt{\tilde T}} \left( \frac{D_0^{(l)}}{c} + 2 L_2 c \right),
\end{align*}
for each $l \in \{0,1\}$, where $D_0^{(l)} = [f^{(l)}(\x_{0}) - f^{(l)}_*]$ and $\omega_{\min} := \min_{t \in [0,T-1]} (1 + \cos \alpha(\x_t))$.
\end{theorembox}
\begin{proof}

We only write the detail for the function $f^{(0)}$ since the proof for $f^{(1)}$ is identical.\\

First, since $f^{(0)}$ is $L_2$-smooth, we have
\begin{align}
f^{(0)}(\x_{t+1}) &\leq f^{(0)}(\x_{t}) + \langle \nabla f^{(0)}(\x_{t}), \x_{t+1} - \x_t \rangle + \frac{L_2}{2} \sqnorm{\x_{t+1} - \x_t} \nonumber \\
&\leq f^{(0)}(\x_{t}) - \eta_t \| \nabla f^{(0)}(\x_{t}) \| -\eta_t \langle \nabla f^{(0)}(\x_{t}), \frac{\nabla f^{(1)}(\x_{t})}{\| \nabla f^{(1)}(\x_{t})\|} \rangle \nonumber \\
&\quad + L_2 \eta_t^2 \sqnorm{\frac{\nabla f^{(0)}(\x_{t})}{\| \nabla f^{(0)}(\x_{t}) \|}} + L_2 \eta_t^2 \sqnorm{\frac{\nabla f^{(1)}(\x_{t})}{\| \nabla f^{(1)}(\x_{t}) \|}} \nonumber \\
&= f^{(0)}(\x_{t}) - \eta_t \| \nabla f^{(0)}(\x_{t}) \| - \eta_t \| \nabla f^{(0)}(\x_{t}) \| \cos \alpha(\x_t) + 2 L_2 \eta_t^2 \nonumber \\
&= f^{(0)}(\x_{t}) - \eta_t (1 + \cos \alpha(\x_t)) \| \nabla f^{(0)}(\x_{t}) \| + 2 L_2 \eta_t^2.
\label{eq:pcngd-decrease}
\end{align}

Since, at least at the beginning of training, both $(1+\cos\alpha(\x_t))$ and $\| \nabla f^{(l)}(\x_{t}) \|$ are strictly positive, Eq.~\ref{eq:pcngd-decrease} implies that, for a small enough learning rate, the per-class loss at time $t+1$ is smaller than that at time $t$, so the MID is avoided. 

Let $\omega_t := (1 + \cos \alpha(\x_t))$ and $\omega_{\min} := \min_{t \in [0,T-1]} (1 + \cos \alpha(\x_t))$. By rearranging the terms in the equation above, we get
\begin{align}
\| \nabla f^{(0)}(\x_{t}) \| &\leq \frac{1}{\eta_t \omega_t} [f^{(0)}(\x_{t}) - f^{(0)}(\x_{t+1})] + \frac{2 L_2 \eta_t}{\omega_t} \nonumber \\
&\leq \frac{1}{\eta_t} \left( \max_{t \in [0,T-1]} \frac{1}{\omega_t} \right) [f^{(0)}(\x_{t}) - f^{(0)}(\x_{t+1})] + 2 L_2 \eta_t \left( \max_{t \in [0,T-1]} \frac{1}{\omega_t} \right) \nonumber \\
&\leq \frac{1}{\eta_t \omega_{\min}} [f^{(0)}(\x_{t}) - f^{(0)}(\x_{t+1})] + \frac{2 L_2 \eta_t}{\omega_{\min}}.
\end{align}

Taking the minimum over $t$, we get
\begin{align}
\min_{s \in [0,\tilde T-1]} \| \nabla f^{(0)}(\x_{s}) \| &\leq \frac{1}{\tilde T} \sum_{t=0}^{\tilde T-1} \|\nabla f^{(0)}(\x_{t}) \| \nonumber \\
&\leq \frac{1}{c \; \omega_{\min} \sqrt{\tilde T}} [f^{(0)}(\x_{0}) - f^{(0)}(\x_{\tilde T})] + \frac{2 L_2 c}{\omega_{\min} \sqrt{\tilde T}} \nonumber \\
&\leq \frac{1}{c \; \omega_{\min} \sqrt{\tilde T}} [f^{(0)}(\x_{0}) - f^{(0)}_*] + \frac{2 L_2 c}{\omega_{\min} \sqrt{\tilde T}} \nonumber \\
&\leq \frac{1}{\omega_{\min} \sqrt{\tilde T}} \left( \frac{1}{c} [f^{(0)}(\x_{0}) - f^{(0)}_*] + 2 L_2 c \right).
\end{align}

\end{proof}

\paragraph{Alternate choice of step size}

We also present a second version of Theorem~\ref{th:PCNGDdecreasing} with an alternate choice of step size.

\begin{theorembox}[Second formal version of Theorem~\ref{th:PCNGDdecreasing}]
Assume that each $f^{(l)}$ is $L_1$-Lipschitz and $L_2$-smooth. 
If for all iterations $t \in [0, T-1]$, with $T < \infty$,  $\cos \alpha(\x_t) \neq -1$, then, for all $\tilde T \in [0, T-1]$, the iterates of PCNGD with step size $\eta_t = \frac{c}{(1 + \cos \alpha(\x_t)) \sqrt{\tilde T}}$, where $c > 0$, satisfy
\begin{align*}
\min_{s \in [0,\tilde T-1]} \| \nabla f^{(l)}(\x_{s}) \| \leq \frac{1}{\sqrt{\tilde T}} \left( \frac{D_0^{(l)}}{c} + \frac{2 L_2 c}{\omega_{\min}} \right),
\end{align*}
for each $l \in \{0,1\}$, where $D_0^{(l)} = [f^{(l)}(\x_{0}) - f^{(l)}_*]$ and $\omega_{\min} := \min_{t \in [0,T-1]} (1 + \cos \alpha(\x_t))^2$.
\end{theorembox}
\begin{proof}

We only write the detail for the function $f^{(0)}$ since the proof for $f^{(1)}$ is identical.\\

First, since $f^{(0)}$ is $L_2$-smooth, we have
\begin{align}
f^{(0)}(\x_{t+1}) &\leq f^{(0)}(\x_{t}) + \langle \nabla f^{(0)}(\x_{t}), \x_{t+1} - \x_t \rangle + \frac{L_2}{2} \sqnorm{\x_{t+1} - \x_t} \nonumber \\
&\leq f^{(0)}(\x_{t}) - \eta_t \| \nabla f^{(0)}(\x_{t}) \| -\eta_t \langle \nabla f^{(0)}(\x_{t}), \frac{\nabla f^{(1)}(\x_{t})}{\| \nabla f^{(1)}(\x_{t})\|} \rangle \nonumber \\
&\quad + L_2 \eta_t^2 \sqnorm{\frac{\nabla f^{(0)}(\x_{t})}{\| \nabla f^{(0)}(\x_{t}) \|}} + L_2 \eta_t^2 \sqnorm{\frac{\nabla f^{(1)}(\x_{t})}{\| \nabla f^{(1)}(\x_{t}) \|}} \nonumber \\
&= f^{(0)}(\x_{t}) - \eta_t \| \nabla f^{(0)}(\x_{t}) \| - \eta_t \| \nabla f^{(0)}(\x_{t}) \| \cos \alpha(\x_t) + 2 L_2 \eta_t^2 \nonumber \\
&= f^{(0)}(\x_{t}) - \eta_t (1 + \cos \alpha(\x_t)) \| \nabla f^{(0)}(\x_{t}) \| + 2 L_2 \eta_t^2 \nonumber\\
&= f^{(0)}(\x_{t}) - \frac{c}{\sqrt{\tilde T}} \| \nabla f^{(0)}(\x_{t}) \| + 2 L_2 \frac{c^2}{\tilde T (1 + \cos \alpha(\x_t))^2}.
\end{align}

Let $\omega_{\min} := \min_{t \in [0,T-1]} (1 + \cos \alpha(\x_t))^2$. By rearranging the terms in the equation above, we get
\begin{align}
\| \nabla f^{(0)}(\x_{t}) \| &\leq \frac{\sqrt{\tilde T}}{c} [f^{(0)}(\x_{t}) - f^{(0)}(\x_{t+1})] + \max_{t \in [0,\tilde T-1]} \frac{2 L_2 c}{\sqrt{\tilde T} (1 + \cos \alpha(\x_t))^2} \nonumber \\
&\leq \frac{\sqrt{\tilde T}}{c} [f^{(0)}(\x_{t}) - f^{(0)}(\x_{t+1})] + \frac{2 L_2 c}{\sqrt{\tilde  T} \omega_{\min}}.
\end{align}

Taking the minimum over $t$, we get
\begin{align}
\min_{s \in [0, \tilde T-1]} \| \nabla f^{(0)}(\x_{s}) \| &\leq \frac{1}{\tilde T} \sum_{t=0}^{\tilde T-1} \|\nabla f^{(0)}(\x_{t}) \| \nonumber \\
&\leq \frac{1}{c \; \sqrt{\tilde T}} [f^{(0)}(\x_{0}) - f^{(0)}(\x_{\tilde T})] + \frac{2 L_2 c}{\sqrt{\tilde T} \omega_{\min}} \nonumber \\
&\leq \frac{1}{c \; \sqrt{\tilde T}} [f^{(0)}(\x_{0}) - f^{(0)}_*] + \frac{2 L_2 c}{\sqrt{\tilde T} \omega_{\min}} \nonumber \\
&\leq \frac{1}{\sqrt{\tilde T}} \left( \frac{1}{c} [f^{(0)}(\x_{0}) - f^{(0)}_*] + \frac{2 L_2 c}{\omega_{\min}} \right).
\end{align}

\end{proof}

\paragraph{Randomized iterate}

Next, we use the same randomization technique as in~\cite{ghadimi:13} where we choose an iterate of SGD at random according to a particular probability distribution.

\begin{theorembox}[R-PCNGD]
Consider a time interval such that $\cos(\alpha(\mathbf{x}_t)) \neq -1$ for all iterations $t \in [0, T-1]$with $T < \infty$. Given any $\tilde{T} \in [0, T-1]$, let the probability mass function $P_R(\cdot)$ be defined as
\begin{equation}
P_R(t) = \text{Prob} \{ R = t \} = \frac{\omega_t}{\sum_{t=0}^{\tilde{T}-1} \omega_t},
\end{equation}
where $\omega_t := (1 + \cos(\alpha(\mathbf{x}_t)))$.

Then, given the step size $\eta_t = \frac{c}{\sqrt{\tilde{T}}}$, we have for $l = \{0,1\}$,
\begin{equation}
\E \| \nabla f^{(l)}(\mathbf{x}_{R}) \| \leq \frac{1}{\sqrt{\tilde{T}} \bar{\omega}} \left[ c^{-1} (f^{(l)}(\mathbf{x}_{0}) - f^{(l)}(\mathbf{x}^*)) +  2 L_2 c \right],
\end{equation}
where $\bar{\omega} = \frac{1}{\tilde{T}} \sum_{t=0}^{\tilde{T}-1} \omega_t$.

\end{theorembox}

\begin{proof}

We only write the detail for the function $f^{(0)}$ since the proof for $f^{(1)}$ is identical.\\

First, since $f^{(0)}$ is $L_2$-smooth, we have
\begin{align}
f^{(0)}(\x_{t+1}) &\leq f^{(0)}(\x_{t}) + \langle \nabla f^{(0)}(\x_{t}), \x_{t+1} - \x_t \rangle + \frac{L_2}{2} \sqnorm{\x_{t+1} - \x_t} \nonumber \\
&\leq f^{(0)}(\x_{t}) - \eta_t \| \nabla f^{(0)}(\x_{t}) \| -\eta_t \langle \nabla f^{(0)}(\x_{t}), \frac{\nabla f^{(1)}(\x_{t})}{\| \nabla f^{(1)}(\x_{t})\|} \rangle \nonumber \\
&\qquad + L_2 \eta_t^2 \sqnorm{\frac{\nabla f^{(0)}(\x_{t})}{\| \nabla f^{(0)}(\x_{t}) \|}} + L_2 \eta_t^2 \sqnorm{\frac{\nabla f^{(1)}(\x_{t})}{\| \nabla f^{(1)}(\x_{t}) \|}} \nonumber \\
&= f^{(0)}(\x_{t}) - \eta_t \| \nabla f^{(0)}(\x_{t}) \| - \eta_t \| \nabla f^{(0)}(\x_{t}) \| \cos \alpha(\x_t) + 2 L_2 \eta_t^2 \nonumber \\
&= f^{(0)}(\x_{t}) - \eta_t (1 + \cos \alpha(\x_t)) \| \nabla f^{(0)}(\x_{t}) \| + 2 L_2 \eta_t^2.
\end{align}

Let $\omega_t := (1 + \cos \alpha(\x_t))$. By rearranging the terms in the equation above, we get
\begin{equation}
\omega_t \| \nabla f^{(0)}(\x_{t}) \| \leq \frac{1}{\eta_t} [f^{(0)}(\x_{t}) - f^{(0)}(\x_{t+1})] + 2 L_2 \eta_t
\end{equation}

By summing from $t = 0$ to $\tilde{T}-1$ and dividing by $\sum_{t=0}^{\tilde{T}-1} \omega_t$,
\begin{align}
\frac{\sum_{t=0}^{\tilde{T}-1} \omega_t \| \nabla f^{(0)}(\x_{t}) \|}{\sum_{t=0}^{\tilde{T}-1} \omega_t} &\leq  \frac{(\sum_{t=0}^{\tilde{T}-1} \eta_t)^{-1}}{\sum_{t=0}^{\tilde{T}-1} \omega_t} (f^{(0)}(\x_{0}) - f^{(0)}_*) +  \frac{2 L_2 \sum_{t=0}^{\tilde{T}-1} \eta_t}{\sum_{t=0}^{\tilde{T}-1} \omega_t} \nonumber \\
&= \frac{(\sum_{t=0}^{\tilde{T}-1} \eta_t)^{-1}}{\tilde{T} \bar{\omega}} (f^{(0)}(\x_{0}) - f^{(0)}_*) +  \frac{2 L_2 \sum_{t=0}^{\tilde{T}-1} \eta_t}{\tilde{T} \bar{\omega}},
\end{align}
where $\bar{\omega}$ is the average of the sequence $\{ \omega_t \}_{t=0}^{\tilde{T}-1}$.\\

Taking $\eta_t = \eta := \frac{c}{\sqrt{\tilde{T}}}$, we obtain
\begin{align}
\frac{\sum_{t=0}^{\tilde{T}-1} \omega_t \| \nabla f^{(0)}(\x_{t}) \|}{\sum_{t=0}^{\tilde{T}-1} \omega_t}
&\leq \frac{1}{c \bar{\omega} \sqrt{\tilde{T}}} (f^{(0)}(\x_{0}) - f^{(0)}_*) +  \frac{2 L_2 \eta}{\bar{\omega}} \nonumber \\
&\leq \frac{1}{c \bar{\omega} \sqrt{\tilde{T}}} (f^{(0)}(\x_{0}) - f^{(0)}_*) +  \frac{2 L_2 c}{\sqrt{\tilde{T}} \bar{\omega}}
\end{align}

Finally, observe that the LHS is an expectation $\E \| \nabla f^{(0)}(\x_{R}) \|$ for an appropriately chosen random variable $\x_R$ according to the distribution $P_R(\cdot)$, therefore
\begin{equation}
\E \| \nabla f^{(0)}(\x_{R}) \| \leq \frac{1}{\sqrt{\tilde{T}} \bar{\omega}} \left[ c^{-1} (f^{(0)}(\x_{0}) - f^{(0)}_*) +  2 L_2 c \right]
\end{equation}

\end{proof}

\paragraph{Convergence under Gradient dominance condition}

We prove convergence of PCNGD under a gradient-dominated condition which is a variant of the Polyak-Łojasiewicz (PL) condition that has been shown to hold for overparametrized neural networks~\cite{liu2022loss}. Instead of requiring this  condition to hold for $f$, we require it to hold for each class separately. Specifically, we say that a function satisfies the \textit{class-GD} inequality if the following holds for each class $l$,
\begin{equation}
\frac12 \| \nabla f^{(l)}(\x) \| \geq \mu^{(l)} | f^{(l)}(\x) - f^{(l)}(\x^*) |, \quad \forall \x \in \R^m,
\end{equation}
for some constant $\mu^{(l)} > 0$. For finite-sum objective functions, the class-GD inequality implies the gradient-dominated inequality.

We are now ready to state the main convergence result for smooth and class-GD functions.

\begin{theorembox}[Formal version of Theorem~\ref{thm:convergence_PL}]
Assume that each $f^{(l)}$ for $l \in \{0,1\}$ is $L_2$-smooth and $\mu$-class-GD.
If for all iterations $t \in [0, T-1]$, with $T < \infty$,$\cos \alpha(\x_t) \neq -1$, then for all $\tilde{T} \in [0, T-1]$,  the iterates of PCNGD with step size $\eta_t = \frac{2t+1}{C_{\mu,l} (t+1)^2}$, we have
\begin{align}
f^{(l)}(\x_{\tilde{T}}) - f^{(l)}_* \leq \frac{8 L_2}{C_{\mu,l}^2 \tilde{T}},
\end{align}
for each $l \in \{0,1\}$, where $C_{\mu,l} := \min_{t \in [0, T-1]} 2 \mu^{(l)} (1 + \cos \alpha(\x_t))$, with $\mu^{(l)} > 0$.
Furthermore, we obtain a linear rate of convergence up to a ball with a constant step size $\eta_t = \eta = \frac{c}{C_{\mu,l}}$ where $c \in (0,1)$:
\begin{align}
f^{(l)}(\x_{\tilde{T}}) - f^{(l)}_* \leq (1 - c)^{\tilde{T}-1} (f^{(l)}(\x_{0}) - f^{(l)}_*) + \frac{2 L_2 c}{C_{\mu,l}^2}.
\end{align}
\end{theorembox}
\begin{proof}
We again provide a proof for $f^{(0)}$ as the proof for $f^{(1)}$ is identical.

First, since $f^{(0)}$ is $L_2$-smooth, we have
\begin{align}
f^{(0)}(\x_{t+1}) &\leq f^{(0)}(\x_{t}) + \langle \nabla f^{(0)}(\x_{t}), \x_{t+1} - \x_t \rangle + \frac{L_2}{2} \sqnorm{\x_{t+1} - \x_t} \nonumber \\
&\leq f^{(0)}(\x_{t}) - \eta_t \| \nabla f^{(0)}(\x_{t}) \| -\eta_t \langle \nabla f^{(0)}(\x_{t}), \frac{\nabla f^{(1)}(\x_{t})}{\| \nabla f^{(1)}(\x_{t})\|} \rangle \nonumber \\
&\qquad + L_2 \eta_t^2 \sqnorm{\frac{\nabla f^{(0)}(\x_{t})}{\| \nabla f^{(0)}(\x_{t}) \|}} + L_2 \eta_t^2 \sqnorm{\frac{\nabla f^{(1)}(\x_{t})}{\| \nabla f^{(1)}(\x_{t}) \|}} \nonumber \\
&= f^{(0)}(\x_{t}) - \eta_t \| \nabla f^{(0)}(\x_{t}) \| - \eta_t \| \nabla f^{(0)}(\x_{t}) \| \cos \alpha(\x_t) + 2 L_2 \eta_t^2 \nonumber \\
&= f^{(0)}(\x_{t}) - \eta_t (1 + \cos \alpha(\x_t)) \| \nabla f^{(0)}(\x_{t}) \| + 2 L_2 \eta_t^2.
\end{align}

Using the gradient-dominated condition, and subtracting $f^{(0)}_*$ from both sides,
\begin{equation}
f^{(0)}(\x_{t+1}) - f^{(0)}_* \leq (1 - 2 \eta_t \mu^{(l)} (1 + \cos \alpha(\x_t))) (f^{(0)}(\x_{t}) - f^{(0)}_*) + 2 L_2 \eta_t^2.
\end{equation}

Note that we need $0<(1 - 2 \eta_t \mu^{(l)} (1 + \cos \alpha(\x_t))) < 1$ for all $t$, \textit{i.e.}
\begin{align}
\eta_t < \frac{1}{2 \mu^{(l)} (1 + \cos \alpha(\x_t))} < \frac{1}{C_{\mu,l}},
\end{align}
where $C_{\mu,l} := \min_{t \in [0, T-1]} 2 \mu^{(l)} (1 + \cos \alpha(\x_t))$.

\underline{Decreasing step size}
Choose $\eta_t = \frac{2t+1}{C_{\mu,l} (t+1)^2}$, then
\begin{equation}
f^{(0)}(\x_{t+1}) - f^{(0)}_* \leq \left( \frac{t^2}{(t+1)^2} \right) (f^{(0)}(\x_{t}) - f^{(0)}_*) + \frac{2 L_2 (2t+1)^2}{C_{\mu,l}^2 (t+1)^4}.
\end{equation}

Multiplying both sides by $(t+1)^2$, and letting $\delta_f(t) = t^2 (f^{(0)}(\x_t) - f^{(0)}_*)$, we obtain
\begin{align}
\delta_f(k+1) &\leq \delta_f(k) + \frac{2 L_2 (2t+1)^2}{C_{\mu,l}^2 (t+1)^2} \nonumber \\
&\leq \delta_f(k) + \frac{8 L_2}{C_{\mu,l}^2},
\end{align}
where the last inequality is due to $\frac{2t+1}{t+1} \leq 2$

Summing up this inequality from $t=0$ to $\tilde{T}-1$, we conclude
\begin{align}
& \delta_f(\tilde{T}) \leq \delta_f(0) + \frac{8 L_2 \tilde{T}}{C_{\mu,l}^2} \\
\implies & \tilde{T}^2 (f^{(0)}(\x_{\tilde{T}}) - f^{(0)}_*) \leq \frac{8 L_2 \tilde{T}}{C_{\mu,l}^2} \\
\implies & f^{(0)}(\x_{\tilde{T}}) - f^{(0)}_* \leq \frac{8 L_2}{C_{\mu,l}^2 \tilde{T}}.
\end{align}

\underline{Constant step size: $\eta_t = \eta > 0$}
Choose $\eta = \frac{c}{C_{\mu,l}}$ for $c \in (0,1)$, then
\begin{align}
f^{(0)}(\x_{t}) - f^{(0)}_* &\leq (1 - c)^{t-1} (f^{(0)}(\x_{0}) - f^{(0)}_*) + 2 L_2 \eta^2 \sum_{i=0}^{t-1} (1 - c)^i \nonumber \\
&\leq (1 - c)^{t-1} (f^{(0)}(\x_{0}) - f^{(0)}_*) + 2 L_2 \eta^2 \sum_{i=0}^\infty (1 - c)^i \nonumber \\
&\leq (1 - c)^{t-1} (f^{(0)}(\x_{0}) - f^{(0)}_*) + \frac{2 L_2 \eta^2}{c} \\
&= (1 - c)^{t-1} (f^{(0)}(\x_{0}) - f^{(0)}_*) + \frac{2 L_2 c}{C_{\mu,l}^2},
\end{align}
where we used the fact that the last term in the second line is a geometric series in the last equation.

\end{proof}

\subsection{Stochastic algorithms}\label{app:StocAlg}

We will now analyze the convergence property of PCNSGD whose update rule is given by
\begin{equation}
\x_{t+1} = \x_t - \eta_t \left( \frac{\nabla_{\tilde n} f^{(0)}(\x_t)}{\| \nabla_{\tilde n} f^{(0)}(\x_t) \|} + \frac{\nabla_{\tilde n} f^{(1)}(\x_t)}{\| \nabla_{\tilde n} f^{(1)}(\x_t) \|} \right)\,,
\end{equation}
where the subscript $\tilde{n}$ indicates gradients that are taken over batches of size $\tilde{n}$.
In the following, we will use the shorthand notation $g^{(l)}(\x_t) := \nabla_{\tilde n} f^{(l)}(\x_t)$ (with $l=0,1$) to denote the stochastic gradients.

\begin{mybox}{gray}
\begin{theorem}[Formal version of Theorem~\ref{th:PCNSGDdecreasing}]
Assume that each $f^{(l)}$ is $L_1$-Lipschitz and $L_2$-smooth and $\E \| \nabla f^{(l)}(\x_{t}) - g^{(l)}(\x_{t}) \|^2 \leq \sigma_l^2$ (where $\sigma_l > 0$).
If  for all iterations $t \in [0, T-1]$, with $T < \infty$, $\cos \alpha(\x_t) \neq -1$ , then for all $\tilde{T} \in [0, T-1]$, the iterates of PCNSGD with step size $\eta_t = \frac{c}{\sqrt{\tilde{T} }}$ (where $c > 0$) satisfy
\begin{align*}
\min_{s \in [0,\tilde{T} -1]} \E \| \nabla f^{(l)}(\x_{s}) \| \leq \frac{1}{\omega_{\min} \sqrt{\tilde{T} }} \left( \frac{D_0^{(l)}}{c} + 2 L_2 c \right) + \sigma_l \left( 1 + \frac{2}{\omega_{\min}} \right),
\end{align*}
for each $l \in \{0,1\}$, where $D_0^{(l)} = \E[f^{(l)}(\x_{0}) - f^{(l)}_*]$ and $\omega_{\min} = \min_{t \in [0,T-1]} (1 + \cos \alpha(\x_t)) >0$.
\end{theorem}
\end{mybox}

\begin{proof}

We only write the detail for the function $f^{(0)}$ since the proof for $f^{(1)}$ is identical. We also use the shorthand notation $\sigma := \sigma_0$. \\

First, since $f^{(0)}$ is $L_2$-smooth, we have
\begin{align}
f^{(0)}(\x_{t+1}) &\leq f^{(0)}(\x_{t}) + \langle \nabla f^{(0)}(\x_{t}), \x_{t+1} - \x_t \rangle + \frac{L_2}{2} \sqnorm{\x_{t+1} - \x_t} \nonumber \\
&\leq f^{(0)}(\x_{t}) - \eta_t \langle \nabla f^{(0)}(\x_{t}), \frac{g^{(0)}(\x_{t})}{\| g^{(0)}(\x_{t})\|} \rangle - \eta_t \langle \nabla f^{(0)}(\x_{t}), \frac{g^{(1)}(\x_{t})}{\| g^{(1)}(\x_{t})\|} \rangle \nonumber \\
&\quad + L_2 \eta_t^2 \sqnorm{\frac{g^{(0)}(\x_{t})}{\| g^{(0)}(\x_{t}) \|}} + L_2 \eta_t^2 \sqnorm{\frac{g^{(1)}(\x_{t})}{\| g^{(1)}(\x_{t}) \|}} \nonumber \\
&= f^{(0)}(\x_{t}) \underbrace{-\eta_t \langle \nabla f^{(0)}(\x_{t}), \frac{g^{(0)}(\x_{t})}{\| g^{(0)}(\x_{t})\|} \rangle}_{:= (A)} \underbrace{- \eta_t \langle \nabla f^{(0)}(\x_{t}), \frac{g^{(1)}(\x_{t})}{\| g^{(1)}(\x_{t})\|} \rangle}_{:= (B)} + 2 L_2 \eta_t^2.
\end{align}

By simple manipulations, term $(A)$ can be bounded as follows,
\begin{align}
(A) &= -\eta_t \langle \nabla f^{(0)}(\x_{t}) - g^{(0)}(\x_{t}), \frac{g^{(0)}(\x_{t})}{\| g^{(0)}(\x_{t})\|} \rangle - \eta_t \frac{\langle g^{(0)}(\x_{t}), g^{(0)}(\x_{t}) \rangle}{\| g^{(0)}(\x_{t}) \|} \nonumber \\
&= -\eta_t \langle \nabla f^{(0)}(\x_{t}) - g^{(0)}(\x_{t}), \frac{g^{(0)}(\x_{t})}{\| g^{(0)}(\x_{t})\|} \rangle - \eta_t \| g^{(0)}(\x_{t}) \| \nonumber \\
&\leq \eta_t \| \nabla f^{(0)}(\x_{t}) - g^{(0)}(\x_{t}) \| - \eta_t \| g^{(0)}(\x_{t}) \|,
\end{align}
where we used Cauchy-Schwarz in the last inequality.

Similarly for term $(B)$,
\begin{align}
(B) &= -\eta_t \langle \nabla f^{(0)}(\x_{t}), \frac{g^{(1)}(\x_{t})}{\| g^{(1)}(\x_{t})\|} \rangle \nonumber \\
&\leq \eta_t \| \nabla f^{(0)}(\x_{t}) - g^{(0)}(\x_{t}) \| - \eta_t \| g^{(0)}(\x_{t}) \| \cos \alpha(\x_t).
\end{align}

Let $ \omega_{\min} = \min_{t \in [0,T-1]} (1 + \cos \alpha(\x_t))$. Combining the last three equations and taking the expectation (over initial conditions and batches) yields
\begin{align}
\E[f^{(0)}(\x_{t+1}) - f^{(0)}(\x_{t})] &\leq 2 \eta_t \E \| \nabla f^{(0)}(\x_{t}) - g^{(0)}(\x_{t}) \| - \eta_t \E [ \| g^{(0)}(\x_{t}) \| (1 + \cos \alpha(\x_t)) ] + 2 L_2 \eta_t^2 \nonumber \\
&\leq 2 \eta_t \sqrt{\E \| \nabla f^{(0)}(\x_{t}) - g^{(0)}(\x_{t}) \|^2} - \eta_t \omega_{\min} \E [ \| g^{(0)}(\x_{t}) \|] + 2 L_2 \eta_t^2 \nonumber \\
&\leq 2 \eta_t \sigma - \eta_t \omega_{\min} \E [ \| g^{(0)}(\x_{t}) \|] + 2 L_2 \eta_t^2.
\label{eq:pcnsgd-mono}
\end{align}
Notice how, barring the first term in the last line, Eq.~\eqref{eq:pcnsgd-mono} resembles the structure of Eq.~\eqref{eq:pcngd-decrease}.
If $\sigma$ is finite the loss function is not monotonic on average. In order to have a per-class loss function which is monotonic on average and avoid the MID, $\sigma$ needs to be small, which corresponds to a large batch size.

Since $\cos(\alpha_t)\neq-1$, we can rearrange the terms in Eq.~\eqref{eq:pcnsgd-mono}, getting
\begin{equation}
\E \| g^{(0)}(\x_{t}) \| \leq \frac{1}{\omega_{\min}} \left( 2 \sigma + \frac{1}{\eta_t} \E[f^{(0)}(\x_{t}) - f^{(0)}(\x_{t+1})] + 2 L_2 \eta_t \right),
\end{equation}

therefore
\begin{align}
\E \| \nabla f^{(0)}(\x_{t}) \| &\leq \E \| g^{(0)}(\x_{t}) \| + \E \| \nabla f^{(0)}(\x_{t}) - g^{(0)}(\x_{t}) \| \nonumber \\
&\leq \frac{2 \sigma}{\omega_{\min}} + \frac{1}{\omega_{\min} \eta_t} \E[f^{(0)}(\x_{t}) - f^{(0)}(\x_{t+1})] + \frac{2 L_2 \eta_t}{\omega_{\min}} + \sqrt{\E \| \nabla f^{(0)}(\x_{t}) - g^{(0)}(\x_{t}) \|^2} \nonumber \\
&\leq \frac{2 \sigma}{\omega_{\min}} + \frac{1}{\omega_{\min} \eta_t} \E[f^{(0)}(\x_{t}) - f^{(0)}(\x_{t+1})] + \frac{2 L_2 \eta_t}{\omega_{\min}} + \sigma.
\end{align}

Taking the minimum over $t$, and using $\eta_t = \frac{c}{\sqrt{\tilde{T} }}$, we get

\begin{align}
\min_{s \in [0,\tilde{T} -1]} \E \| \nabla f^{(0)}(\x_{s}) \| &\leq \frac{1}{\tilde{T} } \sum_{t=0}^{\tilde{T} -1} \E \|\nabla f^{(0)}(\x_{t}) \| \nonumber \\
&\leq \frac{1}{c \; \omega_{\min} \sqrt{\tilde{T} }} \E[f^{(0)}(\x_{0}) - f^{(0)}(\x_{\tilde{T} })] + \frac{2 L_2 c}{\omega_{\min} \sqrt{\tilde{T} }} + \frac{2 \sigma}{\omega_{\min}} + \sigma \nonumber \\
&\leq \frac{1}{c \; \omega_{\min} \sqrt{\tilde{T} }} \E[f^{(0)}(\x_{0}) - f^{(0)}_*] + \frac{2 L_2 c}{\omega_{\min} \sqrt{\tilde{T} }} + \frac{2 \sigma}{\omega_{\min}} + \sigma \nonumber \\
&\leq \frac{1}{\omega_{\min} \sqrt{\tilde{T} }} \left( \frac{1}{c} \E[f^{(0)}(\x_{0}) - f^{(0)}_*] + 2 L_2 c \right) + \frac{2 \sigma}{\omega_{\min}} + \sigma,
\end{align}
where we used $f^{(0)}_* \leq f^{(0)}(\x_{\tilde{T} })$ by definition of $f^{(0)}_*$.

\end{proof}

\subsection{Projection of the PCNSGD steps onto the full-batch gradient}\label{app:proj}
Here, we elicit more explicitly how Eq.~\eqref{eq:proj} is obtained.
On the left hand side we write the projection of the per-batch gradients on the full batch. Then we use Eq.~\ref{eq:pcn} for the per-batch gradients, and keep the leading orders:
\begin{equation}\label{eq:proj-long}
\begin{aligned}
 &\left(\frac{\nabla f_{\tilde{n_l}}^{(l)}}{\lVert \nabla f_{\tilde{n_l}}^{(l)} \rVert} \right)\cdot \frac{\nabla f_{n_l}^{(l)}}{\lVert \nabla f_{n_l}^{(l)} \rVert}  = \Bigg(  \frac{\nabla f_{n_l}^{(l)}}{\lVert \nabla f_{n_l}^{(l)}\rVert} - \frac{ \nabla f_{n_l}^{(l)} \left(\nabla f_{n_l}^{(l)} \cdot \Z^{(l)}\right)}{\sqrt{\tilde{n_l}}\lVert \nabla f_{n_l}^{(l)} \rVert^3} -\frac{ \nabla f_{n_l}^{(l)}  \lVert \Z^{(l)}\rVert^2}{2\tilde{n_l}\lVert \nabla f_{n_l}^{(l)} \rVert^3} +\\
 &+ \frac{ 3 \nabla f_{n_l}^{(l)} \left(\nabla f_{n_l}^{(l)} \cdot \Z^{(l)}\right)^2}{2\tilde{n_l}\lVert \nabla f_{n_l}^{(l)} \rVert^5}+ 
 \frac{\Z^{(l)}}{\sqrt{\tilde{n_l}}\lVert \nabla f_{n_l}^{(l)}\rVert} - \frac{ \Z^{(l)} \left(\nabla f_{n_l}^{(l)} \cdot \Z^{(l)}\right)}{\tilde{n_l}\lVert \nabla f_{n_l}^{(l)} \rVert^3} 
 +o\left(\frac{1}{\tilde{n_l}}\right) \Bigg)\cdot \frac{\nabla f_{n_l}^{(l)}}{\lVert \nabla f_{n_l}^{(l)} \rVert} \\
 & = 1 - \frac{  \lVert \Z^{(l)}\rVert^2 (1-\cos(\theta)^2)}{2\tilde{n_l}\lVert \nabla f_{n_l}^{(l)} \rVert^2} + o\left(\frac{1}{\tilde{n_l}}\right)\,.
\end{aligned}
\end{equation}
Here, $\theta$ indicates the angle between $\Z^{(l)}$ and $\nabla f_{n_l}^{(l)}$.

\section{Multi-class analysis}\label{app:multiclass-gd}

In the following, we demonstrate how the analysis derived in previous sections can be adapted to multi-class problems. We only derive the proof for the deterministic case (full-batch) with one choice of step-size, but the analysis can be adapted similarly to other settings.

\paragraph{Gradient descent analysis}

The update is
\begin{align}
\x_{t+1} &= \x_t - \eta_t \nabla f(\x_t) \\
&= \x_t - \eta_t \sum_{i=0}^{L-1} \nabla f^{(i)}(\x_t).
\end{align}

\begin{mybox}{gray}
\begin{theorem}[Formal, multiclass, version of Theorem~\ref{th:GDdecreasing} - multi-class]\label{th:GDdecreasingMulti}
Assume that each $f^{(l)}$ is $L_1$-Lipshitz and $L_2$-smooth and let $\alpha^{(l)}(\x_t) = \angle(\nabla f^{(l)}(\x_{t}), \sum_{i\neq l} \nabla f^{(i)} (\x_{t}))$. If, for all iterations $t \in [0, T-1]$, with $T < \infty$, $\| \sum_{i\neq l} \nabla f^{(i)} (\x_{t})) \| \neq 0$ and $\cos(\alpha^{(l)}(\x_t)) > - \frac{1}{C_t^{(l)}}$ where $C_t^{(l)} := \frac{\| \sum_{i \neq l} \nabla f^{(i)}(\x_{t}) \|}{\| \nabla f^{(l)}(\x_{t}) \|}$, then  for all $\tilde{T} \in [0, T-1]$  the iterates of gradient descent with step size $\eta_t = \min \left( \frac{1 + \cos(\alpha^{(l)}(\x_t)) C_t^{(l)}}{2 \left(1 + \left(C_t^{(l)}\right)^2\right) L_2}, \frac{c}{\sqrt{\tilde{T}}} \right)$ where $c > 0$ satisfy
\begin{equation*}
\min_{s \in [0,\tilde{T}-1]} \| \nabla f^{(l)}(\x_{s}) \|^2 \leq \frac{2 (1 + C_{\max}^{(l)}) L_2}{(\omega_{min}^{(l)})^2 \tilde{T}} D_0^{(l)} + \frac{1}{\omega_{min}^{(l)} c \sqrt{\tilde{T}}}  D_0^{(l)},
\end{equation*}
for each $l \in \{0,1\}$, where $D_0^{(l)} = [f^{(l)}(\x_{0}) - f^{(l)}_*]$, $\omega_{min}^{(l)} = \min_{t \in [0,T-1]} 1 + \cos(\alpha^{(l)}(\x_t)) C_t^{(l)}$, and $C_{\max}^{(l)} = \max_{t \in [0,T-1]}  \left(C_t^{(l)}\right)^2$.
\end{theorem}
\end{mybox}
\begin{proof}

We only write the detail for the function $f^{(0)}$ since the proof for the generic $f^{(l)}$ is identical. We use the shorthand notation $\alpha := \alpha^{(0)}$, $C_t := C_t^{(0)}$.\\\\

Since each function $f^{(0)}$ is $L_2$-smooth, we have
\begin{align}
&f^{(0)}(\x_{t+1}) \leq f^{(0)}(\x_{t}) + \langle \nabla f^{(0)}(\x_{t}), \x_{t+1} - \x_t \rangle + \frac{L_2}{2} \sqnorm{\x_{t+1} - \x_t} \\
&= f^{(0)}(\x_{t}) - \eta_t \langle \nabla f^{(0)}(\x_{t}), \nabla f(\x_{t}) \rangle + \frac{L_2 \eta_t^2}{2} \sqnorm{\nabla f(\x_{t})} \nonumber \\
&\leq f^{(0)}(\x_{t}) - \eta_t \sqnorm{\nabla f^{(0)}(\x_{t})} - \eta_t \langle \nabla f^{(0)}(\x_{t}), \sum_{i\neq 0} \nabla f^{(i)}(\x_t) \rangle + L_2 \eta_t^2 \sqnorm{\nabla f^{(0)}(\x_{t})} + L_2 \eta_t^2 \sqnorm{\sum_{i\neq 0} \nabla f^{(i)}(\x_t)} \nonumber \\
&= f^{(0)}(\x_{t}) - \eta_t (1 - L_2 \eta_t) \sqnorm{\nabla f^{(0)}(\x_{t})} - \eta_t \langle \nabla f^{(0)}(\x_{t}), \sum_{i\neq 0} \nabla f^{(i)}(\x_t) \rangle + L_2 \eta_t^2 \sqnorm{\sum_{i\neq 0} \nabla f^{(i)}(\x_t)} \nonumber \\
&= f^{(0)}(\x_{t}) - \eta_t (1 - L_2 \eta_t) \sqnorm{\nabla f^{(0)}(\x_{t})} - \eta_t \cos(\alpha(\x_t)) \| \nabla f^{(0)}(\x_{t}) \| \| \sum_{i\neq 0} \nabla f^{(i)}(\x_t) \| + L_2 \eta_t^2 \sqnorm{\sum_{i\neq 0} \nabla f^{(i)}(\x_t)}, \nonumber
\end{align}
where the inequality in the third line is simply due to $\| \x + \y \|_2^2 \leq 2 \| \x \|_2^2 + 2 \| \y \|_2^2$ for any $\x, \y \in \R^d$.

Let $C_t := \frac{\| \sum_{i\neq 0} \nabla f^{(i)}(\x_t) \|}{\| \nabla f^{(0)}(\x_{t}) \|}$. Note that, in the presence of only two classes, this definition reduces to the same as the one used in other sections. We get
\begin{align}
& f^{(0)}(\x_{t+1}) \leq f^{(0)}(\x_{t}) - \eta_t \left(1 + \cos(\alpha(\x_t)) C_t - L_2 \eta_t - L_2 \eta_t C_t^2\right) \sqnorm{\nabla f^{(0)}(\x_{t})} \nonumber \\
\implies & \eta_t (1 + \cos(\alpha(\x_t)) C_t - (1 + C_t^2) L_2 \eta_t) \sqnorm{\nabla f^{(0)}(\x_{t})} \leq f^{(0)}(\x_{t}) - f^{(0)}(\x_{t+1}).
\end{align}

At this point, we see that, in order to have a monotonic decrease of the loss related to class 0, we need the following condition on the angle $\alpha(\x_t)$:
\begin{equation}\label{eq:multi_cos_cond_GD}
\cos(\alpha(\x_t)) > -\frac{1}{C_t}\,.
\end{equation}

Taking $\eta_t = \min \left( \frac{1 + \cos(\alpha(\x_t)) C_t}{2 (1 + C_t^2) L_2}, \frac{c}{\sqrt{\tilde{T}}} \right)$, we have
\begin{equation*}
\eta_t (1 + \cos(\alpha(\x_t)) C_t) - \eta_t^2 (1 + C_t^2) L_2 \geq \frac{\eta_t}{2} (1 + \cos(\alpha(\x_t)) C_t),
\end{equation*}
therefore
\begin{equation}
\frac{\eta_t}{2} (1 + \cos(\alpha(\x_t)) C_t) \sqnorm{\nabla f^{(0)}(\x_{t})} \leq f^{(0)}(\x_{t}) - f^{(0)}(\x_{t+1}).
\end{equation}

Let $\omega_t := 1 + \cos(\alpha(\x_t)) C_t$, then
\begin{equation}
\sqnorm{\nabla f^{(0)}(\x_{t})} \leq \frac{2}{\omega_t \eta_t}  (f^{(0)}(\x_{t}) - f^{(0)}(\x_{t+1})).
\end{equation}

Let $\omega_{min}^{(0)} = \min_{t \in [0,T-1]} \omega_t$ and $C_{\max} := \max_{t \in [0,T-1]} C_t^2$. By summing from $t = 0$ to $\tilde{T}-1$,
\begin{align}
\min_{s \in [0,\tilde{T}-1]} \| \nabla f^{(0)}(\x_{s}) \|^2 &\leq \frac{1}{\tilde{T}} \sum_{t=0}^{\tilde{T}-1} \|\nabla f^{(0)}(\x_{t}) \|^2 \nonumber \\
&\leq \frac{1}{\omega_{min}^{(0)} \tilde{T}} \max \left( \frac{2 (1 + C_{\max}) L_2}{\omega_{min}^{(0)}}, \frac{\sqrt{\tilde{T}}}{c} \right) [f^{(0)}(\x_{0}) - f^{(0)}(\x_{\tilde{T}})] \nonumber \\
&\leq \frac{1}{\omega_{min}^{(0)} \tilde{T}} \left( \frac{2 (1 + C_{\max}) L_2}{\omega_{min}^{(0)}} + \frac{\sqrt{\tilde{T}}}{c} \right) [f^{(0)}(\x_{0}) - f^{(0)}_*] \nonumber \\
&\leq \frac{2 (1 + C_{\max}) L_2}{(\omega_{min}^{(0)})^2 \tilde{T}} [f^{(0)}(\x_{0}) - f^{(0)}_*] + \frac{1}{\omega_{min}^{(0)} c \sqrt{\tilde{T}}}  [f^{(0)}(\x_{0}) - f^{(0)}_*],
\label{eq:min_multi}
\end{align}
where $f^{(0)}_*$ denotes the global minimum of $f^{(0)}(\x)$.  Note that  Eq.~\eqref{eq:min_multi} is related to the imbalance, since both $c$ and $\omega_{min}^{(0)}$ depend on $C_t$, which at least at the beginning of the dynamics depends on the imbalance.

\end{proof}

We now want to get an intuition on the meaning of Th.~\ref{th:GDdecreasingMulti} and how condition~\eqref{eq:multi_cos_cond_GD} varies with imbalance and number of classes.
Since the numerator of $C_t$ is the norm of a sum of vectors, the value of $C_t$ depends on the mutual angles between the per-class gradients. This these are \textit{a priori} not known, we will use the worst-case scenario to elucidate the role of condition~\eqref{eq:multi_cos_cond_GD}.
The worse-case scenario for the optimization of class 0 is when all the $(L-1)$ gradient vectors  $\left\{ \nabla f^{(i)}(\x_{t}) \right\}_{i \neq 0}$ are collinear and pointing in the same direction 
(and, in the case of imbalance, class 0 is the minority class). 
Using the extensivity of the gradient,  we can write $\| \nabla f^{(i)}(\x_{t}) \| \sim n_i M_i$ where $n_i$ indicates the number of elements inside the class $i$, and $M_i$ the typical gradient norm (we are assuming that fluctuations have a finite variance).
In the case of a balanced dataset, $n_i = \hat{n} \; \forall i$. If all classes are equivalent, we can write $\| \nabla f^{(i)}(\x_{t}) \| \sim \hat{n} M$.
Under these assumptions, we will have, in the worst-case scenario for a balanced dataset
\begin{equation}\label{eq:bal_GD_range}
     C_t = (L-1)\,,
\end{equation}
where we see that condition~\eqref{eq:multi_cos_cond_GD} is increasingly restrictive with the number of classes.

Let us now relax the hypothesis of balanced dataset to consider imbalance. If, again, different classes have similar gradient norms, the difference between per-class gradients reduce to the imbalance between classes. For the worst-case scenario
\begin{equation}\label{eq:imbal_GD_range}
        C_t = \frac{\sum_{i \neq 0} n_i}{n_0}.
\end{equation}
Comparing the condition in case of imbalance with the balanced case (Eq. \eqref{eq:bal_GD_range}) we can see how, if class 0 is a minority class (which is the case we are interested in), the worst-case scenario become more restrictive; the imbalance condition implies, in fact, $\frac{\sum_{i \neq 0} n_i}{n_0} > (L-1)$.

\subparagraph{Example} 
We now show a simple example where the worst-case scenario value derived above represent a good estimation of the general case, \textit{i.e.} the limit where the value of $C_t$ is not influenced by the angles between the set of vectors $\left\{ \nabla f^{(i)}(\x_{t}) \right\}_{i \neq 0}$.
Let us consider a problem with $L$ classes with a big gap between the population of the majority one and all the others. 
We say there are $N$ examples in total, the majority class, $L$, has $n_L$ examples, 
and minority classes have $\epsilon \ll \frac{n_L}{L}$ examples.
In this case, there is no significant difference induced by the interference between the $(L-1)$ vectors and Eq.~\eqref{eq:imbal_GD_range} becomes
\begin{equation}
    C_t \sim \frac{N}{\epsilon} \gg 1\,,
\end{equation}
where we remind the reader that the least restrictive value for $C_t$ is $C_t=1$, and the bigger $C_t$, the more stringent condition~\eqref{eq:multi_cos_cond_GD} becomes.

\paragraph{PCNGD}
We now turn to the analysis of PCNGD in the multi-class scenario. We study the following variant for the multi-class case:
\begin{equation}
\x_{t+1} = \x_t - \eta_t \sum_{i=0}^{L-1} \frac{\nabla f^{(i)}(\x_t)}{\| \nabla f^{(i)}(\x_t) \|},
\end{equation}
which recovers the binary update for $L=2$:
\begin{equation}
\x_{t+1} = \x_t - \eta_t \left( \frac{\nabla f^{(0)}(\x_t)}{\| \nabla f^{(0)}(\x_t) \|} + \frac{\nabla f^{(1)}(\x_t)}{\| \nabla f^{(1)}(\x_t) \|} \right).
\end{equation}

\begin{theorembox}[Formal version of Theorem~\ref{th:PCNGDdecreasing}]
Assume that each $f^{(l)}$ is $L_1$-Lipschitz and $L_2$-smooth and let $\alpha^{(l)}(\x_t) = \angle(\nabla f^{(l)}(\x_{t}), \sum_{i\neq l} \frac{\nabla f^{(i)} (\x_{t}))}{\| \nabla f^{(i)} (\x_{t})) \|})$. 
If, for all iterations $t \in [0, T-1]$, with $T < \infty$, $\cos \alpha^{(l)}(\x_t) \neq -1$, then for all $\tilde{T} \in [0, T-1]$ the iterates of PCNGD with step size $\eta_t = \frac{c}{\sqrt{\tilde{T}}}$ where $c > 0$ satisfy
\begin{align*}
\min_{s \in [0,\tilde{T}-1]} \| \nabla f^{(l)}(\x_{s}) \| \leq \frac{1}{\omega^{(l)}_{\min} \sqrt{\tilde{T}}} \left( \frac{D_0^{(l)}}{c (K-1)} + L_2 c \right),
\end{align*}
for each $l \in \{0, \ldots L-1 \}$, where $D_0^{(l)} = [f^{(l)}(\x_{0}) - f^{(l)}_*]$ and $\omega^{(l)}_{\min} := \min_{t \in [0,T-1]} (1 + \cos \alpha^{(l)}(\x_t))$.
\end{theorembox}
\begin{proof}

We only write the detail for the function $f^{(0)}$ since the proof for any $f^{(l)}$ is identical. We use the shorthand notation $\alpha := \alpha^{(0)}$.\\

\blocco{
First, since $f^{(0)}$ is $L_2$-smooth, we have
\begin{align}
f^{(0)}(\x_{t+1}) &\leq f^{(0)}(\x_{t}) + \langle \nabla f^{(0)}(\x_{t}), \x_{t+1} - \x_t \rangle + \frac{L_2}{2} \sqnorm{\x_{t+1} - \x_t} \nonumber \\
&\leq f^{(0)}(\x_{t}) - \eta_t \| \nabla f^{(0)}(\x_{t}) \| -\eta_t \langle \nabla f^{(0)}(\x_{t}), \sum_{i \neq 0} \frac{\nabla f^{(i)}(\x_{t})}{\| \nabla f^{(i)}(\x_{t})\|} \rangle + \frac{L_2}{2} \eta_t^2 \sqnorm{\sum_{i=0}^{L-1} \frac{\nabla f^{(i)}(\x_{t})}{\| \nabla f^{(i)}(\x_{t}) \|}} \nonumber \\
&\leq f^{(0)}(\x_{t}) - \eta_t \| \nabla f^{(0)}(\x_{t}) \| - \eta_t (L-1) \| \nabla f^{(0)}(\x_{t}) \| \cos \alpha(\x_t) + \frac{L_2}{2} L^2 \eta_t^2 \nonumber \\
&= f^{(0)}(\x_{t}) - \eta_t (1 + (L-1) \cos \alpha(\x_t)) \| \nabla f^{(0)}(\x_{t}) \| + \frac{L_2}{2} L^2 \eta_t^2.
\label{eq:pcngd-decrease_multi}
\end{align}
}

First, since $f^{(0)}$ is $L_2$-smooth, we have
\begin{align}
f^{(0)}(\x_{t+1}) &\leq f^{(0)}(\x_{t}) + \langle \nabla f^{(0)}(\x_{t}), \x_{t+1} - \x_t \rangle + \frac{L_2}{2} \sqnorm{\x_{t+1} - \x_t} \nonumber \\
&\leq f^{(0)}(\x_{t}) - \eta_t \| \nabla f^{(0)}(\x_{t}) \| -\eta_t \langle \nabla f^{(0)}(\x_{t}), \sum_{i \neq 0} \frac{\nabla f^{(i)}(\x_{t})}{\| \nabla f^{(i)}(\x_{t})\|} \rangle + \frac{L_2}{2} \eta_t^2 \sqnorm{\sum_{i=0}^{K-1} \frac{\nabla f^{(i)}(\x_{t})}{\| \nabla f^{(i)}(\x_{t}) \|}} \nonumber \\
&= f^{(0)}(\x_{t}) - \eta_t \| \nabla f^{(0)}(\x_{t}) \| - \eta_t \left\| \sum_{i \neq 0} \frac{\nabla f^{(i)}(\x_{t})}{\| \nabla f^{(i)}(\x_{t})\|} \right\| \| \nabla f^{(0)}(\x_{t}) \| \cos \alpha(\x_t) + \frac{L_2}{2} \sqnorm{\sum_{i=0}^{L-1} \frac{\nabla f^{(i)}(\x_{t})}{\| \nabla f^{(i)}(\x_{t}) \|}} \eta_t^2 \nonumber \\
&= f^{(0)}(\x_{t}) - \eta_t \left( 1 + \left\| \sum_{i \neq 0} \frac{\nabla f^{(i)}(\x_{t})}{\| \nabla f^{(i)}(\x_{t})\|} \right\| \cos \alpha(\x_t) \right) \| \nabla f^{(0)}(\x_{t}) \| + \frac{L_2}{2} \sqnorm{\sum_{i=0}^{L-1} \frac{\nabla f^{(i)}(\x_{t})}{\| \nabla f^{(i)}(\x_{t}) \|}} \eta_t^2.
\label{eq:pcngd-decrease_multi}
\end{align}
Taking a sufficiently small $\eta_t$ the monotonicity condition for the per-class loss function (considering only first-order term in $\eta_t $) is:
\begin{equation}\label{eq:multi_cos_cond_PCNGD}
    \cos \alpha^{(l)}(\x_t) > -\frac{1}{\tilde C_t}
\end{equation}
with $\tilde C_t = \tilde C_t^{(0)} \equiv \left\| \sum_{i \neq 0} \frac{\nabla f^{(i)}(\x_{t})}{\| \nabla f^{(i)}(\x_{t})\|} \right\|$.
Similarly as done for GD we can, also in this case, starting from the set of per-class norms, get  $\tilde C_t$ in the worst-case scenario, \textit{i.e.}
\begin{equation}\label{eq:imbal_PCNGD_range}
    C_t = (L-1)
\end{equation}
Note that this is identical to the one derived for GD in the balanced case. Now, however the interval does not change with imbalance, \textit{i.e.} for PCNGD in the imbalanced case, we get the same condition for GD in the balanced case (see Fig.~\ref{fig:PCNGD-multi_scheme}).\newline
Finally, we note that although the condition on the angle does not depend on imbalance, is still does depend on the number of classes, in PCNGD as for GD.
\begin{figure}[t]
    \centering
    \includegraphics[width=0.95\columnwidth]{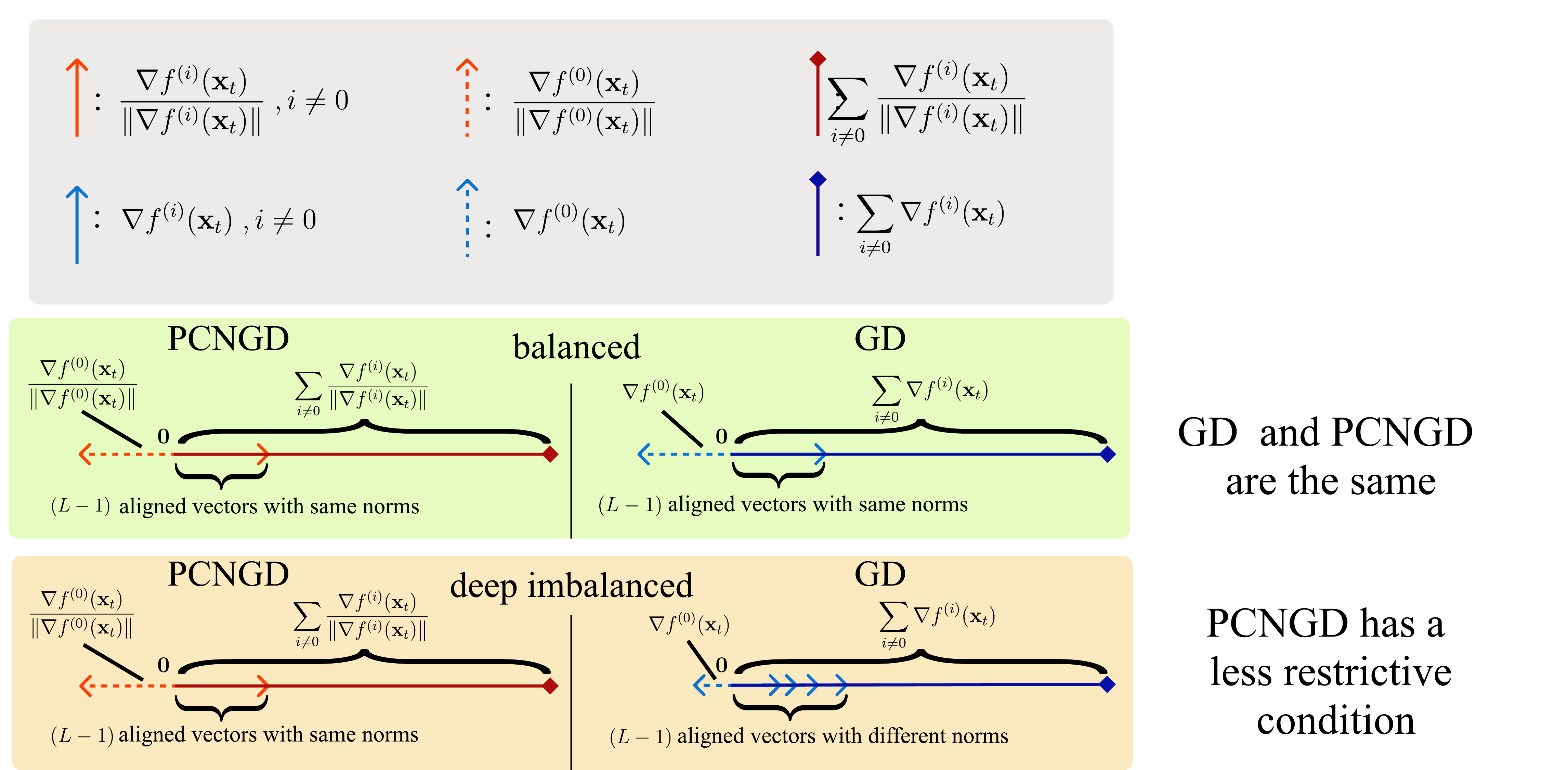}
    \caption{
    Comparison between GD (left schemes) and PCNGD (right schemes) for both balcanced (upper schemes) and imbalanced cases (bottom scheme). All diagrams represent the $L$ per-class gradient vectors (normalized vectors in the PCNGD cases); all vectors are centered on the same point, $0$. From Eq.~\eqref{eq:multi_cos_cond_PCNGD} and Eq.~\eqref{eq:multi_cos_cond_GD}, we know that the convergence condition depends on $C_t$ (and $\tilde C_t$ respectively ) which, in turns, depends on the norms of the gradient contribution coming from classes different from $0$ (red vectors in the diagram). In particular, fixed $\| \nabla f^{(0)}(\x_{t})\|$,  the bigger the norm of the red vector is (with respect to the norm of the purple vector), the more restrictive the condition becomes. Here we report a visual low-dimensional representation of the worst-case scenario.}
   \label{fig:PCNGD-multi_scheme}
   \vspace{-10pt}
\end{figure}

\blocco{
\textcolor{blue}{
Could do the following:
$$
f^{(0)}(\x_{t}) - \eta_t (1 - (K-1) | \cos \alpha(\x_t)) | \cdot \| \nabla f^{(0)}(\x_{t}) \| + \frac{L_2}{2} K^2 \eta_t^2.
$$
Alternative: simply keep the expression $\sum_i \frac{\y_i}{\| \y_i \|}$.
}

\textcolor{red}{
$$
\langle \x, \sum_{i=0}^{K-1} \frac{\y_i}{\| \y_i \|} \rangle = \| \x \| \| \sum_{i=0}^{K-1} \frac{\y_i}{\| \y_i \|} \| \cos(\x, \sum_{i=0}^{K-1} \frac{\y_i}{\| \y_i \|})
$$
$$
\x_1 = (1,0), \x_2=(0,1)
$$
Sum is
$$
(1,1)
$$
then norm is $\sqrt{2}$.
}

\textcolor{red}{
al: Problem here: we need $(1 + (K-1) \cos \alpha(\x_t)) > 0$! I think this is unavoidable, but we can still show it's better than GD. We might also want to look into clipping (Sra et al.) paper, need to discuss this in person.
}
}

Since, at least at the beginning of training, both $(1+\cos\alpha(\x_t))$ and $\| \nabla f^{(l)}(\x_{t}) \|$ are strictly positive, Eq.~\ref{eq:pcngd-decrease_multi} implies that, for a small enough learning rate, the per-class loss at time $t+1$ is smaller than that at time $t$, so the MID is avoided. 

Let $\omega_{\min} := \min_{t \in [0,T-1]} (1 + \cos \alpha(\x_t))$. By rearranging the terms in the equation above, we get
\begin{equation}
\| \nabla f^{(0)}(\x_{t}) \| \leq \frac{1}{\eta_t (L-1) \omega_{\min}} [f^{(0)}(\x_{t}) - f^{(0)}(\x_{t+1})] + \frac{K^2 L_2 \eta_t}{2 (L-1) \omega_{\min}}.
\end{equation}

Taking the minimum over $t$, we get
\begin{align}
\min_{s \in [0,\tilde{T}-1]} \| \nabla f^{(0)}(\x_{s}) \| &\leq \frac{1}{\tilde{T}} \sum_{t=0}^{\tilde{T}-1} \|\nabla f^{(0)}(\x_{t}) \| \nonumber \\
&\leq \frac{1}{c \; \omega_{\min} (L-1) \sqrt{\tilde{T}}} [f^{(0)}(\x_{0}) - f^{(0)}(\x_{\tilde{T}})] + \frac{K^2 L_2 c}{2 (L-1) \omega_{\min} \sqrt{\tilde{T}}} \nonumber \\
&\leq \frac{1}{c \; \omega_{\min} (L-1) \sqrt{\tilde{T}}} [f^{(0)}(\x_{0}) - f^{(0)}_*] + \frac{K^2 L_2 c}{2 (L-1) \omega_{\min} \sqrt{\tilde{T}}} \nonumber \\
&\leq \frac{1}{\omega_{\min} (L-1) \sqrt{\tilde{T}}} \left( \frac{1}{c} [f^{(0)}(\x_{0}) - f^{(0)}_*] + \frac{L_2}{2} L^2 c \right).
\end{align}

\end{proof}

\blocco{
\textcolor{red}{TODO list}
\begin{itemize}
    \item Figure out the equality in Eq.83, where the $(K-1)$ appears. We probably have to remove the $K-1$.
    \item Explain with graphical example why the $K-1$ term comes from, and that this is a worse case scenario, while the typical case is favorable for balanced GD (and for any PCNGD).
    \item Show that, with imbalance, the condition on multiclass GD gets worse than with balanced data (with balanced, half of the angles are ok in the $K\to\infty$ limit). Requires some work.
    \item Show that, with PCNGD, the condition on imbalanced cases is as with balanced cases in GD.
    \item Change discussion in main paper. Maybe also add something in the abstract and introduction (summary of results) about the intuition we gain from multiclass.
\end{itemize}
}

\section{Algorithms}\label{app:algorithms}
In this section we give a summary presentation (with pseudo-codes) of the various variants of (S)GD introduced in the study. 
\subsection{PCNGD}

\begin{algorithm}
\caption{PCNGD}\label{alg:PCNGD}
\begin{algorithmic}[1]
   \STATE Initialize $\x_0$
   \STATE Split $\mathcal{D}$ into subgroups $\{ \mathcal{D}_l \}$
   \FOR{epoch $e \in [1, \dots, N_e]$}
        \FOR{$l \in [0, \dots, L - 1]$}
            \STATE Calculate $\nabla f^{(l)}(\x_t)$ and $\lVert \nabla f^{(l)}(\x_t) \rVert $
        \ENDFOR
        \STATE $\x_{t+1} = \x_t -\eta_t \left( \sum_l \frac{\nabla f^{(l)}(\x_t)}{\lVert \nabla f^{(l)}(\x_t) \rVert} \right)$  \COMMENT{Update the set of parameters}
   \ENDFOR
\end{algorithmic}
\end{algorithm}

After initializing the network weights, $\x_0$ , (line 1 in Algorithm \ref{alg:PCNGD}) we split the examples of the dataset according to their class (line 2 in Algorithm \ref{alg:PCNGD}). For each epoch we calculate the gradient associated with the individual classes and the corresponding norm (line 4 in Algorithm \ref{alg:PCNGD}). Finally, we use the calculated quantities to perform the update rule (line 5 in Algorithm \ref{alg:PCNGD}).

\subsection{PCNSGD}

\begin{algorithm}
\caption{PCNSGD}\label{alg:PCNSGD}
\begin{algorithmic}[1]
\STATE Initialize $\x_0$
\STATE Split $\mathcal{D}$ into subgroups $\{ \mathcal{D}_l \}$ 
\FOR{epoch $e \in [1, \dots, N_e]$}
    \STATE Shuffle $\{ \mathcal{D}_l \}$
    \STATE Group $\{ \mathcal{D}_l \}$ into $\{ \gamma_t^{(l)} \}_e$
    \FOR{$i \in [1, \dots, N_b]$} 
        \FOR{$l \in [0, \dots, L - 1]$}
            \STATE Select  $\gamma_t^{(l)}$
            \STATE Calculate   $\PCMBG$ and $\lVert \PCMBG \rVert$ 
        \ENDFOR
        \STATE $\x_{t+1} = \x_t -\eta_t \left( \sum_l \frac{\PCMBG}{\lVert \PCMBG \rVert} \right)$ \COMMENT{Update the set of parameters}
    \ENDFOR
\ENDFOR
\end{algorithmic}
\end{algorithm}

After initializing the network weights, $\x_0$ , (line 1 in Algorithm \ref{alg:PCNSGD}) we split the examples of the dataset according to their class (line 2 in Algorithm \ref{alg:PCNSGD}). At the beginning of each epoch we shuffle the elements of each subgroup (line 4 in Algorithm \ref{alg:PCNSGD}) and group them into per-class batches (line 5 in Algorithm \ref{alg:PCNSGD}). Note that the per-class batch sizes are set by the imbalance ratio; consequently the number of per-class batches is the same $\forall l$, \textit{i.e.} $|\{ \gamma_t^{(l)} \}|=N_b^{(l)}= N_b$. \newline
We then begin to iterate over the batch index (line 6 in Algorithm \ref{alg:PCNSGD}); at each step we then select a per-class batch (line 8 in Algorithm \ref{alg:PCNSGD}) and calculate the gradient associated with it and its norm (line 9 in Algorithm \ref{alg:PCNSGD}). Finally, we use the calculated quantities to apply the update rule on the network weights (line 11 in Algorithm \ref{alg:PCNSGD}).

\subsection{PCNSGD+O}

\begin{algorithm}
\caption{PCNSGD+O}\label{alg:PCNSGD+O}
\begin{algorithmic}[1]
\STATE Initialize $\x_0$
\STATE Split $\mathcal{D}$ into subgroups $\{ \mathcal{D}_l \}$ 
\FOR{epoch $e \in [1, \dots, N_e]$}
    \STATE Shuffle $\{ \mathcal{D}_l \}$
    \STATE Group $\{ \mathcal{D}_l \}$ into $\{ \gamma_t^{(l)}$ $\}_e$   \COMMENT{" $0$ " is the label of the majority class} 
    \FOR{$i \in [1, \dots, N_b^{(0)}]$}
        \FOR{$l \in [0, \dots, L - 1]$}
            \IF{$i \% N_b^{(l)} = 0$ }
                \STATE Regroup $\mathcal{D}_l$ into per-class batches 
            \ENDIF
            \STATE Select $\gamma_t^{(l)}$
            \STATE Calculate  $\PCMBG$ and $\lVert \PCMBG \rVert$ 
            \ENDFOR
        \STATE $\x_{t+1} = \x_t -\eta_t \left( \sum_l \frac{\PCMBG}{\lVert \PCMBG \rVert} \right)$ \COMMENT{Update the set of parameters}
    \ENDFOR
\ENDFOR
\end{algorithmic}
\end{algorithm}

After initializing the network weights, $\x_0$ , (line 1 in Algorithm \ref{alg:PCNSGD+O}) we split the examples of the dataset according to their class (line 2 in Algorithm \ref{alg:PCNSGD+O}). At the beginning of each epoch we shuffle the elements of each subgroup (line 4 in Algorithm \ref{alg:PCNSGD+O}) and group them into per-class batches using the same per-class batch size, $|\gamma_t^{(l)}|= |\gamma_t|$  $\forall l$ (line 5 in Algorithm \ref{alg:PCNSGD+O}). We then begin to iterate over the majority class batch index (line 6 in Algorithm \ref{alg:PCNSGD+O}).
Note that since different classes have a different number of elements, $| \mathcal{D}_l|$, we will get a different number of batches for each of them: $|\{ \gamma_t^{(l)} \}| = N_b^{(l)}$, with $N_b^{(0)} = \max_l N_b^{(l)}$ (" $0$ " is the label of the majority class). \newline 
At each step we iterate along the classes.
For each of them we check that per-classes batches are still available to use as input; if the per-class batches associated with a class $l$ are finished we shuffle the elements into the subgroup $\mathcal{D}_l$ and define a new set of per-class-batches $\{ \gamma_l \}$ (line 9 in Algorithm \ref{alg:PCNSGD+O}) as done (for each class) at the beginning of the epoch. After this check we then select a per-class batch $\gamma_l$ (line 11 in Algorithm \ref{alg:PCNSGD+O}).
We calculate the gradient associated with it and its norm (line 12 in Algorithm \ref{alg:PCNSGD+O}). Finally, we use the calculated quantities to apply the update rule on the network weights (line 14 in Algorithm \ref{alg:PCNSGD+O}).

\subsection{SGD+O}

\begin{algorithm}
\caption{SGD+O}\label{alg:SGD+O}
\begin{algorithmic}[1]
\STATE Initialize $\x_0$
\STATE Split $\mathcal{D}$ into subgroups $\{ \mathcal{D}_l \}$ 
\FOR{epoch $e \in [1, \dots, N_e]$}
    \STATE Shuffle $\{ \mathcal{D}_l \}$
    \STATE Group $\{ \mathcal{D}_l \}$ into $\{ \gamma_t^{(l)}$ $\}_e$   \COMMENT{" $0$ " is the label of the majority class} 
    \FOR{$i \in [1, \dots, N_b^{(0)}]$}
        \FOR{$l \in [0, \dots, L - 1]$}
            \IF{$i \% N_b^{(l)} = 0$ }
                \STATE Regroup $\mathcal{D}_l$ into per-class batches 
            \ENDIF
            \STATE Select $\gamma_t^{(l)}$
            \STATE Calculate  $\PCMBG$ 
            \ENDFOR
        \STATE $\x_{t+1} = \x_t -\eta_t \left( \sum_l \PCMBG \right)$ \COMMENT{Update the set of parameters}
    \ENDFOR
\ENDFOR
\end{algorithmic}
\end{algorithm}

After initializing the network weights, $\x_0$ , (line 1 in Algorithm \ref{alg:SGD+O}) we split the examples of the dataset according to their class (line 2 in Algorithm \ref{alg:SGD+O}). At the beginning of each epoch we shuffle the elements of each subgroup (line 4 in Algorithm \ref{alg:SGD+O}) and group them into per-class batches using the same per-class batch size, $|\gamma_t^{(l)}|= |\gamma_t|$  $\forall l$ (line 5 in Algorithm \ref{alg:SGD+O}). We then begin to iterate over the majority class batch index (line 6 in Algorithm \ref{alg:SGD+O}).
Note that since different classes have a different number of elements, $| \mathcal{D}_l|$, we will get a different number of batches for each of them: $|\{ \gamma_t^{(l)} \}| = N_b^{(l)}$, with $N_b^{(0)} = \max_l N_b^{(l)}$ (" $0$ " is the label of the majority class). \newline 
At each step we iterate along the classes.
 For each of them we check that per-classes batches are still available to use as input; if the per-class batches associated with a class $l$ are finished we shuffle the elements into the subgroup $\mathcal{D}_l$ and define a new set of per-class-batches $\{ \gamma_l \}$ (line 9 in Algorithm \ref{alg:SGD+O}) as done (for each class) at the beginning of the epoch. After this check we then select a per-class batch $\gamma_l$ (line 11 in Algorithm \ref{alg:SGD+O}).
We calculate the gradient associated with it (line 12 in Algorithm \ref{alg:SGD+O}). Finally, we use the calculated quantities to apply the update rule on the network weights (line 14 in Algorithm \ref{alg:SGD+O}).

\subsection{PCNSGD+R}

\begin{algorithm}
\caption{PCNSGD+R}\label{alg:PCNSGD+R}
\begin{algorithmic}[1]
\STATE Initialize $\x_0$
\STATE Split $\mathcal{D}$ into subgroups $\{ \mathcal{D}_l \}$
\FOR{epoch $e \in [1, \dots, N_e]$}
    \STATE Shuffle $\{ \mathcal{D}_l \}$
    \STATE Group $\{ \mathcal{D}_l \}$ into $\{ \gamma_t^{(l)}$ $\}_e$.
    \FOR{$i \in [1, \dots, N_b]$ }
        \FOR{$l \in [0, \dots, L - 1]$}
            \STATE Select $\gamma_t^{(l)}$
            \STATE Calculate $\PCMBG$ and  $\lVert \PCMBG \rVert$
            \STATE Calculate $\nabla f^{(l)}(\x_t)$ and  $\lVert \nabla f^{(l)}(\x_t) \rVert$ 
            \STATE Compute $p_l = \left( \frac{\PCMBG}{\lVert \PCMBG \rVert} \right) \cdot \left( \frac{\nabla f^{(l)}(\x_t)}{\lVert \nabla f^{(l)}(\x_t) \rVert} \right)$
        \ENDFOR
        \STATE $\x_{t+1} = \x_t -\eta_t \left( \sum_l \frac{\PCMBG}{p_l \lVert \PCMBG \rVert } \right)$ \COMMENT{Update the set of parameters}
    \ENDFOR
\ENDFOR
\end{algorithmic}
\end{algorithm}

After initializing the network weights, $\x_0$ , (line 1 in Algorithm \ref{alg:PCNSGD+R}) we split the examples of the dataset according to their class (line 2 in Algorithm  \ref{alg:PCNSGD+R}). At the beginning of each epoch we shuffle the elements of each subgroup (line 4 in Algorithm  \ref{alg:PCNSGD+R}) and group them into per-class batches (line 5 in Algorithm \ref{alg:PCNSGD+R}). Note that Per-class batch sizes are set by the imbalance ratio; consequently the number of per-class batches $| \{ \gamma_t^{(l)} \}|=N_b^{(l)}= N_b$  is the same $\forall l$. \newline
We then begin to iterate over the batch index (line 6 in Algorithm \ref{alg:PCNSGD+R}); at each step, for each class, we then select a per-class batch (line 8 in Algorithm \ref{alg:PCNSGD+R}) and calculate the gradient associated with it and its norm (line 9 in Algorithm \ref{alg:PCNSGD+R}).
Next, we calculate the per-class gradient associated with the entire dataset and its norm (line 10 in Algorithm \ref{alg:PCNSGD+R}). We then calculate the projections of the normalized mini-batch gradients along the corresponding full-batch directions (line 11 in Algorithm \ref{alg:PCNSGD+R}). 
Finally, we use the calculated quantities to apply the update rule on the network weights (line 13 in Algorithm \ref{alg:PCNSGD+R}).

\newpage

\section{Models and data}\label{app:numerical}
The codes needed to reproduce the experiments presented are available in the following link: \href{https://github.com/EmanueleFrancazi/PCNGD-Algorithms}{https://github.com/EmanueleFrancazi/PCNGD-Algorithms} file, included in the repository, includes some details regarding the structure and operation of the scripts (how to set parameters, how seed initialization takes place, and so on).

\subsection{Network architecture}\label{Appendix:architecutes_info}

We provide here more extended information about the network architectures.
Before doing that, for the sake of clarity, we introduce some notation.
To define a convolutional layer (as well as a pooling one) it is necessary to specify some parameters:
\begin{itemize}
    \item \textbf{in channels}: Number of channels in the input originating from the preceding layer
    \item \textbf{out channels}: Number of channels produced by the convolution
    \item \textbf{stride}: Stride of the convolution
    \item \textbf{padding}: Padding added to all four sides of the input
    \item \textbf{kernel} :  Size of the convolving kernel
\end{itemize}
We used three architectures  for the simulations. In particular, we propose a simple network prototype (\cnn), and a two deeper ones as a prototype for more articulated models (\res and \vgg).

\begin{itemize}

\item \textit{Simple-CNN} (\cnn): As first architecture we chose a convolutional neural network, whose architecture was fixed as follows:
\begin{itemize}
    \item Input data 
    \item Convolutional layer: \textbf{in channels}=3, \textbf{out channels} =16, \textbf{kernel} =5, \textbf{stride}=1, \textbf{padding}=2
    \item activation function: ReLU
    \item Max Pooling: \textbf{kernel} =2, \textbf{stride}=2
    \item Convolutional layer: \textbf{in channels}=16, \textbf{out channels} =32, \textbf{kernel} =5, \textbf{stride}=1, \textbf{padding}=2
    \item activation function: hyperbolic tangent 
    \item Average pooling: \textbf{kernel} =2, \textbf{stride}=2
    \item Output: Fully connected linear layer
    \item Loss function: Cross Entropy
\end{itemize}

Before starting the network training, learning rate (\textit{LR}) and batch size (\textit{BS})\footnote{The batch size HO is not present in GD runs} values need to be fixed through hyperparameter tuning.

\item \textit{ResNet} (\res): {as a second architecture we adopted \textit{ResNet18} \cite{he2016deep}. The output of each convolutional layer is regularized by means of group normalization \cite{wu2018group}. The latter procedure, as opposed to the canonical batch normalization \cite{ioffe2015batch} is independent of the batch size.  This aspect on the one hand poses no performance limitation for particular batch size choices (as pointed out in  \cite{wu2018group}). 
On the other it does not place an explicit dependence on the batch size parameter and thus allows, for example, gradient accumulation to be used. Indeed, such a technique is useful if one wants to forward a batch whose overall size saturates the memory of the machine, or even if one wants to split the batch for reasons of efficiency. In our case, the separation concerned elements belonging to different classes, so as to efficiently collect the gradient at associated with each of them. Group normalization needs an additional parameter to fix the number of blocks within which to group features (\textit{GF}).}

\item \textit{VGG} (\vgg): as a third architecture we adopted \textit{VGG16} \cite{simonyan2014very}.
Here, each convolutional layer is followed by a dropout layer. 
before passing through the dropout (DO) layer, the output of each convolutional layer is regularized by means of group normalization \cite{wu2018group}. 
Compared to \cnn we thus have in \vgg two additional HPs to be fixed through the HP validation process (DO rate and GF).
\end{itemize}

The HP tuning involved the optimization of batch size and learning rate, by exhaustive grid search. The optimal hyperparameters were chosen based on the macro-averaged recall.

\subsection{Dataset and Hyper-parameters}\label{Appendix:Data&HP}
The runs in this work were performed using data from the CIFAR10 and CIFAR100 datasets~\cite{cifar}. Working only with CIFAR images allows us to attribute possible discrepancies between results to the imbalance choice, instead of the nature of the dataset.
In order to create different imbalanced classification settings, we preprocessed the dataset in several ways:
\begin{itemize}
    \item \textit{Simple classes} (\bisevena, \bisevenb\!): We selected two classes and discarded the others. We chose pairs of classes representing similar target items. The training dataset was composed of 5000 images belonging to the majority class and 714 to the minority one, so it had a 7:1 imbalance ratio. The test set was balanced, with 150 images per class. 
    In particular, 2 different pairs of classes were analyzed:
    \begin{itemize}
        \item \bisevena dataset: \texttt{truck} (majority class) and \texttt{car} (minority class),
        \item \bisevenb dataset: \texttt{horse} (majority class) and \texttt{deer} (minority class).
    \end{itemize}
    
    \item \textit{Cifar10 Super-classes (\bisixty\!)}: In order to have a dataset of larger size and imbalance, we created two super classes, \texttt{animals} and \texttt{vehicles}. The first comprises the \texttt{bird}, \texttt{cat}, \texttt{deer}, \texttt{dog}, \texttt{frog} and \texttt{horse} labels (with an equal distribution), and the second encompasses \texttt{airplane}, \texttt{automobile}, \texttt{ship} and \texttt{truck} (with an equal distribution). The \texttt{animals} class contained 30000 elements, and the \texttt{vehicles} class contained 500 elements (60:1 imbalance). The testing data consisted of 600 elements per superclass. We called this the \bisixty dataset.
    
    \item \textit{Cifar10 Multi-class (\multi\!)}: We use the 10 classes of the CIFAR10 dataset.
    The number of images associated with each class, $N_i$, is set by the relation:
    \begin{equation}
        N_i = N_{max} \left(\frac35\right)^i
    \end{equation}
    where $N_{max}=5000$ and $i$ is the label associated with the class, \textit{i.e.} a number between 0 and 9 that identifies the classes in alphabetical order (0: \texttt{airplane}, 1: \texttt{car}, 2: \texttt{bird}, 3: \texttt{cat}, 4: \texttt{deer}, 5: \texttt{dog}, 6: \texttt{frog}, 7: \texttt{horse}, 8: \texttt{ship}, 9: \texttt{truck}). We called this the \multi dataset. The value of the base ($\frac{3}{5}$) was chosen so as to have a $\rho=100$ imbalance ($N_0 \sim 100 N_9$), without excessively reducing the number of images in the least represented classes.
    
    \item \textit{Cifar100 Multi-class (\multib\!)}: 
    { Analogously to \multi  the number of images associated with each class, $N_i$, is set by the relation}:
    \begin{equation}
        N_i = N_{max} \left( 0.955 \right)^i
    \end{equation}
    with $N_{max}=500$ and $i \in [0\mathrel{{.}\,{.}} 99]$ in this case. We called this the \multib dataset. Also for \multib the imbalance ratio is $\rho=100$ ($N_0 \sim 100 N_{99}$). 
    
    
\end{itemize}
For each of the datasets described above, the validation set was constructed similarly to the test set (same criteria for the composition and same size) but using a different subset of images.

In choosing the combination of datasets and architectures for the experiments, we used simple (complex) models for simple (complex) datasets to limit the
number of experiments to a reasonable number.

\paragraph{A note on the meaning of the imbalance ratio}
We note that although the ratio $\rho=100$ between number of examples in the majority and minority classes is the same as in the \multi dataset and in the \multib, in the latter contiguous classes have a very similar number of examples. This renders the \multib dataset effectively less imbalanced. The learning problem remains hard because of the very small number of images per class, but this is unrelated to the dynamical effects induced by class imbalance that we are investigating in this paper.
In general, the same value of $\rho$ indicates a smaller imbalance when the number of classes is large.

\subsection{Execution times}\label{Appendix:ExTimes}
Simulations were run on different servers. The code we provide allows the choice of using either the CPU or the GPU; for the experiments presented, either one or the other was used, depending on availability. The models of the GPUs mounted on the servers used are:
\begin{itemize}
    \item GeForce- RTX 2080 Ti
    \item GP104GL Quadro P4000
    \item GM 200 GeForce GTX TitanX
\end{itemize}

Run times varied depending on the specific device model used and the level of occupancy (if multiple simulations are run in parallel on the same device). For all simulations run on GPUs, a single GPU was always used for each run. 
Other factors may cause the execution time to vary, such as the used batch size. A rigorous comparison of algorithm execution times should be made by setting same conditions for the various algorithms analyzed. This type of comparison is outside the scope of our analysis.\newline
Instead, we report below a rough estimate of the order of magnitude of the execution times of the simulations performed.  We try in this way to give an idea of the differences between the various algorithms. The values below refer to the average run time (expressed in seconds) normalized by the number of iterations ( $\frac{sec}{ iteration}$ units) .\newline
In the case of \cnn the times are lower. 
For the deterministic case GD and PCNGD proceed with similar speeds when run on the same devices. In particular we have an execution time that varies between $\sim 5 \frac{sec}{iteration}$ for \bisevenb to $\sim 30 \frac{sec}{iteration}$ for \multi . \newline
In the stochastic case we have a large gap between execution times of PCNSGD+R and the rest of the algorithms. PCNSGD+R runs with a speed of $\sim 4 \frac{sec}{iteration}$. The remaining algorithms, on the other hand, with a speed of $\sim 0.2 \frac{sec}{iteration}$.\newline
Execution time increases considerably by moving to a more complex architecture. For example, for the \vgg architecture we have an execution time of $ \sim 60 \frac{sec}{iteration}$ for the deterministic algorithms and $ \sim 1.5 \frac{sec}{iteration}$ for the stochastic ones.
For future use, the code records the execution times of the simulations on file.

\section{Additional experiments}
{This section reports the results of additional experiments performed on the data sets described in App.~\ref{Appendix:Data&HP}.} {The loss curves are reported together with the corresponding recall, since it is a clearly interpretable metric of interest for practical applications. Unless specified otherwise, we report the macro-averaged curves, which is the average of the per-class ones. 
All shown curves are averaged over between 4 and 30 random seeds (standard error bars are shown as a shading),
depending on how demanding the run was.
For the dynamics we used a further different random seed.}

\label{app:more-exps}
\subsection{Deterministic optimization}\label{app:det-exps}

Here, we show more plots on the comparison between GD and PCNGD. 

\begin{table*}[tb]
	\footnotesize
	\centering
	\tabcolsep 3pt
	\caption{Summary of experiments with (PCN)GD dynamics. We report test recall and characteristic time $\ttau$ (see main text in App. \ref{app:det-exps}), {as well as the imbalance ratio $\rho$, and the threshold recall $R^*$ used to calculate $\ttau$.}
 We highlight in bold the best values of recall and $\ttau$. Details on the runs are in Sec.~\ref{app:numerical}, and the full learning curves for these runs are shown in Sec.~\ref{app:more-exps}.
    }
	\vspace{-2pt}
	\begin{tabular}{l||c|c|c|c||c||c||c}
		\addlinespace
		\toprule
	Models+Dataset	& \multicolumn{ 2}{c|}{GD} & \multicolumn{ 2}{c||}{PCNGD}          &  \# iterations         & $\rho$   & $R^*$    \\
		\cmidrule{2-5}
		& \multicolumn{ 1}{c|}{Recall [\%]} &\multicolumn{ 1}{c|}{$\ttau$ [steps]} &\multicolumn{ 1}{c|}{Recall [\%]} &\multicolumn{ 1}{c||}{$\ttau$ [steps]}&  &  \\
		\midrule
		\cnn + \bisevena & $80.4 \pm 0.4$ &$272 \pm 6$& {\boldmath$82.8 \pm 0.4$} &\boldmath$16.8 \pm 0.3$& 3000 & 7 & 0.7 \\
		\cnn + \bisevenb & $79.8 \pm 0.7$ &$280 \pm 10 $& {\boldmath$83.8 \pm 0.4$} &\boldmath$12.5 \pm 0.2$& 3000 & 7 & 0.7 \\
		\cnn + \bisixty & $73.3 \pm 0.1$ &$2200 \pm 100$ & {\boldmath$87.6 \pm 0.5$} &\boldmath$122 \pm 4$ & 3500 & 60 & 0.7 \\
		\vgg + \bisevena & $81.0 \pm 0.5$ & $580\pm 20$& {\boldmath$88.8 \pm 0.5$} &\boldmath$200 \pm 5$& 1500 & 7  & 0.7\\
		\cnn + \multi & $41.6 \pm 0.5$ &$170 \pm 6$& {\boldmath$50.4 \pm 0.4$} & \boldmath$15.5 \pm 0.2$ &1500& 100 & 0.3 \\
		\res + \multib & $13.5 \pm 0.2$ &$640 \pm 60$& {\boldmath$14.4 \pm 0.2$} & \boldmath$310 \pm 50$ &1500& 100 & 0.3 \\
		\bottomrule
	\end{tabular}
	\label{tab:GD}
	\vskip-10pt
\end{table*}

\begin{figure}[tbh!]
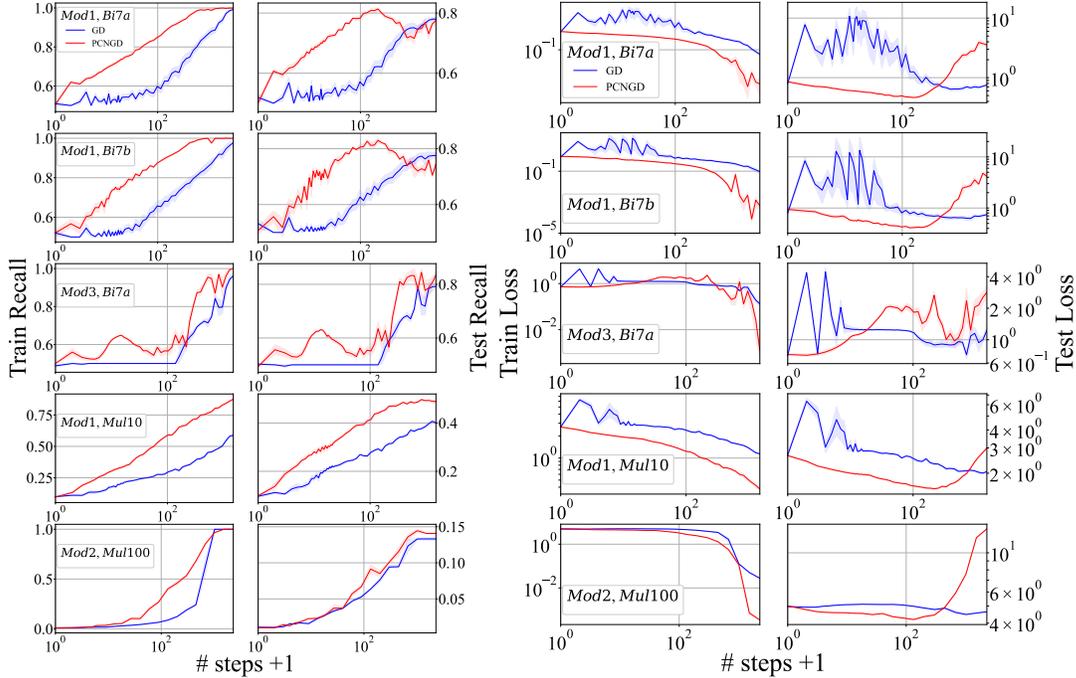

    \centering
    \includegraphics[width=.375\textwidth]{figures/Rec_Det_Group.pdf}
    \includegraphics[width=.45\textwidth]{figures/Loss_Det_Group.pdf}
    \caption{Recall and Loss function for GD and PCNGD algorithms. }
    \label{fig:det_group}
\end{figure}

Fig.~\ref{fig:det_group} shows the recall and loss for several combinations of the models and datasets presented in \ref{Appendix:Data&HP}. 
The careful reader can remark that the learning curves for PCNGD are not strictly monotonic as one could expect from Theorem~\ref{th:PCNGDdecreasing}. The reason is that the monotonous decrease of the loss is only guaranteed for sufficiently small learning rates. For some of our runs, the HP tuning yielded larger learning rates, for which the per-class loss is non-monotonic. This reveals a trade-off between a strictly monotonic decrease of the loss and a larger step size.

\begin{figure}[htb!]
    \centering
    \includegraphics[width=.8\columnwidth]{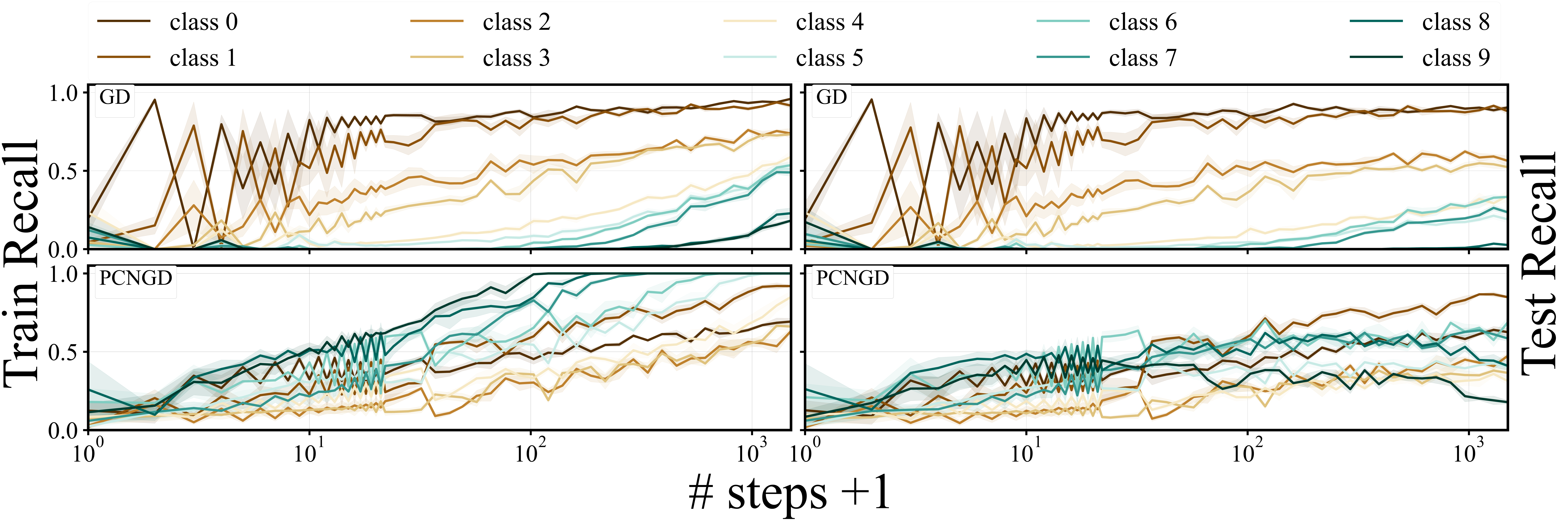}
    \caption{
    Comparison GD (top) and with PCNGD (bottom) Model: \cnn\!, dataset: \multi\!.  The majority class is class 0, the minority class is class 9. }
    \label{fig:Multi_PC_Comp_Det}
\end{figure}

In Fig.~\ref{fig:Multi_PC_Comp_Det} we show an example of the per-class curves in a multiclass problem. As in the binary setting, the GD dynamics exhibit the MID, which is instead not there with PCNGD.

{We summarize the results of these experiments in Tab.~\ref{tab:GD}, where we show the peak test recall after a fixed number of iterations, and the number of time steps, $\ttau$\footnote{Since measurements during the dynamics are not taken at every time step, errors on the quantity $\ttau$ are the $\min$ between the standard error and the interval between measurements.}, that the model requires before the test recall reaches a value $R^*$, also reported on the table. While the test recall is an indicator of the performance at convergence, $\ttau$ is an indicator of the speed of convergence, which is particularly relevant when one needs to assess performance over a limited number of iterations, as in hyperparameter tuning or when training is especially expensive.
We also indicate the total number of iterations performed in each run, and the imbalance ratio, $\rho$, defined as the ratio between the number of examples of majority and minority classes.\footnote{We use this definition because it is the one usually adopted in the literature. However, with this definition, the same value of $\rho$ indicates a larger imbalance when the number of classes is small, than when it is large.}
From Tab.~\ref{tab:GD}, we see that PCNGD systematically outperforms GD in terms of convergence speed, in agreement with our theory.
In addition, also the test recall is consistently higher: we elaborate on this in Sec.~\ref{sec:generalize}.}

\subsection{Stochastic optimization}
Here we show additional experiments regarding stochastic optimization algorithms, as described in the main part of the paper.
\begin{figure}[tbh!]
    \centering
    \includegraphics[width=.45\textwidth]{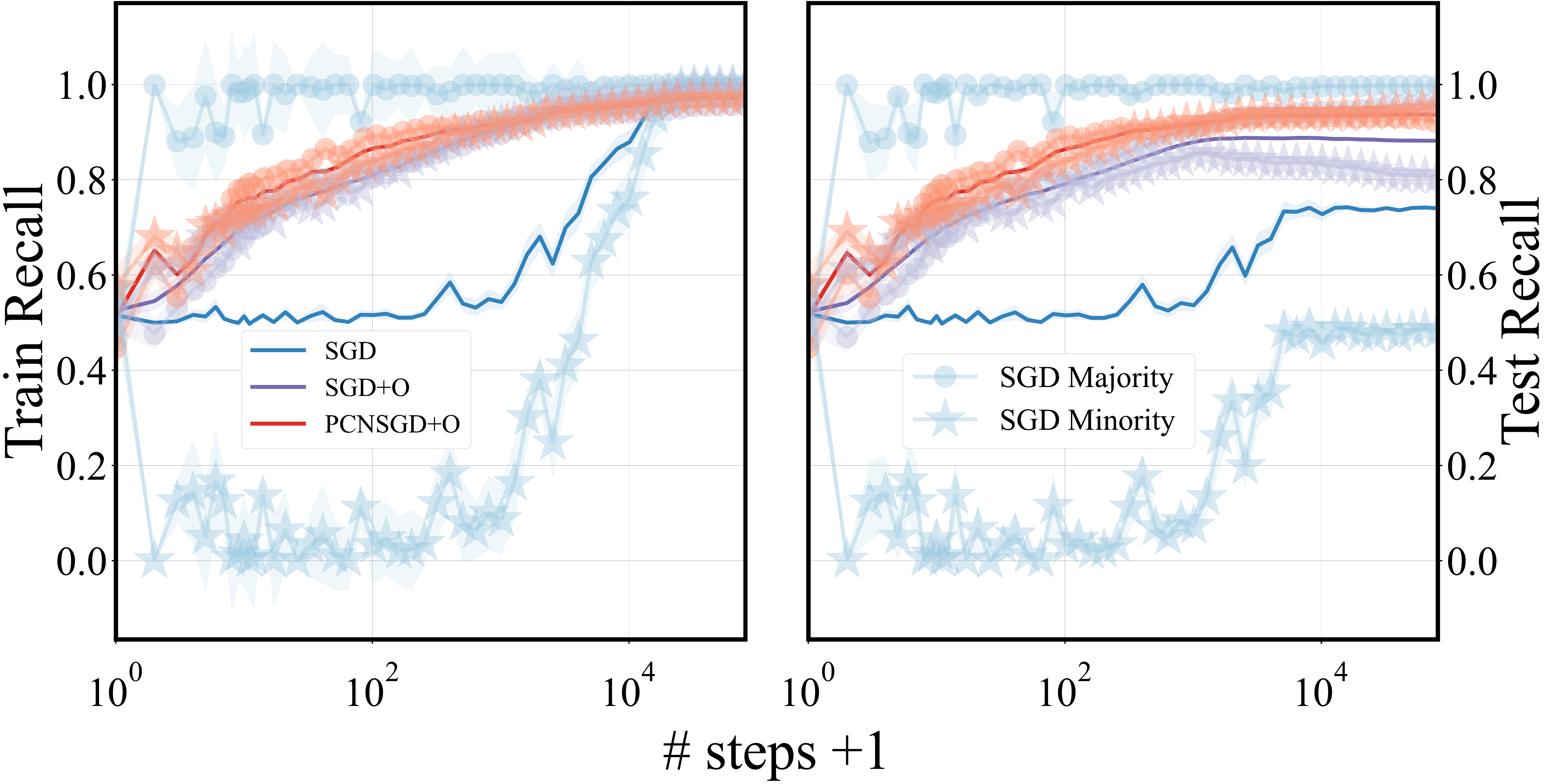}
    \includegraphics[width=.455\textwidth]{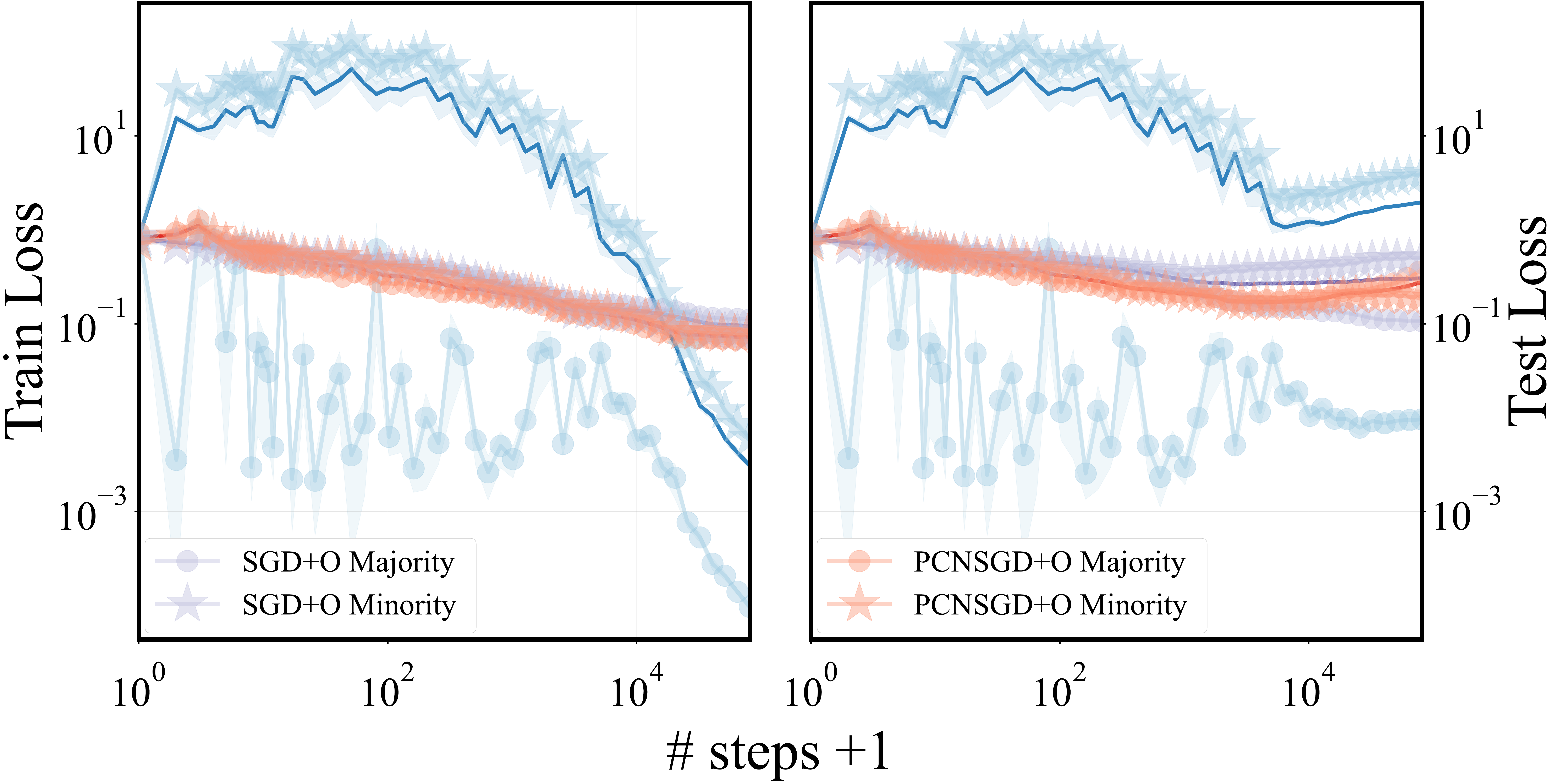}
    \caption{Recall and Loss function for GD and PCNGD algorithms. Model: \cnn\!, dataset: \bisixty\!. The SGD per-class curves clearly display the MID, while those related to the two oversampled algorithms follow closely the macro-averaged trends. }
    \label{figure:60Stoc}
\end{figure}
\begin{figure}[tbh!]
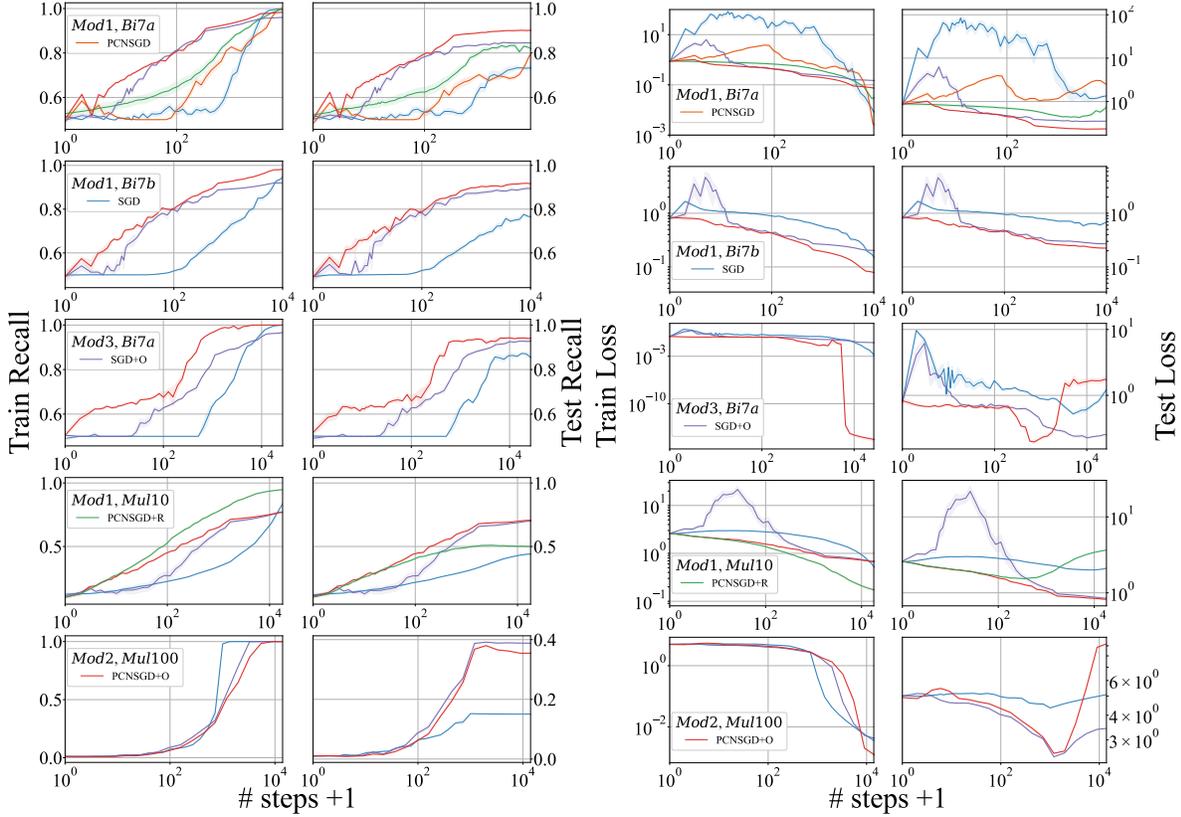

    \centering
    \includegraphics[width=.45\textwidth]{figures/Rec_Stoc_Group.pdf}
    \includegraphics[width=.455\textwidth]{figures/Loss_Stoc_Group.pdf}
    \caption{Recall and Loss function for SGD and PCNSGD algorithms.}
    \label{fig:stoc_group}
\end{figure}
In Fig.~\ref{figure:60Stoc}, the recall and loss curves are shown for the Model: \cnn\!, dataset: \bisixty\!.
\begin{table*}[tb]\footnotesize
	\centering
	\tabcolsep 3pt
	\caption{Same as Tab.~\ref{tab:GD}, but for (PCN)SGD dynamics. 
}
	\vspace{-2pt}
	\begin{tabular}{l||c|c|c|c|c|c||c||c||c}
		\addlinespace
		\toprule
	Models+Dataset	& \multicolumn{ 2}{c|}{SGD}    & \multicolumn{ 2}{c|}{SGD+O} &\multicolumn{ 2}{c||}{PCNSGD+O}             &  \# iterations         & $\rho$   &  $R^*$     \\
		\cmidrule{2-7}
		& \multicolumn{ 1}{c|}{Recall} &\multicolumn{ 1}{c|}{$\ttau$} & \multicolumn{ 1}{c|}{Recall} &\multicolumn{ 1}{c|}{$\ttau$} &\multicolumn{ 1}{c|}{Recall} &\multicolumn{ 1}{c||}{$\ttau$} &  &  \\
		\midrule
		\cnn + \bisevena & $77.0 \pm 0.8$ &$1400 \pm 300$& $84.9 \pm 0.2$ &$20.3 \pm 0.4$& \boldmath$90.9 \pm 0.2$ &\boldmath$14.5 \pm 0.2$& 10000 & 7 & 0.7 \\
		\cnn + \bisevenb & $78.4 \pm 0.5$ &$2200 \pm 100$& $90.3 \pm 0.5$ &$24 \pm 1$&{\boldmath$92.3 \pm 0.2$} &\boldmath$14.9 \pm 0.6 $& 10000 & 7 & 0.7 \\
		\cnn + \bisixty & $81 \pm 1$ &$1540 \pm 60$& $89.3\pm 0.1$&$10.8 \pm 0.2$& \boldmath$94.2 \pm 0.1$ &\boldmath$5.4 \pm 0.1$& 80000 & 60 & 0.7 \\
		\vgg + \bisevena & $88.3 \pm 0.3$&$2400 \pm 100$& $93.5 \pm 0.2$&$410 \pm 20$&\boldmath$95.1 \pm 0.3$ &\boldmath$111 \pm 6$& 25000 & 7 & 0.7 \\
		\cnn + \multi & $44.6 \pm 0.5$ &$5300 \pm 500$& $71.1 \pm 0.7$&$250 \pm 10$& {\boldmath$71.8 \pm 0.4$} &\boldmath$62 \pm 4$& 18000 & 100  & 0.4\\
		\res + \multib & $15.2 \pm 0.1$ &$420 \pm 30$& \boldmath$39.2 \pm 0.2 $& \boldmath$140 \pm 20$& $38.0 \pm 0.2$ &$210 \pm 30$& 18000 & 100 & 0.1 \\
		
		\bottomrule
	\end{tabular}
	\label{tab:SGD}
	\vskip-10pt
\end{table*}
{In Fig.~\ref{fig:stoc_group} we show macro averaged recalls and loss functions for stochastic algorithms. We do not always show the curves for PCNSGD+R for all the experiments because they are computationally expensive. In most of the cases presented, PCNSGD+O outperforms all the other solutions. The initial spike in the loss for the SGD+O algorithm can be reduced by decreasing the learning rate, but at the expense of a worse final performance.}
{The improvement of PCNSGD+O with respect to the other algorithms is more marked when the data imbalance is increased, as we show in Fig.~\ref{figure:60Stoc}, where there is a significant gain with respect to SGD+O along the entire dynamics.} {Similar to the deterministic case, the results of the various experiments are summarized in Tab.~\ref{tab:SGD}. }
We do not report PCNSGD+R on Tab.~\ref{tab:SGD} because of its high computational cost, but we still show some extra PCNSGD+R runs in Fig~\ref{fig:stoc_group}.

\begin{figure}[htb!]
    \centering
    \includegraphics[width=.8\columnwidth]{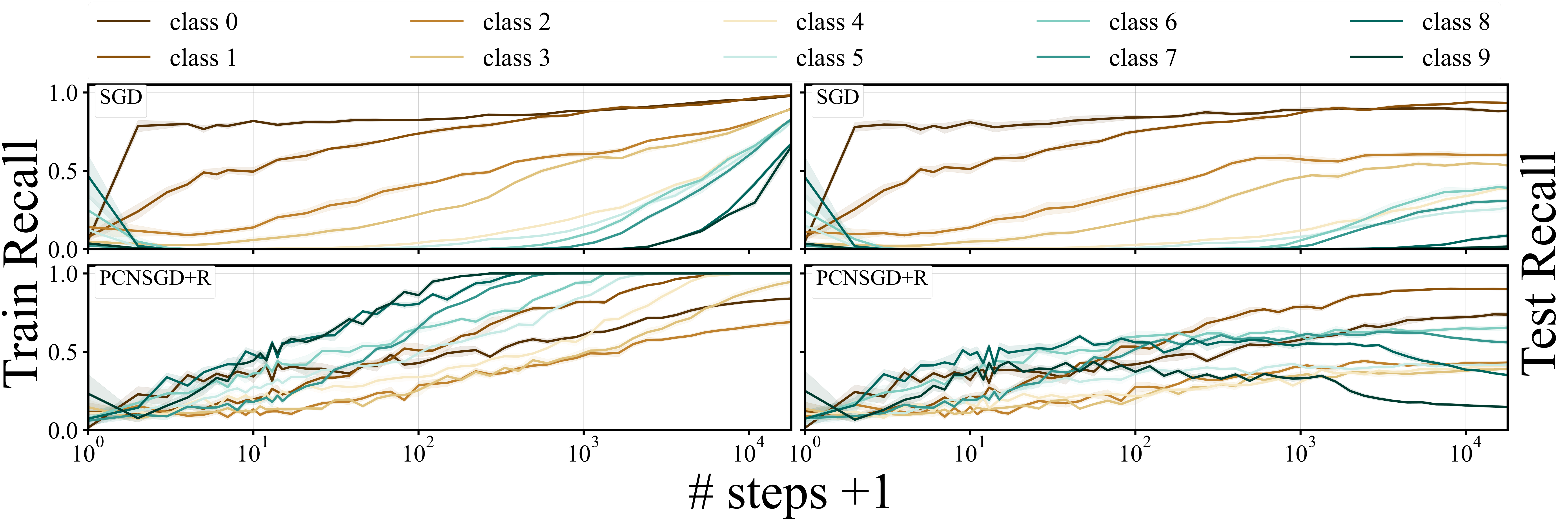}
    \caption{
    Comparison SGD (top) and with PCNSGD+R (bottom) Model: \cnn\!\!, dataset: \multi\!. The majority class is class 0, the minority class is class 9. }
    \label{fig:Multi_PC_Comp}
\end{figure}

\paragraph{The effect of PCNSGD+R on the single classes}
In Fig.~\ref{fig:Multi_PC_Comp} we show a comparison of the per-class learning curves of SGD and PCNSGD+R.
The improvement in class-averaged performance reflects, as in the deterministic case, a more balanced growth in single-class performance. In SGD, the beginning of the dynamics is characterized by a sudden growth of the majority classes and a degrowth of the minority ones, analogously to the \drop we showed in Fig.~\ref{fig:warmup} in the case of binary classification. After the initial boost, the growth of the majority classes' recall slows down and the remaining classes gradually begin to grow following the order set by the imbalance of the data. 
In the case of PCNSGD+R, on the contrary, a more regular growth is observed among the various classes. Note that, while in SGD the fastest classes are the majority ones, with PCNSGD+R the fastest are the minority, but the majority class is not the slowest.

\section{Limitations and Ethics}\label{app:limitations}

\paragraph{Limitations}
We can identify the following limitations to our work:
\begin{itemize}
\item Our theory does not cover dropout, but it could probably be extended by trying to include the averaging effect of Dropout. Alternatively, one could use the formulation of~\cite{arora2020theory} that shows that the Dropout objective is a regularized version of the original objective.
In addition, we did run experiments on a VGG network (which has dropout), the \vgg model, and found results that are consistent with the rest of the architectures.

\item Our theory does not cover non-differentiable activations: There are various standard ways to deal with non-differentiable functions in optimization, \textit{e.g.} subgradients, smoothing techniques, etc. We expect that all these techniques are applicable to PCN(S)GD, and that the convergence theorems can be adapted accordingly. Furthermore, note that the models \cnn and \res have ReLU activations (which is non-differentiable at a single point), so at least empirically we see that our results also hold with ReLUs.
\end{itemize}

\paragraph{Ethics}
Class imbalance, often present in real datasets used to train neural networks, not only impacts performance but also brings with it a number of ethical questions related to representative discrimination and fairness \cite{mehrabi2021survey, buolamwini2018gender, danks2017algorithmic, de2019does, yapo2018ethical}.

The \href{https://www.cs.toronto.edu/~kriz/cifar.html}{CIFAR} datasets, are subsets of the 80 million tiny
images, which is formally withdrawn since it contains some derogatory terms as
categories and offensive images (\href{ http://groups.csail.mit.edu/vision/TinyImages/}{ http://groups.csail.mit.edu/vision/TinyImages/}). However, note that none of the experiments described in the paper was performed on the overall tiny images dataset: the said derogatory images are not present in CIFAR10 and CIFAR100.


%

\end{document}

%% file: defs.tex
\newcommand{\bigO}{\mathcal{O}}

\newcommand{\sqnorm}[1]{\left\|#1\right\|^2}

\newcommand{\n}{\tilde{n}}

\newcommand{\x}{{\bf x}}
\newcommand{\y}{{\bf y}}

\newcommand{\D}{{\mathcal D}}
\newcommand{\E}{{\mathbb E}}

\newcommand{\R}{{\mathbb{R}}}

\newcommand{\Z}{{\mathbf Z}}

\newcommand{\PCMBG}{\nabla_{\tilde n} f^{(l)}(\x_t)}
\newcommand{\APCMBG}{\nabla f_{\tilde n_l}^{(l)}(\x_t)}

\newtcolorbox{mybox}[2][]
{
  colframe = #2!25,
  colback  = #2!25!white!25,
  left=1mm,
  top=1mm,
  #1,
  breakable
}

\newenvironment{theorembox}
   {\begin{mybox}{gray}\begin{theorem}}
   {\end{theorem}\end{mybox}}

%% file: icml2023.bbl
\begin{thebibliography}{55}
\providecommand{\natexlab}[1]{#1}
\providecommand{\url}[1]{\texttt{#1}}
\expandafter\ifx\csname urlstyle\endcsname\relax
  \providecommand{\doi}[1]{doi: #1}\else
  \providecommand{\doi}{doi: \begingroup \urlstyle{rm}\Url}\fi

\bibitem[Qui{\~n}onero-Candela et~al.(2009)Qui{\~n}onero-Candela, Sugiyama,
  Lawrence, and Schwaighofer]{quinonero:09}
Joaquin Qui{\~n}onero-Candela, Masashi Sugiyama, Neil~D Lawrence, and Anton
  Schwaighofer.
\newblock \emph{Dataset shift in machine learning}.
\newblock Mit Press, 2009.

\bibitem[Moreno-Torres et~al.(2012)Moreno-Torres, Raeder, Alaiz-Rodríguez,
  Chawla, and Herrera]{morenotorres:12}
Jose~G. Moreno-Torres, Troy Raeder, Rocío Alaiz-Rodríguez, Nitesh~V. Chawla,
  and Francisco Herrera.
\newblock A unifying view on dataset shift in classification.
\newblock \emph{Pattern Recognition}, 45\penalty0 (1):\penalty0 521--530, 2012.
\newblock ISSN 0031-3203.
\newblock \doi{https://doi.org/10.1016/j.patcog.2011.06.019}.
\newblock URL
  \url{https://www.sciencedirect.com/science/article/pii/S0031320311002901}.

\bibitem[Van~Horn and Perona(2017)]{van2017devil}
Grant Van~Horn and Pietro Perona.
\newblock The devil is in the tails: Fine-grained classification in the wild.
\newblock \emph{arXiv preprint arXiv:1709.01450}, 2017.

\bibitem[Feldman(2020)]{feldman2020does}
Vitaly Feldman.
\newblock Does learning require memorization? a short tale about a long tail.
\newblock In \emph{Proceedings of the 52nd Annual ACM SIGACT Symposium on
  Theory of Computing}, pages 954--959, 2020.

\bibitem[D'souza et~al.(2021)D'souza, Nussbaum, Agarwal, and Hooker]{d2021tale}
Daniel D'souza, Zach Nussbaum, Chirag Agarwal, and Sara Hooker.
\newblock A tale of two long tails.
\newblock \emph{arXiv preprint arXiv:2107.13098}, 2021.

\bibitem[Liu et~al.(2017)Liu, Wang, Zhang, Chen, and Xiang]{liu:17}
Shigang Liu, Yu~Wang, Jun Zhang, Chao Chen, and Yang Xiang.
\newblock Addressing the class imbalance problem in twitter spam detection
  using ensemble learning.
\newblock \emph{Computers \& Security}, 69:\penalty0 35--49, 2017.

\bibitem[Makki et~al.(2019)Makki, Assaghir, Taher, Haque, Hacid, and
  Zeineddine]{makki:19}
Sara Makki, Zainab Assaghir, Yehia Taher, Rafiqul Haque, Mohand-Said Hacid, and
  Hassan Zeineddine.
\newblock An experimental study with imbalanced classification approaches for
  credit card fraud detection.
\newblock \emph{IEEE Access}, 7:\penalty0 93010--93022, 2019.

\bibitem[Kyathanahally et~al.(2021)Kyathanahally, Hardeman, Merz, Bulas, Reyes,
  Isles, Pomati, and Baity-Jesi]{kyathanahally:21}
Sreenath~P Kyathanahally, Thomas Hardeman, Ewa Merz, Thea Bulas, Marta Reyes,
  Peter Isles, Francesco Pomati, and Marco Baity-Jesi.
\newblock Deep learning classification of lake zooplankton.
\newblock \emph{Frontiers in microbiology}, page 3226, 2021.
\newblock \doi{10.3389/fmicb.2021.746297}.

\bibitem[He and Garcia(2009)]{he:09}
Haibo He and Edwardo~A Garcia.
\newblock Learning from imbalanced data.
\newblock \emph{IEEE Transactions on knowledge and data engineering},
  21\penalty0 (9):\penalty0 1263--1284, 2009.

\bibitem[Huang et~al.(2016)Huang, Li, Loy, and Tang]{huang:16}
Chen Huang, Yining Li, Chen~Change Loy, and Xiaoou Tang.
\newblock Learning deep representation for imbalanced classification.
\newblock In \emph{Proceedings of the IEEE conference on computer vision and
  pattern recognition}, pages 5375--5384, 2016.

\bibitem[Japkowicz(2000)]{japkowicz:00}
Nathalie Japkowicz.
\newblock The class imbalance problem: Significance and strategies.
\newblock In \emph{Proc. of the Int’l Conf. on Artificial Intelligence},
  volume~56. Citeseer, 2000.

\bibitem[Alshammari et~al.(2022)Alshammari, Wang, Ramanan, and
  Kong]{alshammari:22}
Shaden Alshammari, Yu-Xiong Wang, Deva Ramanan, and Shu Kong.
\newblock Long-tailed recognition via weight balancing.
\newblock In \emph{Proceedings of the IEEE/CVF Conference on Computer Vision
  and Pattern Recognition}, pages 6897--6907, 2022.

\bibitem[Tang et~al.(2020)Tang, Huang, and Zhang]{tang:20}
Kaihua Tang, Jianqiang Huang, and Hanwang Zhang.
\newblock Long-tailed classification by keeping the good and removing the bad
  momentum causal effect.
\newblock \emph{Advances in Neural Information Processing Systems},
  33:\penalty0 1513--1524, 2020.

\bibitem[Anand et~al.(1993)Anand, Mehrotra, Mohan, and Ranka]{anand:93}
R.~Anand, K.G. Mehrotra, C.K. Mohan, and S.~Ranka.
\newblock An improved algorithm for neural network classification of imbalanced
  training sets.
\newblock \emph{IEEE Transactions on Neural Networks}, 4\penalty0 (6):\penalty0
  962--969, 1993.
\newblock \doi{10.1109/72.286891}.

\bibitem[Ye et~al.(2021)Ye, Zhan, and Chao]{ye:21}
Han-Jia Ye, De-Chuan Zhan, and Wei-Lun Chao.
\newblock Procrustean training for imbalanced deep learning.
\newblock In \emph{Proceedings of the IEEE/CVF International Conference on
  Computer Vision}, pages 92--102, 2021.

\bibitem[Chen et~al.(2018)Chen, Badrinarayanan, Lee, and
  Rabinovich]{chen2018gradnorm}
Zhao Chen, Vijay Badrinarayanan, Chen-Yu Lee, and Andrew Rabinovich.
\newblock Gradnorm: Gradient normalization for adaptive loss balancing in deep
  multitask networks.
\newblock In \emph{International conference on machine learning}, pages
  794--803. PMLR, 2018.

\bibitem[Yu et~al.(2020)Yu, Kumar, Gupta, Levine, Hausman, and
  Finn]{yu2020gradient}
Tianhe Yu, Saurabh Kumar, Abhishek Gupta, Sergey Levine, Karol Hausman, and
  Chelsea Finn.
\newblock Gradient surgery for multi-task learning.
\newblock \emph{Advances in Neural Information Processing Systems},
  33:\penalty0 5824--5836, 2020.

\bibitem[Ghadimi and Lan(2013)]{ghadimi:13}
Saeed Ghadimi and Guanghui Lan.
\newblock Stochastic first-and zeroth-order methods for nonconvex stochastic
  programming.
\newblock \emph{SIAM Journal on Optimization}, 23\penalty0 (4):\penalty0
  2341--2368, 2013.

\bibitem[Nesterov and Polyak(2006)]{nesterov2006cubic}
Yurii Nesterov and Boris~T Polyak.
\newblock Cubic regularization of newton method and its global performance.
\newblock \emph{Mathematical Programming}, 108\penalty0 (1):\penalty0 177--205,
  2006.

\bibitem[Karimi et~al.(2016)Karimi, Nutini, and Schmidt]{karimi2016linear}
Hamed Karimi, Julie Nutini, and Mark Schmidt.
\newblock Linear convergence of gradient and proximal-gradient methods under
  the polyak-{\l}ojasiewicz condition.
\newblock In \emph{Joint European Conference on Machine Learning and Knowledge
  Discovery in Databases}, pages 795--811. Springer, 2016.

\bibitem[Liu et~al.(2022)Liu, Zhu, and Belkin]{liu2022loss}
Chaoyue Liu, Libin Zhu, and Mikhail Belkin.
\newblock Loss landscapes and optimization in over-parameterized non-linear
  systems and neural networks.
\newblock \emph{Applied and Computational Harmonic Analysis}, 59:\penalty0
  85--116, 2022.

\bibitem[Kyathanahally et~al.(2022)Kyathanahally, Hardeman, Reyes, Merz, Bulas,
  Pomati, and Baity-Jesi]{kyathanahally2022ensembles}
S~Kyathanahally, T~Hardeman, M~Reyes, E~Merz, T~Bulas, F~Pomati, and
  M~Baity-Jesi.
\newblock Ensembles of vision transformers as a new paradigm for automated
  classification in ecology.
\newblock \emph{arXiv preprint arXiv:2203.01726}, 2022.

\bibitem[Izmailov et~al.(2018)Izmailov, Podoprikhin, Garipov, Vetrov, and
  Wilson]{izmailov:18}
Pavel Izmailov, Dmitrii Podoprikhin, Timur Garipov, Dmitry Vetrov, and
  Andrew~Gordon Wilson.
\newblock Averaging weights leads to wider optima and better generalization.
\newblock \emph{arXiv preprint arXiv:1803.05407}, 2018.

\bibitem[Darling(1956)]{darling:56}
Donald~A Darling.
\newblock Bv gnedenko and an kolmogorov, limit distributions for sums of
  independent random variables.
\newblock \emph{Bulletin of the American Mathematical Society}, 62\penalty0
  (1):\penalty0 50--52, 1956.

\bibitem[Lam et~al.(2011)Lam, Blanchet, Burch, and Bazant]{lam2011corrections}
Henry Lam, Jose Blanchet, Damian Burch, and Martin~Z Bazant.
\newblock Corrections to the central limit theorem for heavy-tailed probability
  densities.
\newblock \emph{Journal of Theoretical Probability}, 24\penalty0 (4):\penalty0
  895--927, 2011.

\bibitem[Lin et~al.(2017)Lin, Goyal, Girshick, He, and
  Doll{\'a}r]{lin2017focal}
Tsung-Yi Lin, Priya Goyal, Ross Girshick, Kaiming He, and Piotr Doll{\'a}r.
\newblock Focal loss for dense object detection.
\newblock In \emph{Proceedings of the IEEE international conference on computer
  vision}, pages 2980--2988, 2017.

\bibitem[Cao et~al.(2019{\natexlab{a}})Cao, Wei, Gaidon, Arechiga, and
  Ma]{cao2019learning}
Kaidi Cao, Colin Wei, Adrien Gaidon, Nikos Arechiga, and Tengyu Ma.
\newblock Learning imbalanced datasets with label-distribution-aware margin
  loss.
\newblock \emph{Advances in neural information processing systems}, 32,
  2019{\natexlab{a}}.

\bibitem[Leng et~al.(2022)Leng, Tan, Liu, Cubuk, Shi, Cheng, and
  Anguelov]{leng2022polyloss}
Zhaoqi Leng, Mingxing Tan, Chenxi Liu, Ekin~Dogus Cubuk, Xiaojie Shi, Shuyang
  Cheng, and Dragomir Anguelov.
\newblock Polyloss: A polynomial expansion perspective of classification loss
  functions.
\newblock \emph{arXiv preprint arXiv:2204.12511}, 2022.

\bibitem[Sagawa et~al.(2020)Sagawa, Raghunathan, Koh, and
  Liang]{sagawa2020investigation}
Shiori Sagawa, Aditi Raghunathan, Pang~Wei Koh, and Percy Liang.
\newblock An investigation of why overparameterization exacerbates spurious
  correlations.
\newblock In \emph{International Conference on Machine Learning}, pages
  8346--8356. PMLR, 2020.

\bibitem[Jin et~al.(2017)Jin, Ge, Netrapalli, Kakade, and
  Jordan]{jin2017escape}
Chi Jin, Rong Ge, Praneeth Netrapalli, Sham~M Kakade, and Michael~I Jordan.
\newblock How to escape saddle points efficiently.
\newblock In \emph{International conference on machine learning}, pages
  1724--1732. PMLR, 2017.

\bibitem[Daneshmand et~al.(2018)Daneshmand, Kohler, Lucchi, and
  Hofmann]{daneshmand2018escaping}
Hadi Daneshmand, Jonas Kohler, Aurelien Lucchi, and Thomas Hofmann.
\newblock Escaping saddles with stochastic gradients.
\newblock In \emph{International Conference on Machine Learning}, pages
  1155--1164. PMLR, 2018.

\bibitem[Nesterov(2003)]{nesterov2003introductory}
Yurii Nesterov.
\newblock \emph{Introductory lectures on convex optimization: A basic course},
  volume~87.
\newblock Springer Science \& Business Media, 2003.

\bibitem[Chawla et~al.(2004)Chawla, Japkowicz, and Kotcz]{chawla2004special}
Nitesh~V Chawla, Nathalie Japkowicz, and Aleksander Kotcz.
\newblock Special issue on learning from imbalanced data sets.
\newblock \emph{ACM SIGKDD explorations newsletter}, 6\penalty0 (1):\penalty0
  1--6, 2004.

\bibitem[Johnson and Khoshgoftaar(2019)]{johnson:19}
Justin~M Johnson and Taghi~M Khoshgoftaar.
\newblock Survey on deep learning with class imbalance.
\newblock \emph{Journal of Big Data}, 6\penalty0 (1):\penalty0 1--54, 2019.

\bibitem[Chawla et~al.(2002)Chawla, Bowyer, Hall, and Kegelmeyer]{chawla:02}
Nitesh~V Chawla, Kevin~W Bowyer, Lawrence~O Hall, and W~Philip Kegelmeyer.
\newblock Smote: synthetic minority over-sampling technique.
\newblock \emph{Journal of artificial intelligence research}, 16:\penalty0
  321--357, 2002.

\bibitem[Mani and Zhang(2003)]{mani2003knn}
Inderjeet Mani and I~Zhang.
\newblock knn approach to unbalanced data distributions: a case study involving
  information extraction.
\newblock In \emph{Proceedings of workshop on learning from imbalanced
  datasets}, volume 126. ICML United States, 2003.

\bibitem[Chou et~al.(2020)Chou, Chang, Pan, Wei, and Juan]{chou:20}
Hsin-Ping Chou, Shih-Chieh Chang, Jia-Yu Pan, Wei Wei, and Da-Cheng Juan.
\newblock Remix: rebalanced mixup.
\newblock In \emph{European Conference on Computer Vision}, pages 95--110.
  Springer, 2020.

\bibitem[Ren et~al.(2018)Ren, Zeng, Yang, and Urtasun]{ren:18}
Mengye Ren, Wenyuan Zeng, Bin Yang, and Raquel Urtasun.
\newblock Learning to reweight examples for robust deep learning.
\newblock In \emph{International Conference on Machine Learning}, pages
  4334--4343. PMLR, 2018.

\bibitem[An et~al.(2020)An, Ying, and Zhu]{an2020resampling}
Jing An, Lexing Ying, and Yuhua Zhu.
\newblock Why resampling outperforms reweighting for correcting sampling bias
  with stochastic gradients.
\newblock \emph{arXiv preprint arXiv:2009.13447}, 2020.

\bibitem[Dong et~al.(2018)Dong, Gong, and Zhu]{dong:18}
Qi~Dong, Shaogang Gong, and Xiatian Zhu.
\newblock Imbalanced deep learning by minority class incremental rectification.
\newblock \emph{IEEE transactions on pattern analysis and machine
  intelligence}, 41\penalty0 (6):\penalty0 1367--1381, 2018.

\bibitem[Menon et~al.(2020)Menon, Jayasumana, Rawat, Jain, Veit, and
  Kumar]{menon:20}
Aditya~Krishna Menon, Sadeep Jayasumana, Ankit~Singh Rawat, Himanshu Jain,
  Andreas Veit, and Sanjiv Kumar.
\newblock Long-tail learning via logit adjustment.
\newblock \emph{arXiv:2007.07314}, 2020.

\bibitem[Zhu et~al.(2022)Zhu, Wang, Chen, Chen, and Jiang]{zhu:22}
Jianggang Zhu, Zheng Wang, Jingjing Chen, Yi-Ping~Phoebe Chen, and Yu-Gang
  Jiang.
\newblock Balanced contrastive learning for long-tailed visual recognition.
\newblock In \emph{Proceedings of the IEEE/CVF Conference on Computer Vision
  and Pattern Recognition}, pages 6908--6917, 2022.

\bibitem[Hendrycks et~al.(2019)Hendrycks, Lee, and Mazeika]{hendrycks:19}
Dan Hendrycks, Kimin Lee, and Mantas Mazeika.
\newblock Using pre-training can improve model robustness and uncertainty.
\newblock In \emph{International Conference on Machine Learning}, pages
  2712--2721. PMLR, 2019.

\bibitem[Cao et~al.(2019{\natexlab{b}})Cao, Wei, Gaidon, Arechiga, and
  Ma]{cao:19}
Kaidi Cao, Colin Wei, Adrien Gaidon, Nikos Arechiga, and Tengyu Ma.
\newblock Learning imbalanced datasets with label-distribution-aware margin
  loss.
\newblock \emph{arXiv:1906.07413}, 2019{\natexlab{b}}.

\bibitem[He et~al.(2016)He, Zhang, Ren, and Sun]{he2016deep}
Kaiming He, Xiangyu Zhang, Shaoqing Ren, and Jian Sun.
\newblock Deep residual learning for image recognition.
\newblock In \emph{Proceedings of the IEEE conference on computer vision and
  pattern recognition}, pages 770--778, 2016.

\bibitem[Wu and He(2018)]{wu2018group}
Yuxin Wu and Kaiming He.
\newblock Group normalization.
\newblock In \emph{Proceedings of the European conference on computer vision
  (ECCV)}, pages 3--19, 2018.

\bibitem[Ioffe and Szegedy(2015)]{ioffe2015batch}
Sergey Ioffe and Christian Szegedy.
\newblock Batch normalization: Accelerating deep network training by reducing
  internal covariate shift.
\newblock In \emph{International conference on machine learning}, pages
  448--456. PMLR, 2015.

\bibitem[Simonyan and Zisserman(2014)]{simonyan2014very}
Karen Simonyan and Andrew Zisserman.
\newblock Very deep convolutional networks for large-scale image recognition.
\newblock \emph{arXiv preprint arXiv:1409.1556}, 2014.

\bibitem[Krizhevsky et~al.(2009)Krizhevsky, Hinton, et~al.]{cifar}
Alex Krizhevsky, Geoffrey Hinton, et~al.
\newblock Learning multiple layers of features from tiny images.
\newblock 2009.
\newblock URL \url{https://www.cs.toronto.edu/~kriz/cifar.html}.

\bibitem[Arora et~al.(2020)Arora, Arora, Bruna, Cohen, Du, Ge, Gunasekar, Jin,
  Lee, Ma, et~al.]{arora2020theory}
Raman Arora, Sanjeev Arora, Joan Bruna, Nadav Cohen, Simon Du, Rong Ge, Suriya
  Gunasekar, C~Jin, Jason Lee, Tengyu Ma, et~al.
\newblock Theory of deep learning, 2020.

\bibitem[Mehrabi et~al.(2021)Mehrabi, Morstatter, Saxena, Lerman, and
  Galstyan]{mehrabi2021survey}
Ninareh Mehrabi, Fred Morstatter, Nripsuta Saxena, Kristina Lerman, and Aram
  Galstyan.
\newblock A survey on bias and fairness in machine learning.
\newblock \emph{ACM Computing Surveys (CSUR)}, 54\penalty0 (6):\penalty0 1--35,
  2021.

\bibitem[Buolamwini and Gebru(2018)]{buolamwini2018gender}
Joy Buolamwini and Timnit Gebru.
\newblock Gender shades: Intersectional accuracy disparities in commercial
  gender classification.
\newblock In \emph{Conference on fairness, accountability and transparency},
  pages 77--91. PMLR, 2018.

\bibitem[Danks and London(2017)]{danks2017algorithmic}
David Danks and Alex~John London.
\newblock Algorithmic bias in autonomous systems.
\newblock In \emph{Ijcai}, volume~17, pages 4691--4697, 2017.

\bibitem[De~Vries et~al.(2019)De~Vries, Misra, Wang, and Van~der
  Maaten]{de2019does}
Terrance De~Vries, Ishan Misra, Changhan Wang, and Laurens Van~der Maaten.
\newblock Does object recognition work for everyone?
\newblock In \emph{Proceedings of the IEEE/CVF conference on computer vision
  and pattern recognition workshops}, pages 52--59, 2019.

\bibitem[Yapo and Weiss(2018)]{yapo2018ethical}
Adrienne Yapo and Joseph Weiss.
\newblock Ethical implications of bias in machine learning.
\newblock \emph{Proceedings of the 51st Hawaii International Conference on
  System Sciences}, page 5365, 2018.

\end{thebibliography}
